\documentclass[10pt,twocolumn,letterpaper]{article}

\usepackage{ijcb}
\usepackage{times}
\usepackage{epsfig}
\usepackage{graphicx}
\usepackage{amsmath}
\usepackage{amssymb}
\usepackage{hhline}
\usepackage{color, colortbl}
\usepackage{multirow}
\usepackage{times}
\usepackage{epsfig}
\usepackage{graphicx}
\usepackage{amsmath}
\usepackage{amssymb}
\usepackage{subcaption}
\usepackage[utf8]{inputenc}
\usepackage{lipsum} 
\usepackage{booktabs}
\usepackage{xcolor}
\usepackage{makecell}
\usepackage{url}

\newenvironment{tight_itemize}{
\begin{itemize}
  \setlength{\topsep}{0pt}
  \setlength{\itemsep}{2pt}
  \setlength{\parskip}{0pt}
  \setlength{\parsep}{0pt}
}{\end{itemize}}

\definecolor{lightgray}{gray}{0.9}
\definecolor{lightblue}{rgb}{0.93,0.95,1.0}
\definecolor{darkgreen}{rgb}{0.0,0.6,0.0}
\definecolor{blue}{rgb}{1, 0, 0}

\ijcbfinalcopy 

\begin{document}

\title{
Does Face Recognition Error Echo Gender Classification Error?
}

\author{Ying Qiu\\
University of Notre Dame\\

\and
Vítor Albiero\\
University of Notre Dame\\

\and
Michael C. King\\
Florida Insitute of Technology\\

\and
Kevin W. Bowyer\\
University of Notre Dame\\

}

\maketitle

\begin{abstract}
This paper is the first to explore the question of whether images that are classified incorrectly by a face analytics algorithm (\eg, gender classification) are any more or less likely to participate in an image pair that results in a face recognition error.
We analyze results from three different gender classification algorithms (one open-source and two commercial), and two face recognition algorithms (one open-source and one commercial), on image sets representing four demographic groups (African-American female and male, Caucasian female and male).
For impostor image pairs,
our results show that pairs in which one image has a gender classification error have a better impostor distribution than pairs in which both images have correct gender classification,
and so are {\it less} likely to generate a false match error.
For genuine image pairs, our results show that
individuals whose images have a mix of correct and incorrect gender classification have a worse genuine distribution (increased false non-match rate) compared to individuals whose images all have correct gender classification.
Thus, compared to images that generate correct gender classification, images that generate gender classification errors do generate a different pattern of recognition errors, both better (false match) and worse (false non-match).
\end{abstract}


\section{Introduction}

Media reports can seem oblivious to distinctions between face analytics and face recognition.
For example, articles titled ``Facial Recognition Is Accurate, if You’re a White Guy'' in The New York Times~\cite{NYT} and ``Face-recognition Software is Perfect – If You’re a White Man'' in New Scientist~\cite{Revell_New_Scientist} focus on discussing 
Buolamwini and Gebru's work on accuracy of commercial gender classification algorithms~\cite{gender_shades}
but make broad statements about face recognition technology in general.

Facial analytics -- such as classifying the gender, race, and age of a person from their face image --  is an active area of research~\cite{Dantcheva2016,Eidinger2014,Fu2014,Sariyanidi2015}.
Modern approaches train a deep convolutional neural network (CNN).
The trained CNN can then analyze a face image
to classify or estimate some attribute of the person, and a single image is sufficient to generate a result. 
Researchers have noted that accuracy of face analytics tools varies across demographic groups~\cite{gender_shades,Muthukumar_2018, NIST_gender}.

Face recognition differs from face analytics in fundamental ways.
Face recognition involves estimating the similarity of face appearance between two images, rather than analyzing a single image.
Also, two different types of errors can occur in face recognition, a false match 
(judging images of two different persons to be images of the same person)
or a false non-match (judging two different images of the same person to not be similar enough to be images of the same person).
As with face analytics, face recognition error rates vary across demographic groups~\cite{Albiero_WACVW_2020,Cook_TBIOM_2019,Klare_TIFS_2012,Krishnapriya_TTS_2020,Lu_TBIOM_2019,Grother_2019}.

We know of no previous work that investigates whether images that generate face analytics errors generate any different pattern of errors when used for face recognition.
In this first investigation into the topic, 
we analyze whether images that generate gender classification error have any different pattern of face recognition errors compared to images that generate correct gender classification.

It is important to acknowledge at the outset the sensitivity of discussing algorithms for ``gender classification'' or ``gender from face'' and the error patterns of such algorithms.
All work to date in this area (\eg,~\cite{gender_shades,Muthukumar_2018,NIST_gender}) 
assumes that a face image can be analyzed and given a gender classification of ``female'' or ``male'' and that this result can be determined as correct or incorrect by comparison to meta-data.
In pursuing this research, no disrespect is intended to persons who feel that ``gender'' is not comprehensively defined as either ``female'' or ``male''.

In summary, the contributions of this work are:
\begin{tight_itemize}
    \item We present the results of the first investigation into the question of whether images that generate face analytics errors have any different error pattern in terms of face recognition errors.
    \item We investigate the generality of the results across demographic dimensions: (female / male) x (African-American / Caucasian). 
    \item To facilitate transparency and reproducibility, in addition to results from representative commercial facial analytics tools and matcher, we also report results for an open-source gender-from-face algorithm and the open-source ArcFace matcher, and we use a dataset generally available to the research community.
\end{tight_itemize}

\section{Related Work}

There are two streams of related work.
One looks at demographic disparity in accuracy of face analytics, such as gender-from-face.
The other looks at demographic disparity in accuracy of face recognition.
We summarize selected works from each stream here, and point to a recent survey for a broad summary of related work~\cite{Drozdowski_TTS_2020}.

\subsection{Gender-from-face Across Demographics}

Buolamwini and Gebru~\cite{gender_shades} reported the accuracy of three commercial gender classification algorithms 
for a dataset of 1,270 images collected for government officials in three Nordic countries (Iceland, Finland, Sweden) and three African countries (Rwanda, Senegal, South Africa).
They named their image dataset the ``Pilot Parliaments Benchmark'' (PPB).
In Table 4 of~\cite{gender_shades}, they show that 2 of 3 commercial gender-from-face APIs have highest accuracy on the ``L M'' (lighter male) subset of PPB, with the third API having highest accuracy on the ``D M'' (darker male) subset, and that all three commercial APIs have lowest accuracy on the ``D F'' (darker female) subset.

Muthukumar et al.~\cite{Muthukumar_2018} followed up 
\cite{gender_shades} with analysis of a similar dataset.
They also reported that gender classification accuracy is higher for males than females, higher for lighter (Caucasian) faces,
and that accuracy is lowest for darker (African), female faces.
Based on a series of experiments, they conclude that gender classification is mostly stable across skin types, and that the skin type {\it by itself} has a minimal effect on the classification outcome.
Using contrastive post-hoc explanation techniques for neural networks, the authors suggest that differences in face shape (lip, eye and cheek structure), and use of cosmetics (lip and eye makeup) play a larger role.

Diversity in Faces~\cite{Merler_DiF} uses gender and age classification models to add meta-data to images from YFCC100M~\cite{Thomee_YFCC100M}, along with meta-data for face landmark points and measurements.
FairFace~\cite{Karkkainen_FairFace} uses Mechanical Turk to add gender, race and age group meta-data to images from YFCC100M and elsewhere and uses the created dataset to train better face analytic classifiers.
These and similar datasets do not have identity meta-data, and so they may be useful in facial analytics research but cannot generate impostor and genuine distributions needed to study face recognition.

\subsection{Face recognition across demographics}

Face recognition accuracy was reported to vary with demographics at least as early as FRVT 2002 \cite{frvt_2003}.
A well-known early study to report demographic differences in face recognition accuracy is Klare et al.~\cite{Klare_TIFS_2012}. 
Their study used the Pinellas County Sheriff’s Office (PCSO) dataset and multiple
matchers from before the wave of deep learning algorithms in face recognition.
(The PCSO dataset is not available to other researchers.)
They reported accuracy in terms of ROC curves and verification rate at fixed false match rate (FMR). They conclude that ``The female, Black, and younger cohorts are more difficult to recognize for all matchers used in this article (commercial, non-trainable, and trainable)''.

A recent NIST report on demographic differences in face recognition accuracy~\cite{Grother_2019} contains the largest results, in terms of number of face recognition algorithms evaluated and number of datasets tested on, in this area.
The conclusions are largely similar to those in Klare et al.~\cite{Klare_TIFS_2012}, in that accuracy for females is lower than for males,
and accuracy for African-Americans is lower than for Caucasians.

Krishnapriya et al.~\cite{Krishnapriya_TTS_2020} present results indicating that both the impostor distribution and the genuine distribution for females are shifted toward lower accuracy, relative to the distibrutions for males.
They also report on an experiment to determine if darker skin tone is the driving cause of lower accuracy for African-Americans and suggest that it is not. 

Lu et al.~\cite{Lu_TBIOM_2019} report on experiments with different CNN matchers and the IARPA Janus dataset.
They find that accuracy as measured by the ROC curve is worse for women than for men, and that the ROC curve is overall worse for darker skin tones but gets better for the darkest skin tone.

Cook et al.~\cite{Cook_TBIOM_2019} analyze errors from using a COTS face matcher with a dataset of images from eleven different image acquisition kiosks.
They report that ``lower (darker) skin reflectance was associated with lower efficiency (higher transaction times) and accuracy (lower mated similarity scores)'' in their study.

\section{Image Datasets}

A dataset that supports face recognition experiments should inherently also support face analytics experiments, 
but the reverse is not true.
We present results on gender-from-face for the PPB dataset~\cite{gender_shades}, but since face recognition results are not possible with PPB, the main results presented here are obtained with the MORPH dataset~\cite{morph_site, morph_paper}.
The MORPH dataset has been used extensively in  face aging research, and has also been used in the study of demographic variation in recognition accuracy 
\cite{albiero2020skin,Albiero_WACV_2020,Albiero_WACVW_2020,Albiero_IJCB_2021,Krishnapriya_CVPRW_2019, Krishnapriya_TTS_2020}.

MORPH contains mugshot-style images that are nominally frontal pose, neutral expression and acquired with controlled lighting and uniform gray background.
The dataset is generally available to the research community, so that other researchers can reproduce and extend results found on this dataset.
We curated MORPH 3 to remove a small number of duplicate images, twins, and mislabeled images.The dataset used contains substantial numbers of persons and images in four demographic categories: African-American females (A-A F), African-American males (A-A M), Caucasian females (C F), and Caucasian males (C M), as summarized in Table~\ref{table:Morph_numbers}.

\begin{table}
\centering
\small
\begin{tabular}{|l|r|r|r|r|}
\hline
Cohort & Number   & Number & Impostor &  Genuine  \\
 & Persons  & Images & Pairs    &  Pairs  \\
\hline
A-A M   & 8,839  & 56,245 &  1,581,426,316 & 295,574\\
\hline
C M     &  8,835 & 35,276 &  622,042,698 & 137,752\\
\hline
A-A F   & 5,929  & 24,857 &  308,840,189 & 82,607\\
\hline
C F     & 2,798  & 10,941 &  59,813,525 & 33,745\\

\hline
\end{tabular}
\caption{Demographic cohorts in MORPH 3: African-American male (A-A M),  Caucasian male (C M), African-American female (A-A F) and Caucasian female (C F).
Number of images is the number of possible gender-from-face results; images pairs in the impostor and genuine distributions is a function of number of persons and number of images per person. 
}
\label{table:Morph_numbers}
\end{table}

\section{Gender Classification Error Analysis}

Gender classification results were computed for two ``black box'' commercial APIs (Amazon~\cite{amazon_api} and Microsoft (MS)~\cite{microsoft_api}) and one open-source classifier trained for this experiment.
The open-source classifier was trained using ResNet-50~\cite{resnet} modified as proposed in~\cite{arcface} and~\cite{se}. 
Weighted binary cross entropy was used as the loss function, where each gender was weighted according to number of samples. 
For training and validation, we used a collection of datasets: AAF~\cite{aaf}, AFAD~\cite{ordinal}, AgeDB~\cite{agedb}, CACD~\cite{cacd}, IMDB-WIKI~\cite{imdb_wiki}, IMFDB~\cite{imfdb}, MegaAgeAsian~\cite{megaageasian}, and UTKFace~\cite{utkface}.
The variety of datasets is meant to improve the generalizability of the trained model.
Training and validation sets were created using a 90/10 split; training contains 466,256 images and validation contains 51,803 images.
The trained model achieves an accuracy of 96.25\% on the validation set.
To improve transparency and reproducibility of our results, the code and pre-trained model will be publicly released.

\begin{table}
\centering
   \begin{tabular}{|l|r|r|r|r|}
    \hline
 Classifier & D F & D M & L F & L M \\
 \hline
    Microsoft & 95.9 & 93.4 & 99.3 & 100.0 \\ \hline
    Amazon  & 96.7 & 99.4 & 99 & 99.7 \\ \hline
    Open source & 80.4 & 98.7 & 97.3 & 99.7 \\ \hline
    \end{tabular}
    \caption{Gender classification accuracy on PPB~\cite{gender_shades,Raji_2019}.}
    \label{ppb_accuracy}
\end{table}
\begin{table}
\centering
\begin{tabular}{|l|r|r|r|r|}
\hline
Classifier  & A-A M  &  C M  & A-A F & C F \\
\hline
Microsoft   & 99.2   &  99.8 & 96.3 & 99.0 \\
\hline
Amazon      & 98.1   &  99.5  & 92.9 & 97.9  \\
\hline
Open source & 98.4   &  99.3 & 82.8 & 92.2  \\
\hline
\end{tabular}
\caption{
Gender classification accuracy on MORPH. 
Accuracy is higher for males than females,
higher for Caucasians than African-Americans 
(as in \cite{gender_shades,Muthukumar_2018}).
The MS API gives no classification for a small number of images: 1 C F, 4 A-A F, 32 C M, and 35 A-A M; these results are counted as classification errors in this table.
}
\label{table:gender_pred_morph}
\end{table}

\begin{table}
\centering
\begin{tabular}{|l|r|r|r|r|}
\hline
Classifier  & A-A M  &  C M  & A-A F & C F \\
\hline
Microsoft   & 440   &  73 & 930 & 112 \\
\hline
Amazon      & 1,053   &  171  & 1,759 & 231  \\
\hline
Open source & 913   &  258 & 4,278 & 853  \\
\hline \hline
In common      & 169   &  15   & 540  & 68  \\
\% Microsoft   & 38\%  & 21\%  & 58\% & 61\% \\
\% Amazon      & 16\%  & 9\%  & 31\% & 29\% \\
\% Open        & 19\%  & 6\%  & 13\% & 8\% \\
\hline
\end{tabular}
\caption{
Number of errors individually and in common across classifiers.
Each classifier has a substantial fraction of its errors that are not errors for the other classifiers.
}
\label{table:common_errors}
\end{table}

For cross-reference with Buolamwini and Gebru~\cite{gender_shades}, we benchmark the accuracy of the gender classifiers used in this paper on the PPB dataset.
Results are shown in Table~\ref{ppb_accuracy}.
The two commercial gender classification APIs in our study have similar relative accuracy between demographics as found
in~\cite{gender_shades}, but generally higher accuracy overall.
In particular, the accuracy for 
darker female (D F) images is much higher than reported in~\cite{gender_shades}.
However, following the publicity generated by
\cite{gender_shades},
and given the availability of the PPB images, it is possible that PPB has effectively been used in training current commercial APIs.
The accuracy pattern for our gender classifier trained for use in this work, and not trained using the PPB dataset, more closely follows the accuracy pattern in~\cite{gender_shades}.

The gender classification accuracy for MORPH is shown in Table~\ref{table:gender_pred_morph}.
The accuracy pattern across demographics in these results largely follows that reported in~\cite{gender_shades} for the PPB dataset.
For all three gender classification algorithms, accuracy is generally higher for male than for female, higher for Caucasian than for African-American, and lowest for African-American female.
Also, the accuracy disparity for the open source algorithm is similar to that reported in~\cite{gender_shades} for the three commercial APIs evaluated at that time.
Overall, the MS API gives the highest accuracy on  MORPH.

\begin{figure}[t]
    \centering
       \begin{subfigure}[t]{0.1\textwidth}
            \includegraphics[width=\textwidth]{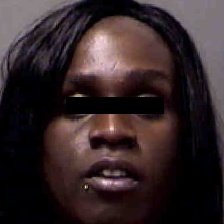}
     \caption{A-A M}
   \end{subfigure}
   \quad
   \begin{subfigure}[t]{0.1\textwidth}
     \includegraphics[width=\textwidth]{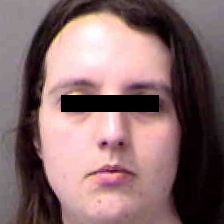}
     \caption{C M}
   \end{subfigure}
   \quad
   \begin{subfigure}[t]{0.1\textwidth}
     \includegraphics[width=\textwidth]{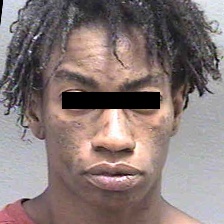}
     \caption{A-A F}
   \end{subfigure}
   \quad
   \begin{subfigure}[t]{0.1\textwidth}
     \includegraphics[width=\textwidth]{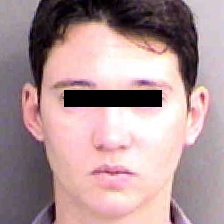}
     \caption{C F}
   \end{subfigure}
   \caption{Example images that generate an error for all three gender classification algorithms.}
   \label{fig:error_samples}
\end{figure}

As indicated in Table~\ref{table:common_errors}, The different gender classification algorithms make errors on some of the same images, but there are also substantial differences between algorithms.
For example, for African-American male, 
only 169 of the 440 images for which the MS API makes a classification error also result in an error for both the Amazon API and the open-source API, even though those APIs make many more errors.
This reinforces that we have three distinct gender classification algorithms.
Each of these three exhibits the general trends of different accuracy across demographic groups that have been reported by previous researchers.
Example images classified incorrectly (according to the metadata records) by all three gender classification algorithms are shown in Figure \ref{fig:error_samples}. It is noted that the atypical length of hair and the image exposure are possible underlying factors for these example images.

Note that all example face images shown in this paper have black rectangles added over the eye regions for increased anonymity / privacy of the individuals.  The original images in MORPH of course do not have the rectangles, and were extracted from public records. 

\section{Face Recognition Error Analysis}

We present results for a pre-trained version 
(available at~\cite{insightface})
of the well-known ArcFace~\cite{arcface} matcher, 
(available here~\cite{insightface})
and also for a current commercial off-the-shelf face recognition API, referred to only as COTS due to license restrictions.
Face detection and alignment is performed on each image prior to the open-source side of the experiments using either img2pose~\cite{albiero2020img2pose} or RetinaFace~\cite{retinaface}.
For the ArcFace matcher, 512-d features are extracted from aligned faces and then matched using cosine distance.
For the commercial matcher, full images are input to the COTS and genuine and impostor distribution are returned.

We first analyze whether and how the impostor distribution varies for images that generate gender classification error.
Then we consider how the genuine distribution varies.
Space limits prevent showing impostor and genuine distribution plots for all six combinations of gender classifier and face matchers.
However, the general pattern of results is similar across all six and is summarized in Table 1 of the supplemental material.
Impostor and genuine distributions are shown for the combination of the highest-accuracy gender-from-face API (MS) and the higher-accuracy face matcher (ArcFace).

\begin{table}
\centering
\begin{tabular}{|l|r|r|r|r|r|}
\hline
Cohort &  & incorrect: & correct: & correct: \\
       &  & incorrect  & incorrect & correct \\
\hline
\multirow{2}{*}{A-A M}  & $\square$ &95,695 &24,551,490 &1,556,779,131  \\
                        \cline{2-5}
                        & $\blacksquare$ &0.04905  &-0.00185 &0.00002 \\
\hline
\multirow{2}{*}{C M}    & $\square$ &2,526  &2,568,785 &619,471,387  \\
                        \cline{2-5}
                        & $\blacksquare$ &0.01642   &-0.00557 &0.00003 \\
\hline
\multirow{2}{*}{A-A F } & $\square$ &428,817  &22,246,621 &286,164,751   \\
                        \cline{2-5}
                        & $\blacksquare$ &0.01638  &-0.01114 &0.00084 \\                        
\hline
\multirow{2}{*}{C F}    & $\square$ &6,130 &1,212,224 &58,595,171  \\
                        \cline{2-5}
                        & $\blacksquare$ &0.01150  &-0.00670 &0.00010\\
\hline
\end{tabular}
\caption{Number of image pairs ($\square$) and the relative average impostor scores difference from overall impostor distribution ($\blacksquare$) for ArcFace and MS classifier. The complete set of results by the two matchers and three gender classifiers is available in the supplemental material.
}
\label{table:image_pairs_of_impostor_distribution}
\end{table}

\begin{figure*}
  \centering
  \begin{subfigure}[b]{1\linewidth}
      \centering
      \begin{subfigure}[b]{0.245\linewidth}
        \centering
          \includegraphics[width=\linewidth, height=3.2cm]{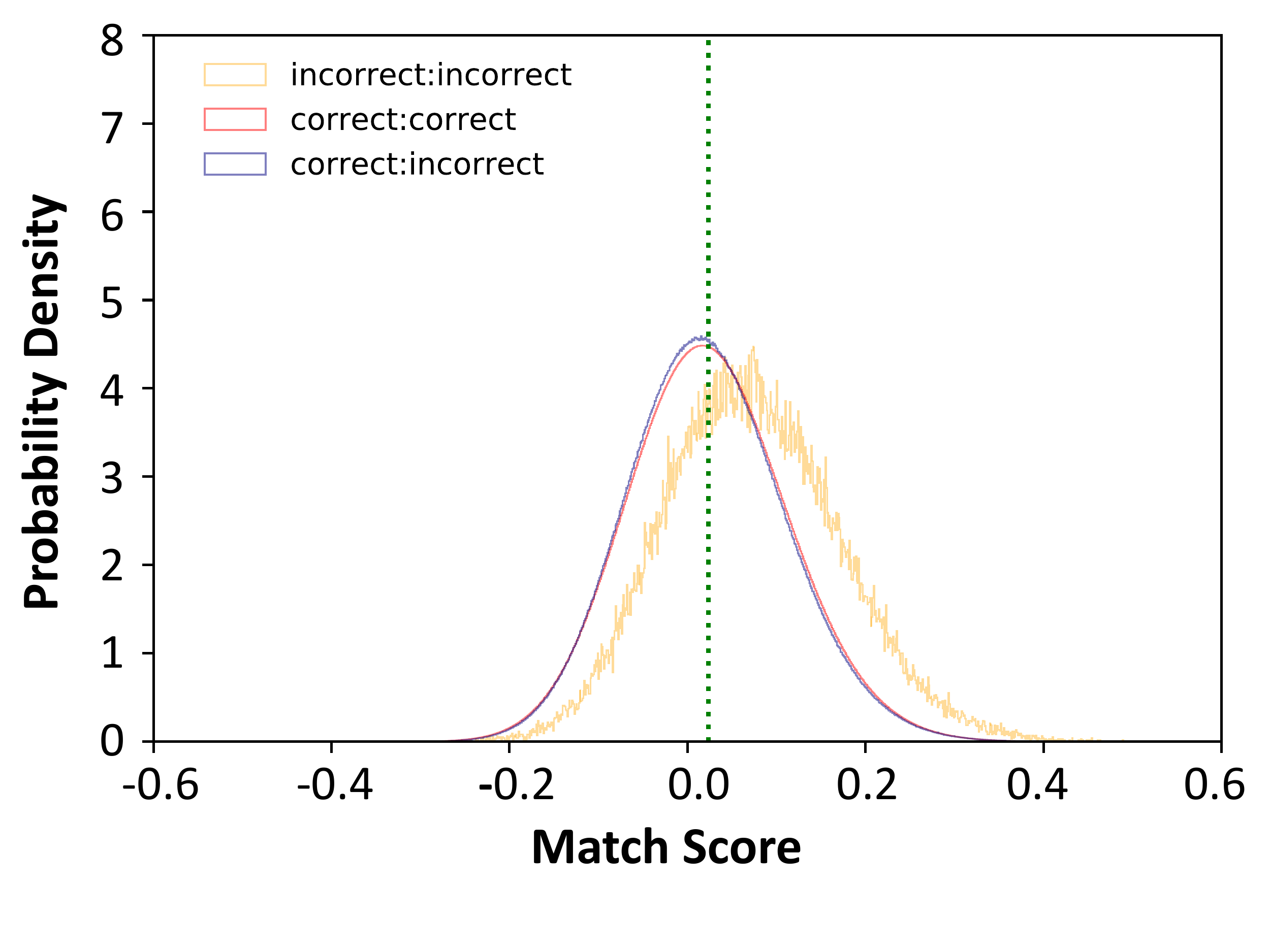}
          \caption{African-American Males}
      \end{subfigure}
      \begin{subfigure}[b]{0.245\linewidth}
        \centering
          \includegraphics[width=\linewidth]{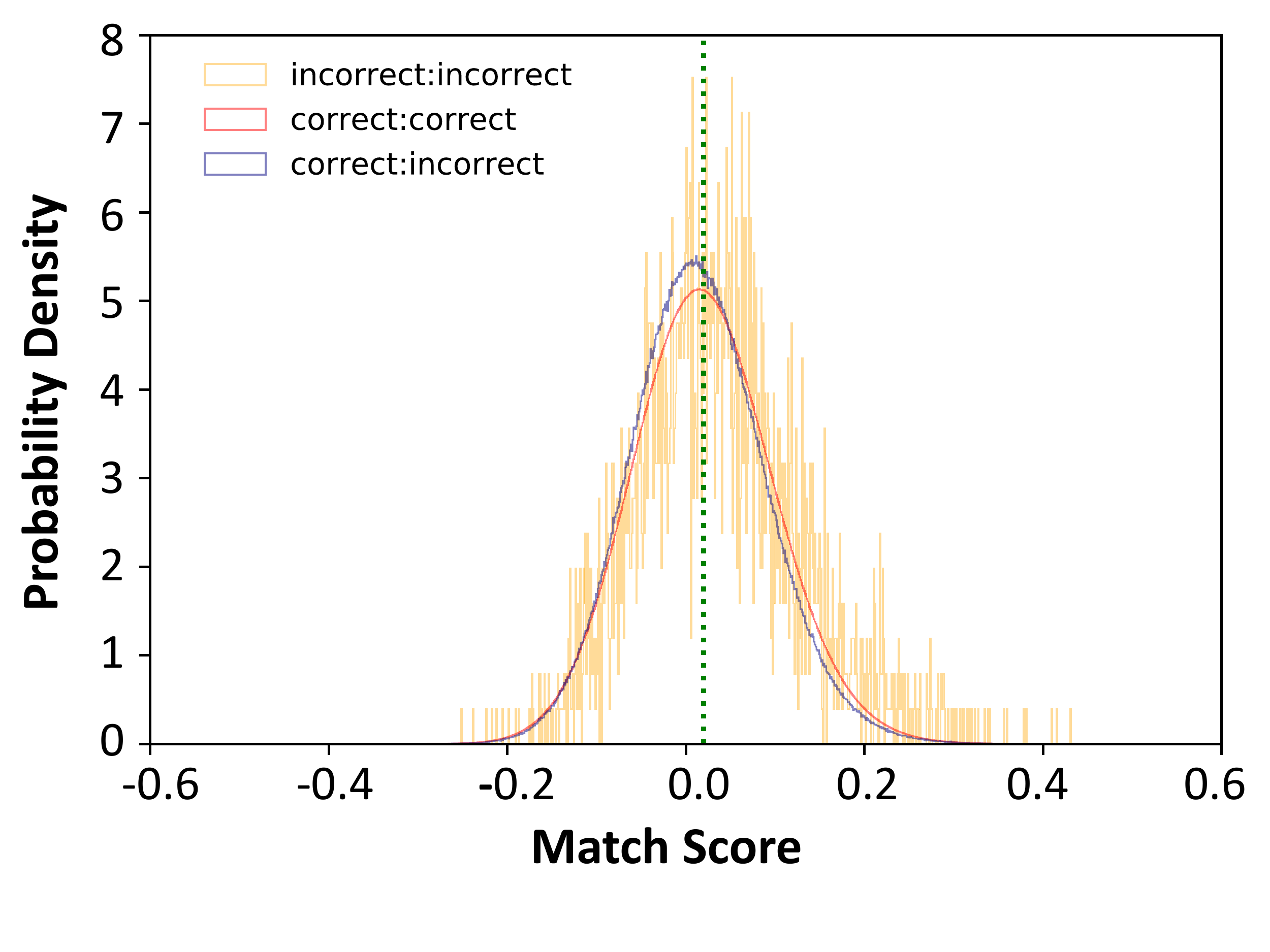}
          \caption{Caucasian Males}
      \end{subfigure}
      \centering
      \begin{subfigure}[b]{0.245\linewidth}
        \centering
          \includegraphics[width=\linewidth]{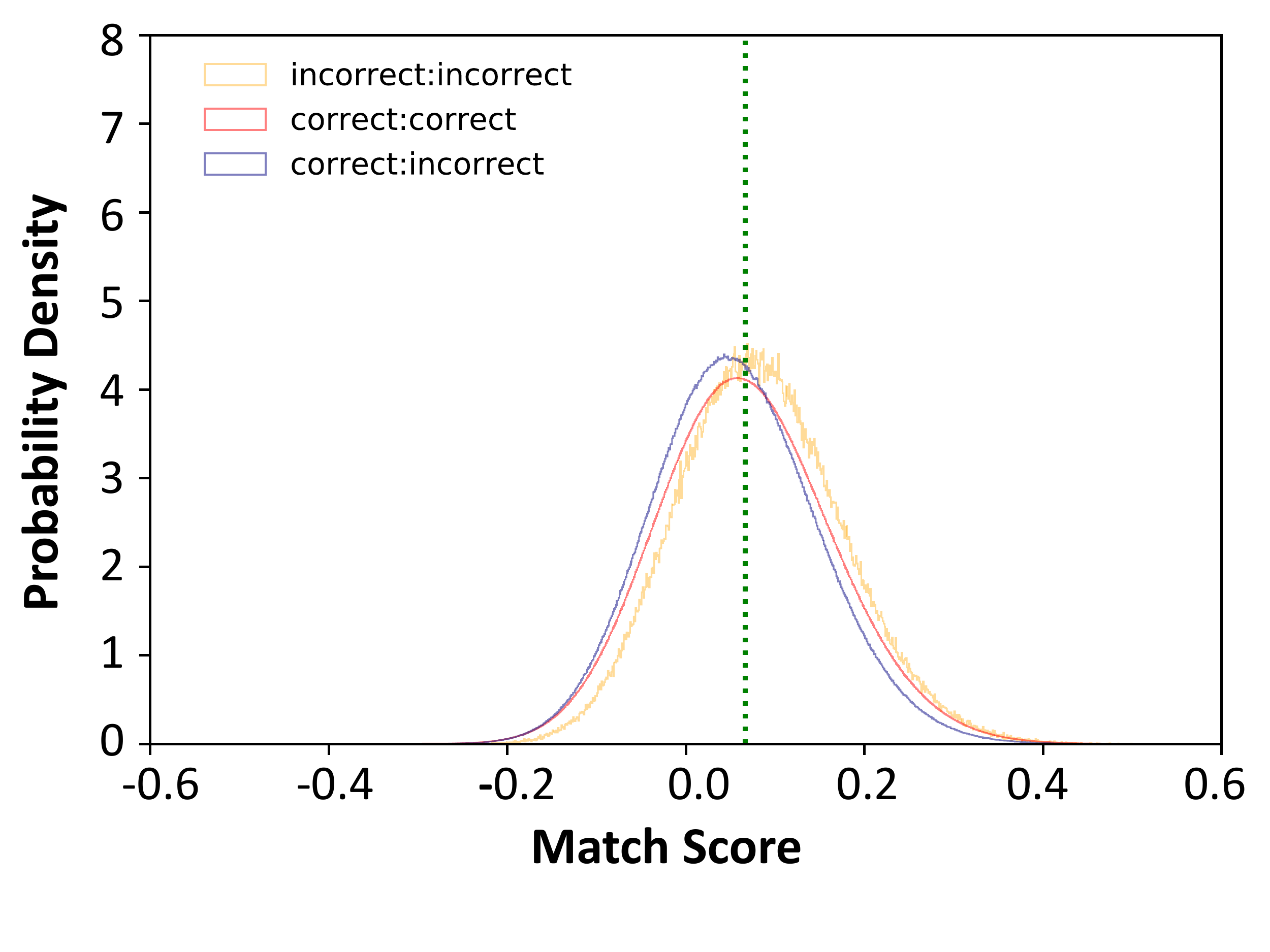}
          \caption{African-American Females}
      \end{subfigure}
      \begin{subfigure}[b]{0.245\linewidth}
        \centering
          \includegraphics[width=\linewidth]{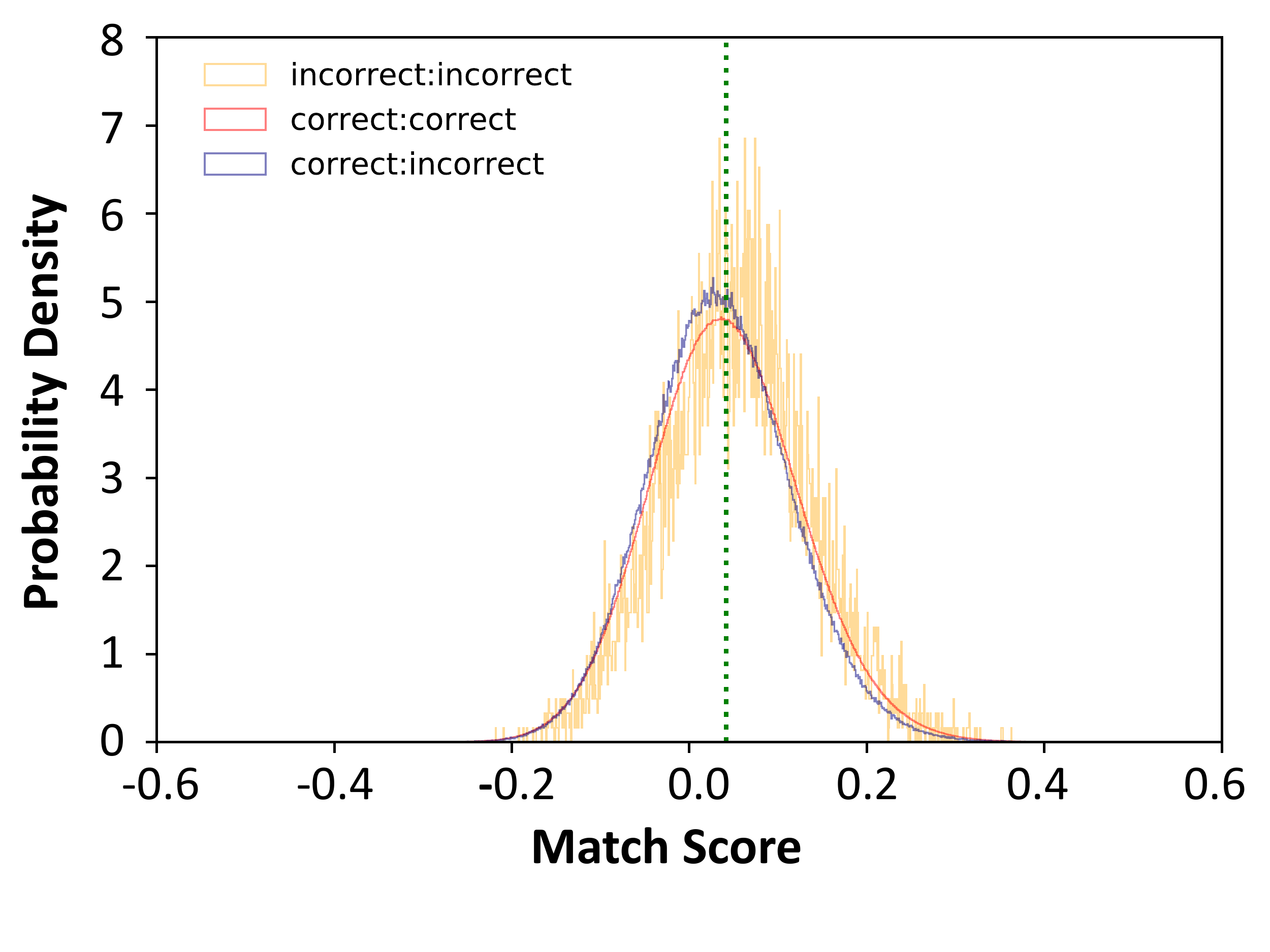}
          \caption{Caucasian Females}
      \end{subfigure}
  \end{subfigure}
  \caption{ArcFace impostor distribution split into images without gender error, with at least one with gender error, and with both with gender error. Gender predicted using Microsoft Face API. The complete set of impostor distribution by the two matchers and three gender classifiers is available in the supplemental material.}
  \label{fig:imp_arcface_ms}
  \vspace{-0.5em}
\end{figure*}
\begin{figure*}[t]
  \begin{subfigure}[b]{1\linewidth}
      \begin{subfigure}[b]{0.24\linewidth}
          \begin{subfigure}[b]{.49\columnwidth}
            \centering
            \includegraphics[width=\linewidth]{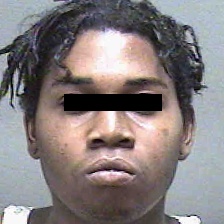}
          \end{subfigure}
          \hfill 
          \begin{subfigure}[b]{.49\columnwidth}
            \centering
            \includegraphics[width=\linewidth]{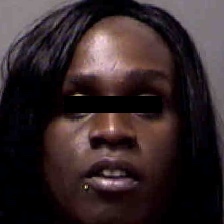}
          \end{subfigure}
          \vspace{-0.5em}
      \end{subfigure}
      \hfill 
      \begin{subfigure}[b]{0.24\linewidth}
          \begin{subfigure}[b]{.49\columnwidth}
            \centering
            \includegraphics[width=\linewidth]{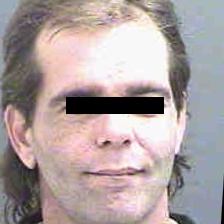}
          \end{subfigure}
          \hfill 
          \begin{subfigure}[b]{.49\columnwidth}
            \centering
            \includegraphics[width=\linewidth]{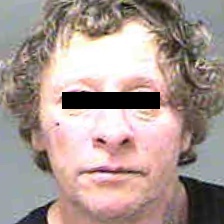}
          \end{subfigure}
          \vspace{-0.5em}
      \end{subfigure}
      \hfill 
      \begin{subfigure}[b]{0.24\linewidth}
          \begin{subfigure}[b]{.49\columnwidth}
            \centering
            \includegraphics[width=\linewidth]{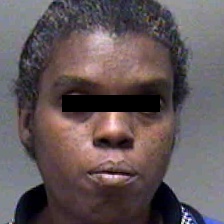}
          \end{subfigure}
          \hfill 
          \begin{subfigure}[b]{.49\columnwidth}
            \centering
            \includegraphics[width=\linewidth]{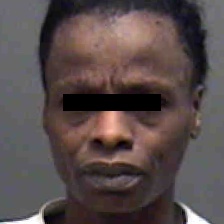}
          \end{subfigure}
          \vspace{-0.5em}
      \end{subfigure}
      \hfill 
      \begin{subfigure}[b]{0.24\linewidth}
          \begin{subfigure}[b]{.49\columnwidth}
            \centering
            \includegraphics[width=\linewidth]{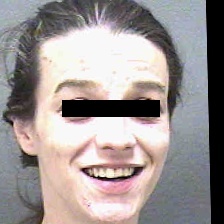}
          \end{subfigure}
          \hfill 
          \begin{subfigure}[b]{.49\columnwidth}
            \centering
            \includegraphics[width=\linewidth]{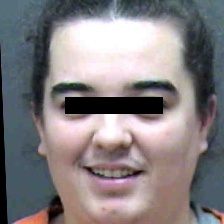}
          \end{subfigure}
          \vspace{-0.5em}
      \end{subfigure}
  \end{subfigure}
  \begin{subfigure}[b]{1\linewidth}
      \begin{subfigure}[b]{0.24\linewidth}
          \begin{subfigure}[b]{.49\columnwidth}
            \centering
            \includegraphics[width=\linewidth]{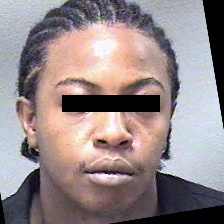}
          \end{subfigure}
          \hfill 
          \begin{subfigure}[b]{.49\columnwidth}
            \centering
            \includegraphics[width=\linewidth]{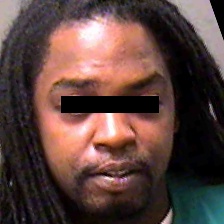}
          \end{subfigure}
          \vspace{-0.5em}
      \end{subfigure}
      \hfill 
      \begin{subfigure}[b]{0.24\linewidth}
          \begin{subfigure}[b]{.49\columnwidth}
            \centering
            \includegraphics[width=\linewidth]{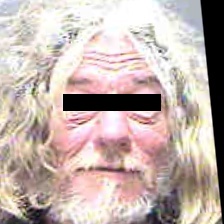}
          \end{subfigure}
          \hfill 
          \begin{subfigure}[b]{.49\columnwidth}
            \centering
            \includegraphics[width=\linewidth]{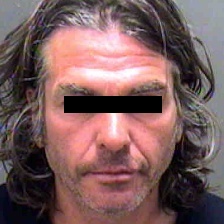}
          \end{subfigure}
          \vspace{-0.5em}
      \end{subfigure}
      \hfill 
      \begin{subfigure}[b]{0.24\linewidth}
          \begin{subfigure}[b]{.49\columnwidth}
            \centering
            \includegraphics[width=\linewidth]{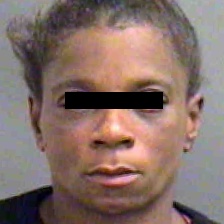}
          \end{subfigure}
          \hfill 
          \begin{subfigure}[b]{.49\columnwidth}
            \centering
            \includegraphics[width=\linewidth]{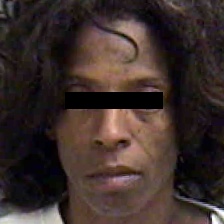}
          \end{subfigure}
          \vspace{-0.5em}
      \end{subfigure}
      \hfill 
      \begin{subfigure}[b]{0.24\linewidth}
          \begin{subfigure}[b]{.49\columnwidth}
            \centering
            \includegraphics[width=\linewidth]{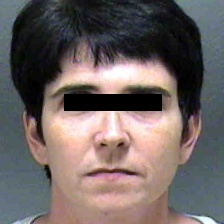}
          \end{subfigure}
          \hfill 
          \begin{subfigure}[b]{.49\columnwidth}
            \centering
            \includegraphics[width=\linewidth]{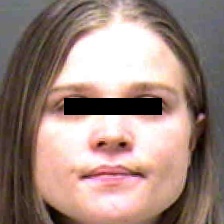}
          \end{subfigure}
          \vspace{-0.5em}
      \end{subfigure}
  \end{subfigure}
  \begin{subfigure}[b]{1\linewidth}
      \begin{subfigure}[b]{0.24\linewidth}
          \begin{subfigure}[b]{.49\columnwidth}
            \centering
            \includegraphics[width=\linewidth]{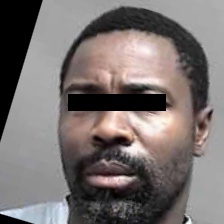}
          \end{subfigure}
          \hfill 
          \begin{subfigure}[b]{.49\columnwidth}
            \centering
            \includegraphics[width=\linewidth]{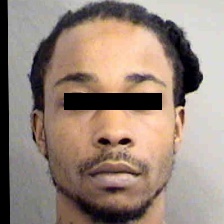}
          \end{subfigure}
          \caption{A-A M}
          \vspace{-0.5em}
      \end{subfigure}
      \hfill 
      \begin{subfigure}[b]{0.24\linewidth}
          \begin{subfigure}[b]{.49\columnwidth}
            \centering
            \includegraphics[width=\linewidth]{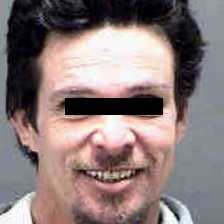}
          \end{subfigure}
          \hfill 
          \begin{subfigure}[b]{.49\columnwidth}
            \centering
            \includegraphics[width=\linewidth]{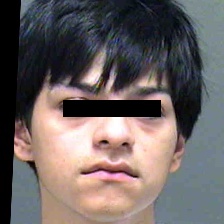}
          \end{subfigure}
          \caption{C M}
          \vspace{-0.5em}
      \end{subfigure}
      \hfill 
      \begin{subfigure}[b]{0.24\linewidth}
          \begin{subfigure}[b]{.49\columnwidth}
            \centering
            \includegraphics[width=\linewidth]{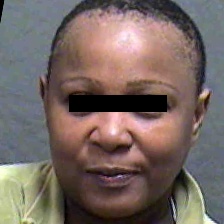}
          \end{subfigure}
          \hfill 
          \begin{subfigure}[b]{.49\columnwidth}
            \centering
            \includegraphics[width=\linewidth]{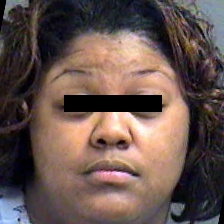}
          \end{subfigure}
          \caption{A-A F}
          \vspace{-0.5em}
      \end{subfigure}
      \hfill 
      \begin{subfigure}[b]{0.24\linewidth}
          \begin{subfigure}[b]{.49\columnwidth}
            \centering
            \includegraphics[width=\linewidth]{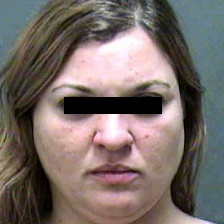}
          \end{subfigure}
          \hfill 
          \begin{subfigure}[b]{.49\columnwidth}
            \centering
            \includegraphics[width=\linewidth]{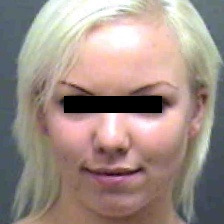}
          \end{subfigure}
          \caption{C F}
          \vspace{-0.5em}
      \end{subfigure}
  \end{subfigure}
  \caption{Example of image pairs from component distributions: type III (the first row), type II (the second row and the first image of each image pair has gender classification error), type I (the third row) with ArcFace and Microsoft gender classifier.}
  \label{fig:example_image_pairs_arcface_microsoft}
\end{figure*}

\subsection{Impostor Distribution Analysis}
The impostor distribution is the set of similarity scores for all impostor (``non-mated'') image pairs.
In the context of gender classification errors, there are three possible types of impostor image pairs:
\begin{itemize}
    \item Type I: (correct:correct), both images have correct gender classification. 
    \item Type II: (incorrect:correct), one  has incorrect gender classification and one has correct classification.
    \item Type III: (incorrect:incorrect), both images have incorrect classification.
\end{itemize}
The number of image pairs in the three component distributions is given in 
Table~\ref{table:image_pairs_of_impostor_distribution}.
This observation allows use to subdivide the overall impostor distribution into three component impostor distributions.
If the gender classification result of an image has no impact on false match errors, then there should be no significant difference between the three component distributions.
If an image resulting in gender classification error makes it more likely to be involved in false match errors, 
then the type II and III distributions should be shifted to higher similarity values than the type I distribution.

Figure \ref{fig:imp_arcface_ms} shows the component impostor distributions obtained using the MS gender-from-face API with the similarity scores obtained from matching ArcFace features.
The dashed vertical line marks the mean similarity score for the overall impostor distribution (the three component distributions taken together).
Note that the qualitative positioning of the three component distributions is the same across all four demographics.
The type II component distribution is centered at the lowest similarity value.
Thus the (incorrect:correct) image pairs should then have the 
lowest false match rate (FMR). 
A possible interpretation is that because the images in a (incorrect:correct) pair have been classified as having different gender appearance, they on average have a lower similarity score.
Type I and III component distributions are composed of image pairs where both images have the same gender classification. Both component distributions are shifted toward higher similarity scores. However, the type III distribution has higher similarity scores than type I according to Table~\ref{table:image_pairs_of_impostor_distribution}, meaning it would have the highest FMR of the three. 

It is also clear for all four demographic cohorts that the component distribution for incorrect:incorrect image pairs appears quite noisy.
This is because this component distribution is composed of a relatively small number of image pairs. 
The more accurate the gender classifier, the fewer images there are to form the incorrect:correct and incorrect:incorrect component distributions.
The number of pairs in each component distribution for the MS + ArcFace results are given in  Table~\ref{table:image_pairs_of_impostor_distribution}.
The number of (incorrect:incorrect) image pairs is less than 1\% of the number of (incorrect:correct) pairs for A-A M, less than 0.1\% for C M, less than 2\% for A-A F, and less than 1\% for C F.
Thus, for a given image with incorrect gender classification, the number of pairings that result in an increased chance of FM is a small fraction of the number that result in a decreased chance of FM.

\begin{figure*}[t]
  \begin{subfigure}[b]{1\linewidth}
      \begin{subfigure}[b]{0.24\linewidth}
          \begin{subfigure}[b]{.49\columnwidth}
            \centering
            \includegraphics[width=\linewidth]{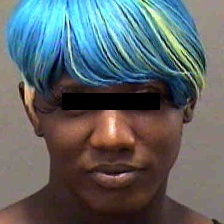}
          \end{subfigure}
          \hfill 
          \begin{subfigure}[b]{.49\columnwidth}
            \centering
            \includegraphics[width=\linewidth]{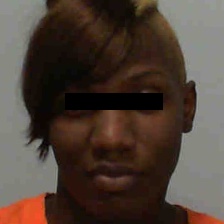}
          \end{subfigure}
          \vspace{-0.5em}
      \end{subfigure}
      \hfill 
      \begin{subfigure}[b]{0.24\linewidth}
          \begin{subfigure}[b]{.49\columnwidth}
            \centering
            \includegraphics[width=\linewidth]{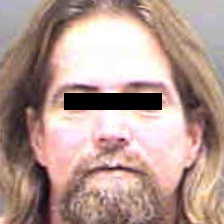}
          \end{subfigure}
          \hfill 
          \begin{subfigure}[b]{.49\columnwidth}
            \centering
            \includegraphics[width=\linewidth]{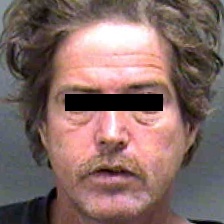}
          \end{subfigure}
          \vspace{-0.5em}
      \end{subfigure}
      \hfill 
      \begin{subfigure}[b]{0.24\linewidth}
          \begin{subfigure}[b]{.49\columnwidth}
            \centering
            \includegraphics[width=\linewidth]{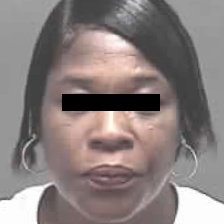}
          \end{subfigure}
          \hfill 
          \begin{subfigure}[b]{.49\columnwidth}
            \centering
            \includegraphics[width=\linewidth]{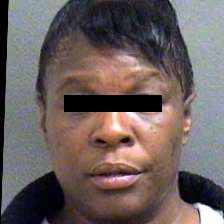}
          \end{subfigure}
          \vspace{-0.5em}
      \end{subfigure}
      \hfill 
      \begin{subfigure}[b]{0.24\linewidth}
          \begin{subfigure}[b]{.49\columnwidth}
            \centering
            \includegraphics[width=\linewidth]{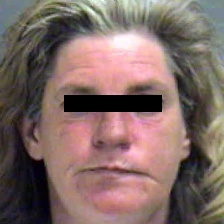}
          \end{subfigure}
          \hfill 
          \begin{subfigure}[b]{.49\columnwidth}
            \centering
            \includegraphics[width=\linewidth]{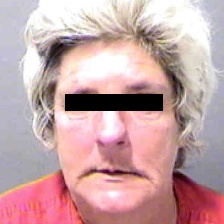}
          \end{subfigure}
          \vspace{-0.5em}
      \end{subfigure}
  \end{subfigure}
  \begin{subfigure}[b]{1\linewidth}
      \begin{subfigure}[b]{0.24\linewidth}
          \begin{subfigure}[b]{.49\columnwidth}
            \centering
            \includegraphics[width=\linewidth]{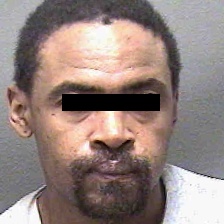}
          \end{subfigure}
          \hfill 
          \begin{subfigure}[b]{.49\columnwidth}
            \centering
            \includegraphics[width=\linewidth]{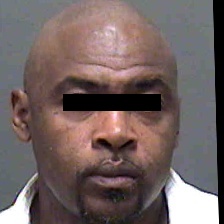}
          \end{subfigure}
          \caption{A-A M}
          \vspace{-0.5em}
      \end{subfigure}
      \hfill 
      \begin{subfigure}[b]{0.24\linewidth}
          \begin{subfigure}[b]{.49\columnwidth}
            \centering
            \includegraphics[width=\linewidth]{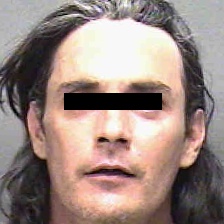}
          \end{subfigure}
          \hfill 
          \begin{subfigure}[b]{.49\columnwidth}
            \centering
            \includegraphics[width=\linewidth]{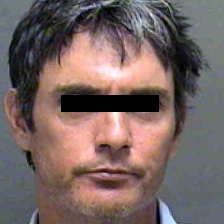}
          \end{subfigure}
          \caption{C M}
          \vspace{-0.5em}
      \end{subfigure}
      \hfill 
      \begin{subfigure}[b]{0.24\linewidth}
          \begin{subfigure}[b]{.49\columnwidth}
            \centering
            \includegraphics[width=\linewidth]{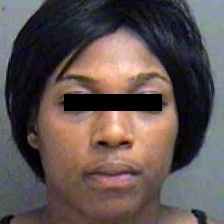}
          \end{subfigure}
          \hfill 
          \begin{subfigure}[b]{.49\columnwidth}
            \centering
            \includegraphics[width=\linewidth]{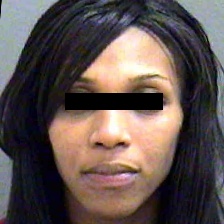}
          \end{subfigure}
          \caption{A-A F}
          \vspace{-0.5em}
      \end{subfigure}
      \hfill 
      \begin{subfigure}[b]{0.24\linewidth}
          \begin{subfigure}[b]{.49\columnwidth}
            \centering
            \includegraphics[width=\linewidth]{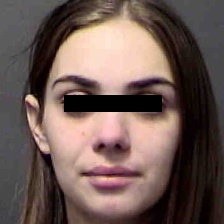}
          \end{subfigure}
          \hfill 
          \begin{subfigure}[b]{.49\columnwidth}
            \centering
            \includegraphics[width=\linewidth]{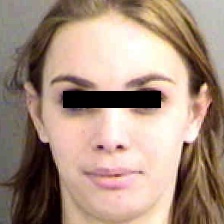}
          \end{subfigure}
          \caption{C F}
          \vspace{-0.5em}
      \end{subfigure}
  \end{subfigure}
  \caption{Example of pairs from authentic distribution composed by subjects that have at least one image with gender classification error (top), and subjects without any gender classification error (bottom). The gender error that put the subjects from the top into this distribution could occur on a third image not shown here, as is the case for the C M subject. Samples picked from the each distribution 10th percentile.}
  \label{fig:auth_samples}
  \vspace{-1em}
\end{figure*}

There are small differences between the four demographic cohorts in the relative positioning of the component distributions.
The type I and II distributions appear to be less separated for the African-American male cohort than for the other three cohorts.
Also, the type III distribution for African-American male appears to be more separated from the other components than for the other cohorts.

Example image pairs from the 10\% highest similarity scores of the component distributions of the four cohorts are shown in Figure \ref{fig:example_image_pairs_arcface_microsoft}. (incorrect:incorrect) and (correct:correct) image pairs are in the first and third rows of Figure \ref{fig:example_image_pairs_arcface_microsoft}, respectively. In these image pairs, both images are either classified as male or female by MS API, consequently the two faces seem subjectively more similar in terms of gender appearance as compared with (incorrect:correct) image pairs in the second row.

\begin{table}[t]
\small
\centering
\setlength\tabcolsep{3.5pt}
\begin{tabular}{|l|r|r|r|r|r|r|r|}
\hline
\multicolumn{1}{|c|}{\multirow{2}{*}{Cohort}} & \multicolumn{1}{l|}{\multirow{2}{*}{Total}} & \multicolumn{3}{c|}{Gender Error $>$ 0.5} & \multicolumn{3}{c|}{Gender Error $=$ 1} \\ \cline{3-8} 
\multicolumn{1}{|c|}{} & \multicolumn{1}{l|}{} & \multicolumn{1}{c|}{Open} & \multicolumn{1}{c|}{MS} & \multicolumn{1}{c|}{Amazon} & \multicolumn{1}{c|}{Open} & \multicolumn{1}{c|}{MS} & \multicolumn{1}{c|}{Amazon} \\ \hline
A-A M & 8,839 & 94 & 58 & 120 & 13 & 10 & 26 \\ \hline
C M & 8,835 & 68 & 16 & 48 & 8 & 6 & 10 \\ \hline
A-A F & 5,929 & 911 & 150 & 351 & 181 & 30 & 65 \\ \hline
C F & 2,798 & 194 & 25 & 50 & 45 & 5 & 8 \\ \hline
\end{tabular}
\caption{Number of subjects with half or more of their images with classification error (Gender Error $>$ 0.5), and number with all images with classification error (Gender Error $=$ 1).} 
\label{tab:error_0_5_1}
\end{table}

To check that the different numbers of images in the four demographic cohorts did not affect our results, we created a balanced subset of the data, in which the same number of subjects, with same number of images, and same age distribution, is randomly selected from MORPH for each cohort.
We performed a parallel analysis on this balanced dataset.
The results follow a qualitatively similar pattern as for the whole dataset. These results can be found in Figure 3 and Table 2 of the supplemental material.

\subsection{Genuine Distribution Analysis}

\begin{figure*}[t]
  \centering
  \begin{subfigure}[b]{1\linewidth}
      \centering
      \begin{subfigure}[b]{0.245\linewidth}
        \centering
          \includegraphics[width=\linewidth]{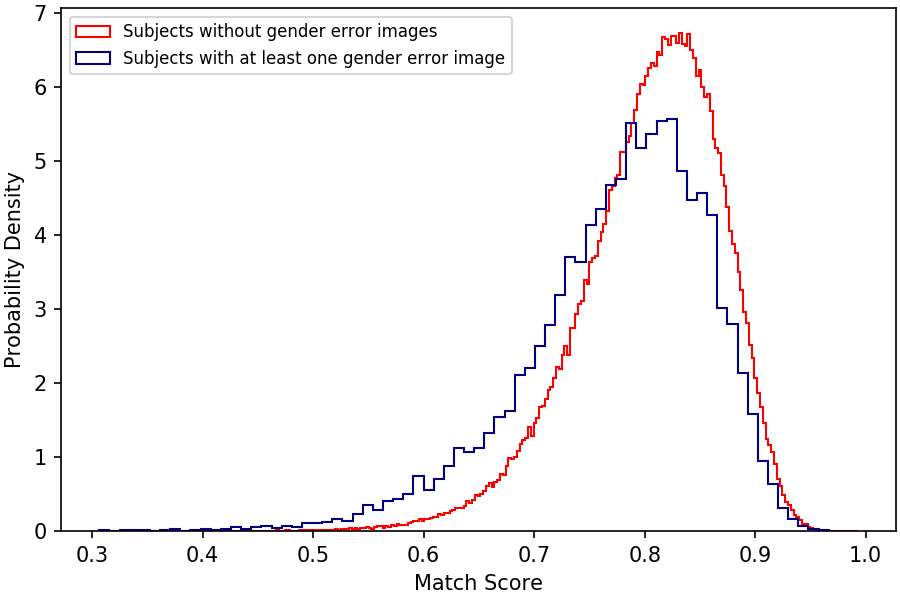}
          \caption{African-American Males}
      \end{subfigure}
      \centering
      \begin{subfigure}[b]{0.245\linewidth}
        \centering
          \includegraphics[width=\linewidth]{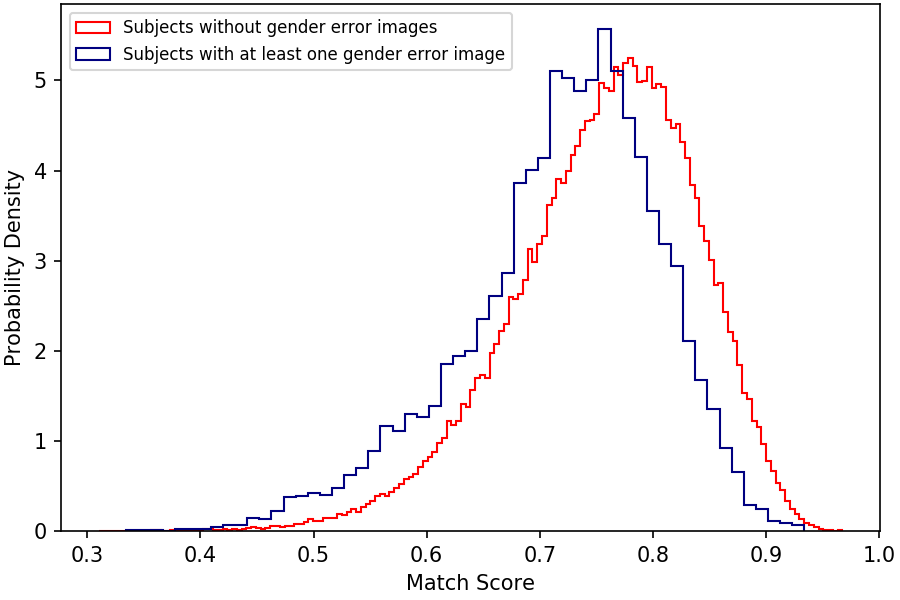}
          \caption{Caucasian Males}
      \end{subfigure}
    \centering
      \begin{subfigure}[b]{0.245\linewidth}
        \centering
          \includegraphics[width=\linewidth]{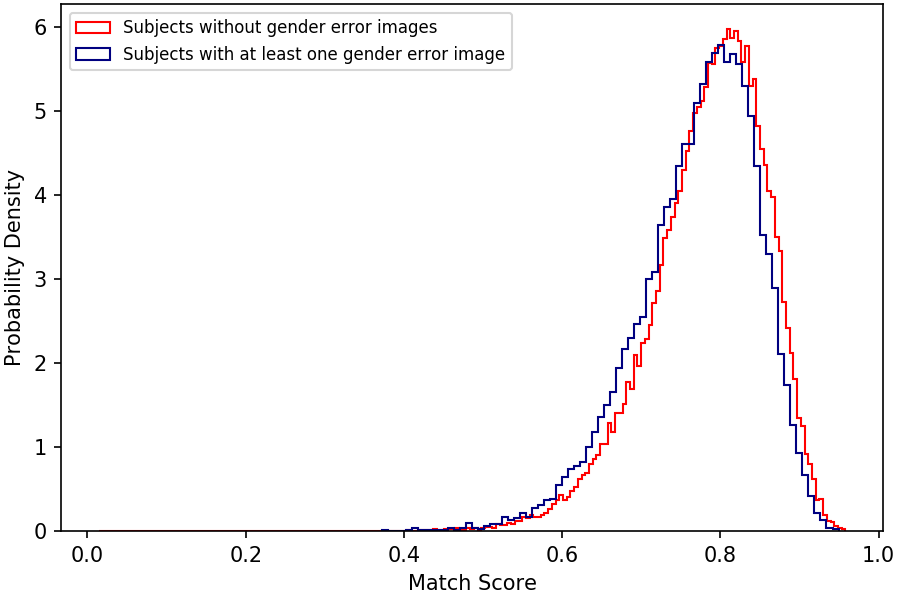}
          \caption{African-American Females}
      \end{subfigure}
      \begin{subfigure}[b]{0.245\linewidth}
        \centering
          \includegraphics[width=\linewidth]{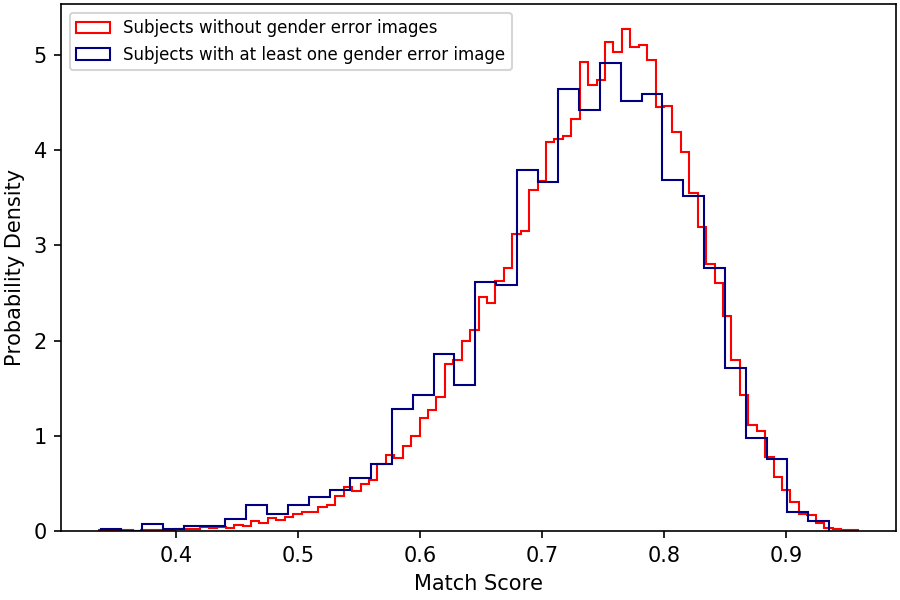}
          \caption{Caucasian Females}
      \end{subfigure}
  \end{subfigure}
  \caption{
  Arcface genuine distribution broken out by subjects with no image having gender classification error and subjects with at least one image with gender classification error (using MS gender classification API). The complete set of impostor distribution by the two matchers and three gender classifiers is available in the supplemental material.
  }
  \label{fig:auth_arcface_ms}
\end{figure*}

The genuine distribution contains similarity scores for pairs of images of the same person.
There is inherently less data in the genuine distribution than in the impostor distribution, and using the same definition of component distributions as used for the impostor distribution would not make sense.
We consider the genuine distribution split into two component distributions: (a) persons whose images all have correct gender classification, and (b) persons with one or more images that have incorrect gender classification.
Samples from these two distributions are shown in Figure~\ref{fig:auth_samples}.

As shown in Table~\ref{tab:error_0_5_1}, there are only a few people that have all their images with incorrect gender classification, which is too small of a sample to analyze in a meaningful way. (See Figure 1 in the supplemental material for a histogram of the percentage of images with error versus total images per subject.)
If images with incorrect gender classification have no effect on false non-match errors, then the two component distributions should be essentially the same.
If images with incorrect gender classification cause an increase in false non-match errors, 
then the component distribution for persons with mixed gender classification result should be shifted toward lower similarity scores.

The component genuine distributions for the four demographic cohorts, computed with the MS API and the ArcFace matcher, are shown in Figure~\ref{fig:auth_arcface_ms}.
It is immediately apparent that, for males,
the component genuine distribution for persons whose images have mixed gender classification results is shifted noticeably toward lower similarity scores. (This agrees with the difference between average genuine similarity scores of two component genuine distributions as indicated in Table~\ref{tab:mean_diff_auth}.)
The difference in the component distributions for females is not as readily apparent, but the component distributions are in fact statistically significantly different at the 0.05 level by a two sample KS test.

The mixed-result component distribution has a ``blocky'' appearance that is due to the relatively small number of image pairs in the distribution.
Therefore, conclusions regarding the difference for this demographic should be regarded as tentative.

\begin{table}[t]
\small
\centering
\setlength\tabcolsep{4pt}
\begin{tabular}{|l|r|r|r|r|r|r|}
\hline
\multicolumn{1}{|c|}{\multirow{2}{*}{Cohort}} & \multicolumn{3}{c|}{ArcFace} & \multicolumn{3}{c|}{COTS} \\ \cline{2-7} 
\multicolumn{1}{|c|}{} & \multicolumn{1}{c|}{Open} & \multicolumn{1}{c|}{MS} & \multicolumn{1}{c|}{Amazon} & \multicolumn{1}{c|}{Open} & \multicolumn{1}{c|}{MS} & \multicolumn{1}{c|}{Amazon} \\ \hline
A-A M & 0.014 & 0.03 & 0.016 & 0.013 & 0.026 & 0.015 \\ \hline
C M & 0.044 & 0.038 & 0.039 & 0.019 & 0.036 & 0.016 \\ \hline
A-A F & 0.011 & 0.013 & 0.012 & 0.003 & 0.005 & 0.002 \\ \hline
C F & 0.01 & 0.008 & 0.03 & 0.008 & 0.005 & 0.025 \\ \hline
\end{tabular}
\caption{Difference between the average genuine similarity scores for subjects with all correct gender classification compared to subjects with at least one image with gender classification error. For both matchers, all three gender classifiers, and all four demographics, the distribution for subjects with one or more images with classification error has lower average similarity score.}
\label{tab:mean_diff_auth}
\end{table}

\section{Conclusions and Discussion}

Media reports often fail to distinguish between face analytics and face recognition, and treat error patterns in face analytics as if they necessarily tell something about error patterns in face recognition.
We report results of an investigation into whether and how images that generate incorrect classification on a particular face analytic, gender classification, have a different error pattern when used for face recognition.

Based on results for four demographic cohorts of images,
three gender classification algorithms and two face recognition algorithms, we offer the following general conclusions.
\begin{itemize}
    \item 
    An image that generates an error for a gender classification algorithm is slightly less likely to participate in an (impostor) image pair that generates a false match face recognition error.    
    \item 
    An image that generates an error for a gender classification algorithm is more likely to participate in a (genuine) image pair that generates a false non-match face recognition error.
    \item 
Image pairs in which one or both images generate a gender classification error are typically only a small fraction (less than 2\%) of the impostor image pairs, and naturally the fraction becomes smaller as the gender classifier becomes more accurate.
\end{itemize}

{\small
\bibliographystyle{ieee}
\bibliography{main.bib}
}

-\vfill
\clearpage
\setcounter{page}{1}
\setcounter{figure}{0}
\setcounter{table}{0}

\onecolumn
\begin{center}
\Large
\textbf{Does Face Recognition Error Echo Gender Classification Error?
\\-- Supplemental Material --}
\end{center}

\begin{figure*}[h]
  \centering
  \begin{subfigure}[b]{1\linewidth}
      \centering
      \begin{subfigure}[b]{0.327\linewidth}
        \centering
          \includegraphics[width=\linewidth]{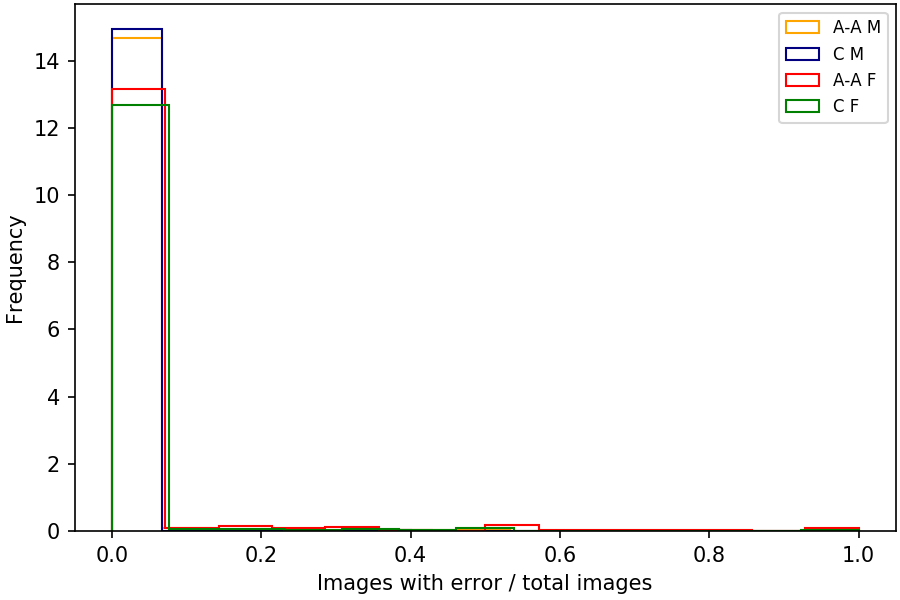}
          \caption{Microsoft Face API}
      \end{subfigure}
      \hfill 
      \begin{subfigure}[b]{0.327\linewidth}
        \centering
          \includegraphics[width=\linewidth]{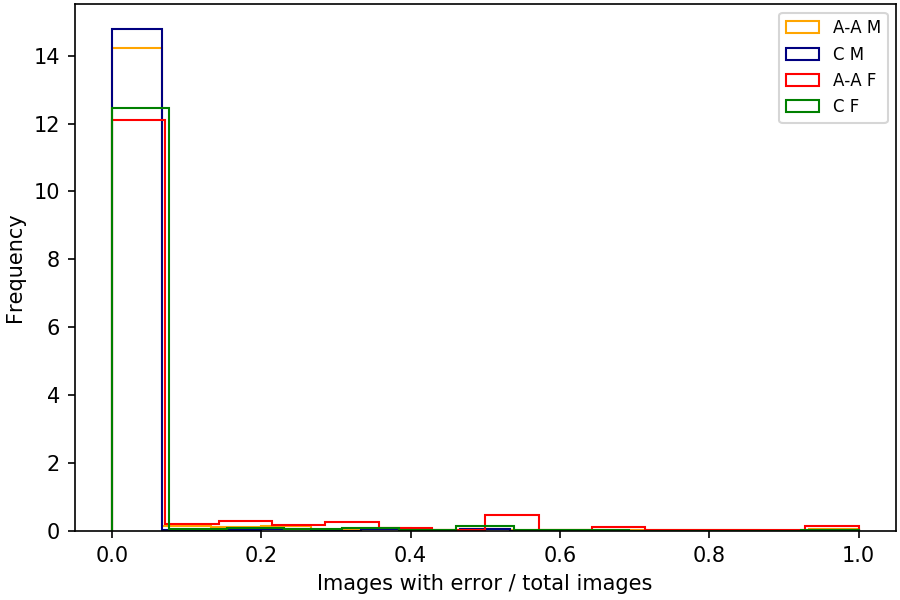}
          \caption{Amazon Face API}
      \end{subfigure}
      \hfill 
      \begin{subfigure}[b]{0.327\linewidth}
        \centering
          \includegraphics[width=\linewidth]{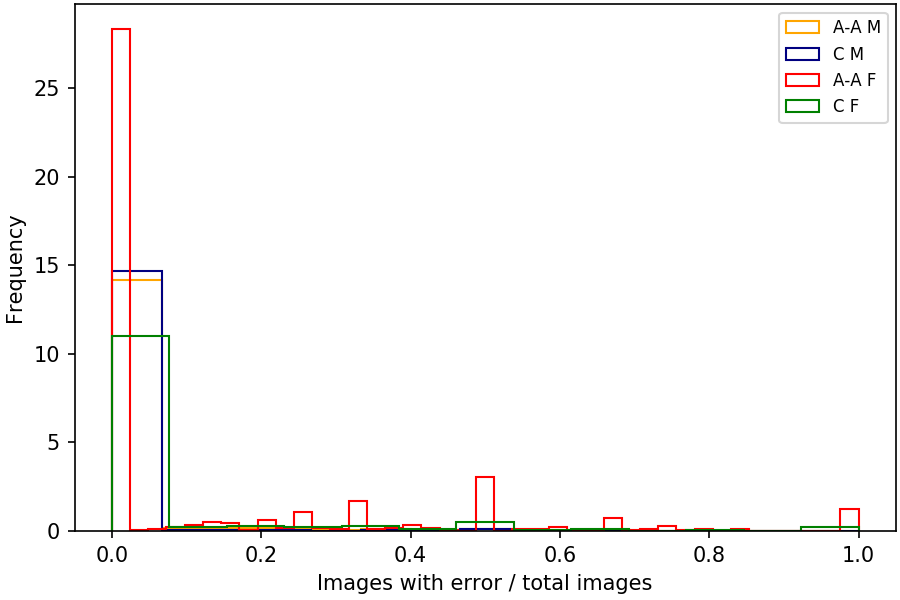}
          \caption{Open Source}
      \end{subfigure}
  \end{subfigure}
  \caption{Distribution of the number of images with gender error divided by the total number of images per subject.}
\end{figure*}
\begin{table*}[ht]
    \centering

        \begin{tabular}{|l|r|r|r|r|r|r|r|r|}
        \hline
\multirow{2}{*}{Gender Classifier} & \multicolumn{1}{|c|}{\multirow{2}{*}{Cohort}} & \multicolumn{1}{|c|}{\multirow{2}{*}{}} & \multicolumn{3}{c|}{ArcFace} & \multicolumn{3}{c|}{COTS} \\ \cline{4-9} 
& &\multicolumn{1}{|c|}{} & \multicolumn{1}{c|}{Two Errors} & \multicolumn{1}{c|}{One Error} & \multicolumn{1}{c|}{No Error} &\multicolumn{1}{c|}{Two Errors} & \multicolumn{1}{c|}{One Error} & \multicolumn{1}{c|}{No Error} 
\\ \hline
\multirow{8}{*}{Microsoft}  & \multirow{2}{*}{A-A M}  & $\square$ & 95,695  & 24,551,490 &  1,556,779,131 &95,257 &24,494,377 &1,556,611,750 \\
                        \cline{3-9}
                        &  & $\blacksquare$ & 0.04905  & -0.00185 & 0.00002 &0.03420 & -0.00590&0.00010\\
                        \cline{2-9}

                        & \multirow{2}{*}{C M}  & $\square$ & 2,526  & 2,568,785 & 619,471,387 &2,526 & 2,568,566&619,365,796 \\
                        \cline{3-9}
                        &  & $\blacksquare$ & 0.01642   & -0.00557 & 0.00003 &-0.04050 &-0.04330 &0.00020 \\
                       \cline{2-9}

                        & \multirow{2}{*}{A-A F }   & $\square$ & 428,817  & 22,246,621 &  286,164,751  &42,887 &22,244,762 &286,116,942 \\
                        \cline{3-9}
                        &   & $\blacksquare$ & 0.01638  & -0.01114& 0.00084 &-0.01100 &-0.02030 & 0.00160 \\ 
                        \cline{2-9}

                        & \multirow{2}{*}{C F}    & $\square$ & 6,130 & 1,212,224 &  58,595,171 &6,130 &1,212,112 &58,584,349 \\
                        \cline{3-9}
                        &   & $\blacksquare$ & 0.01150  & -0.00670 & 0.00010 &0.00600 &-0.01040 &0.00020 \\
                        \hline
\multirow{8}{*}{Amazon}  & \multirow{2}{*}{A-A M}  & $\square$ &552,158  &58,111,025 &1,522,763,133  &552,158 &58,106,813 &1,522,542,413 \\
                        \cline{3-9}
                        &  & $\blacksquare$ &0.5625   &-0.00088  &0.00001  &0.04800 &-0.00420 &0.00020\\
                        \cline{2-9}

                        & \multirow{2}{*}{C M}  & $\square$ &14,494 &6,002,026  &616,026,178 &14,494 &6,001,514 &615,920,880 \\
                        \cline{3-9}
                        &  & $\blacksquare$ &0.01871    &-0.00371  &0.00004  &0.01350 &-0.00710 & 0.00010 \\
                       \cline{2-9}

                        & \multirow{2}{*}{A-A F }   & $\square$ &1,541,983 &40,619,357  &266,678,849    &1,541,983 &40,615,839 &266,632,699 \\
                        \cline{3-9}
                        &   & $\blacksquare$ &0.01826   &-0.00348 &0.00043  &0.00070 &-0.01020 &0.00160 \\ 
                        \cline{2-9}

                        & \multirow{2}{*}{C F}    & $\square$ &26,248  &2,472,585  &57,314,692   &26,248 &2,472,354 &57,303,989 \\
                        \cline{3-9}
                        &   & $\blacksquare$ &0.02120   &-0.00260  &0.00010  &0.01100 &-0.00670 &0.00030 \\
                       \hline

\multirow{8}{*}{Open Source}  & \multirow{2}{*}{A-A M}  & $\square$ &415,358  &50,511,664  &1,530,499,294  &415,358 &50,508,012 &1,530,278,014 \\
                        \cline{3-9}
                        &  & $\blacksquare$ &0.04116   &-0.00071  &0.00001  &0.03060 &-0.00550 &0.00020\\
                        \cline{2-9}

                        & \multirow{2}{*}{C M}  & $\square$ &33,069   &9,032,439  &612,977,190  &33,069 &9,031,667 &612,872,152 \\
                        \cline{3-9}
                       &  & $\blacksquare$ &0.01464    &-0.00344  &0.00005  &-0.00440 &-0.01520 &0.00020 \\
                       \cline{2-9}

                        & \multirow{2}{*}{A-A F }   & $\square$ &1,541,983   &40,619,357  &266,678849    &9,134,307 &87,991,477 &211,664,737 \\
                        \cline{3-9}
                        &   & $\blacksquare$ &0.01354   &-0.00295 &0.00064  &-0.00260 &-0.00970 &0.00410 \\ 
                        \cline{2-9}

                        & \multirow{2}{*}{C F}    & $\square$ &362,090  &8,600,857  &50,850,578   &362,090 &8,600,004 &50,840,497 \\
                        \cline{3-9}
                        &   & $\blacksquare$ &0.02080   &0.00170  &-0.00050  &0.02180 &0.00060 &-0.00020 \\
                        \hline

\end{tabular}
\caption{Number of image pairs ($\square$) and the relative average impostor match scores with respect to average match scores of the overall impostor distribution ($\blacksquare$)
}
\label{table:impostor_ImagePairs_Mean}
\end{table*}
\begin{figure*}[t]
  \centering
    \begin{subfigure}[b]{1\linewidth}
      \centering
      \captionsetup[subfigure]{labelformat=empty}
      \begin{subfigure}[b]{0.239\linewidth}
        \centering
          \vspace{-0.5em}
          \caption{Caucasian Males}
          \includegraphics[width=\linewidth]{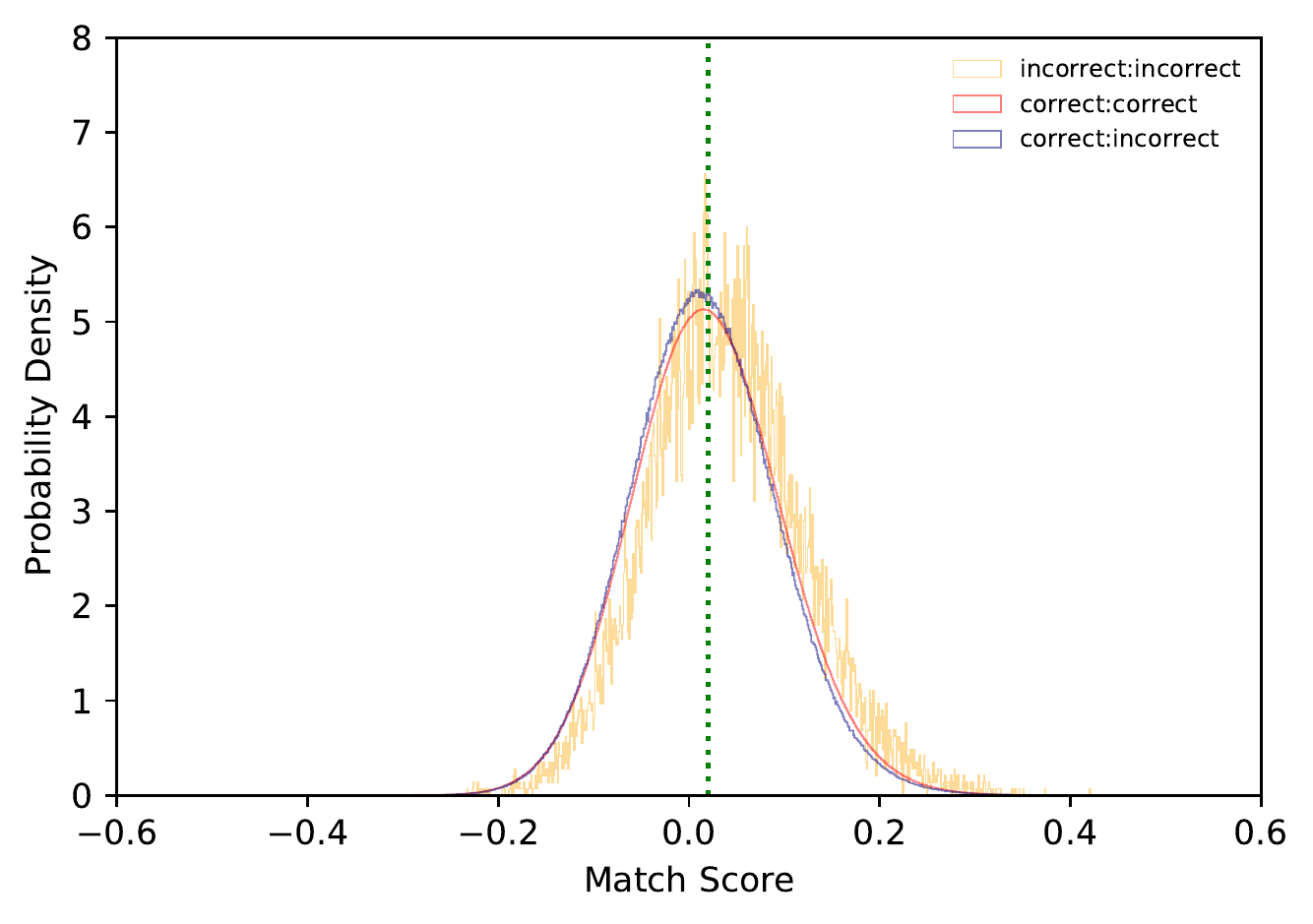}
      \end{subfigure}
      \begin{subfigure}[b]{0.239\linewidth}
        \centering
          \captionsetup[subfigure]{labelformat=empty}
          \vspace{-0.5em}
          \caption{Caucasian Females}
          \includegraphics[width=\linewidth]{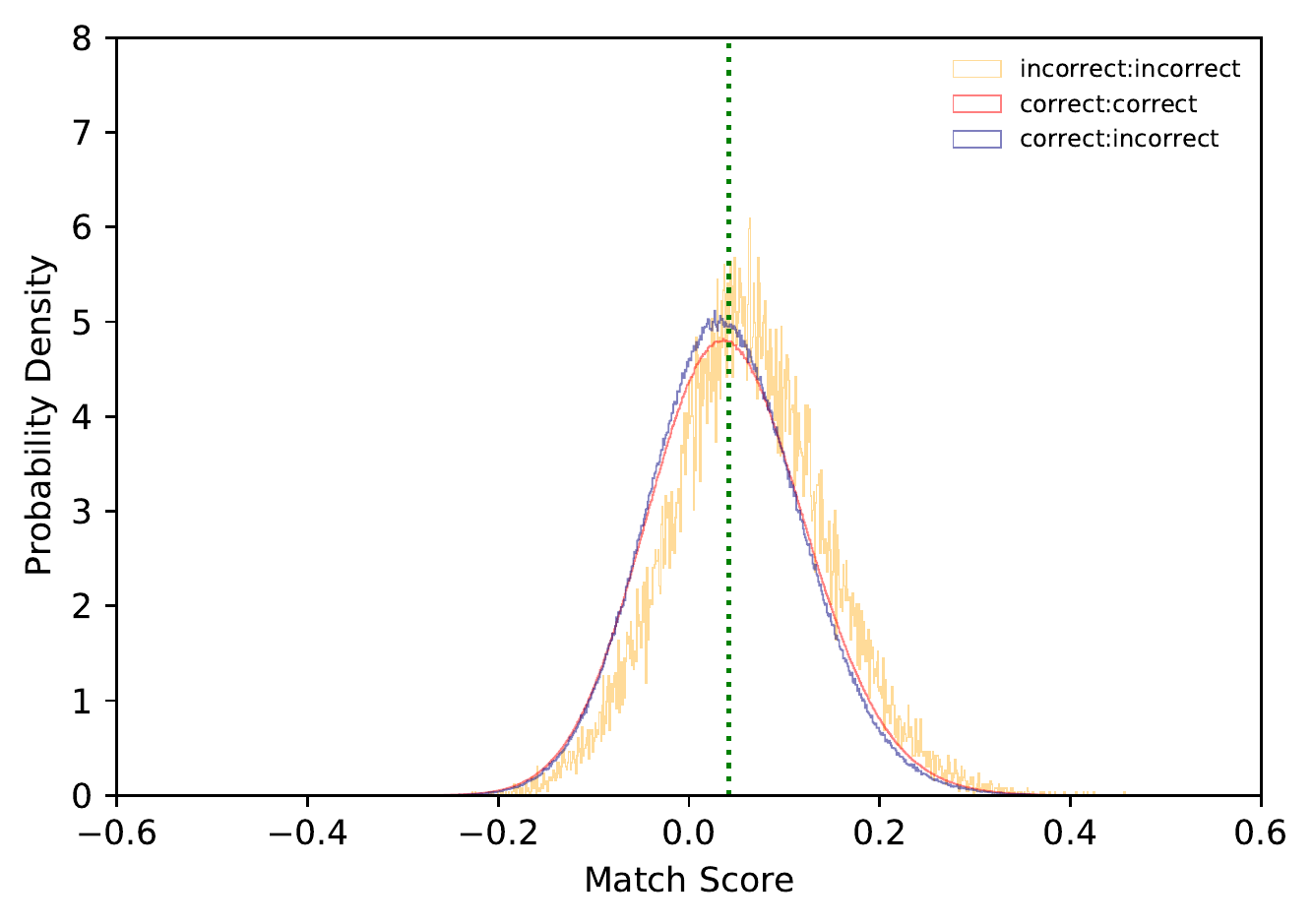}
      \end{subfigure}
      \centering
      \begin{subfigure}[b]{0.239\linewidth}
        \centering
          \captionsetup[subfigure]{labelformat=empty}
          \vspace{-0.5em}
          \caption{African-American Males}
          \includegraphics[width=\linewidth]{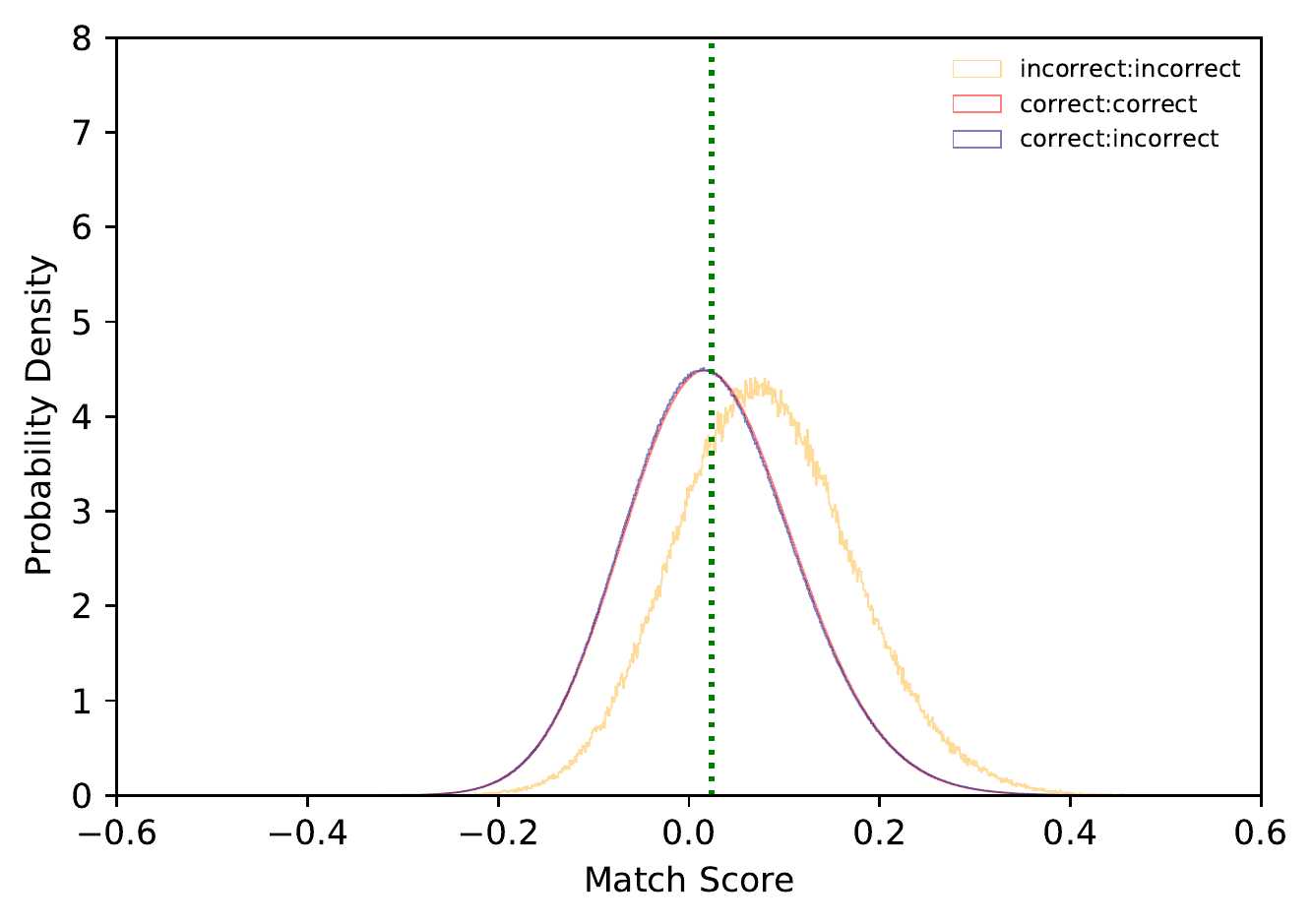}
      \end{subfigure}
      \begin{subfigure}[b]{0.239\linewidth}
        \centering
          \captionsetup[subfigure]{labelformat=empty}
          \vspace{-0.5em}
          \caption{African-American Females}
          \includegraphics[width=\linewidth]{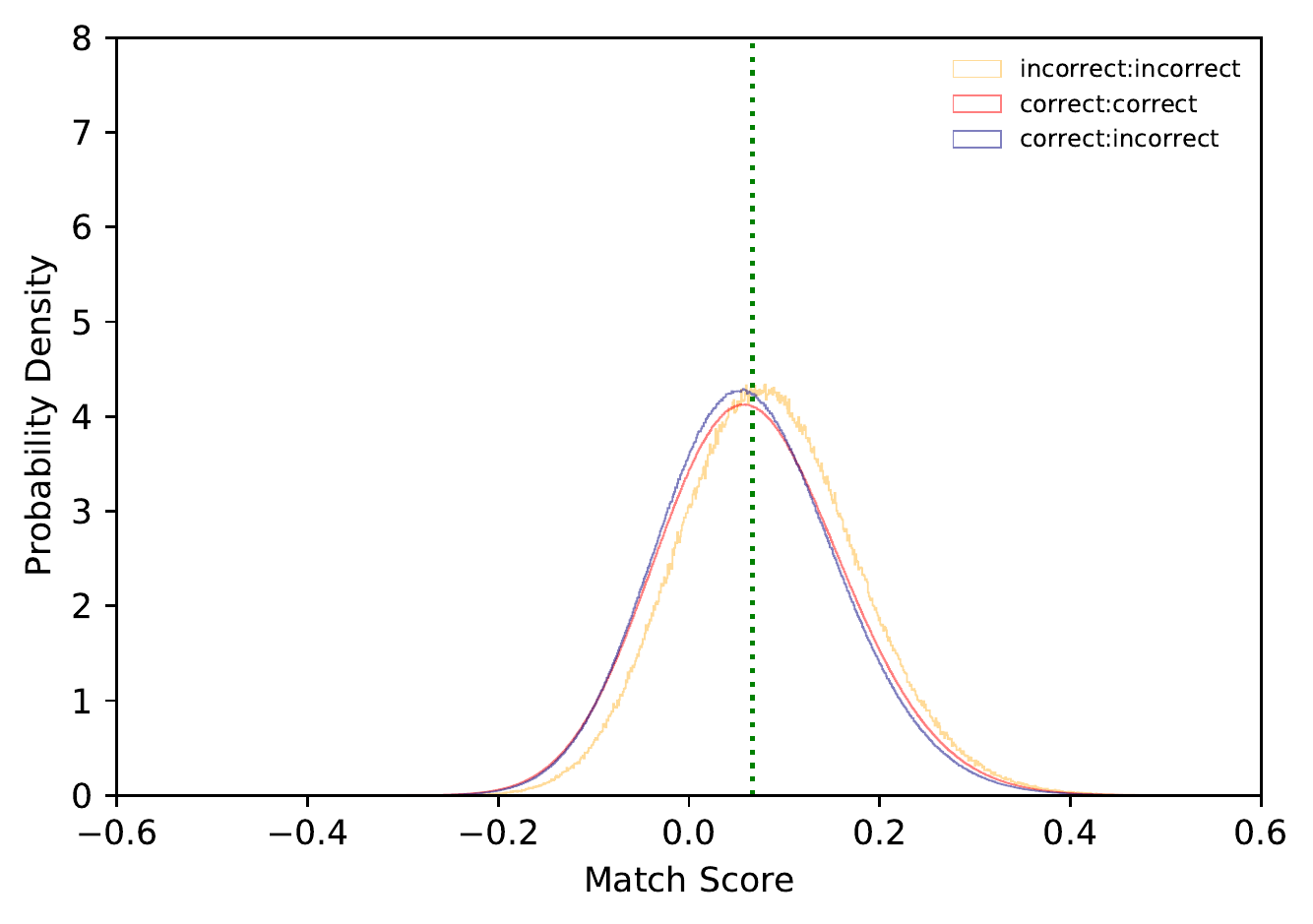}
      \end{subfigure}
      \addtocounter{subfigure}{-4}
      \caption{ArcFace and Amazon Face API}
      \vspace{0.5em}
  \end{subfigure}
  \centering
  
  \begin{subfigure}[b]{1\linewidth}
      \centering
      \begin{subfigure}[b]{0.239\linewidth}
        \centering
          \includegraphics[width=\linewidth]{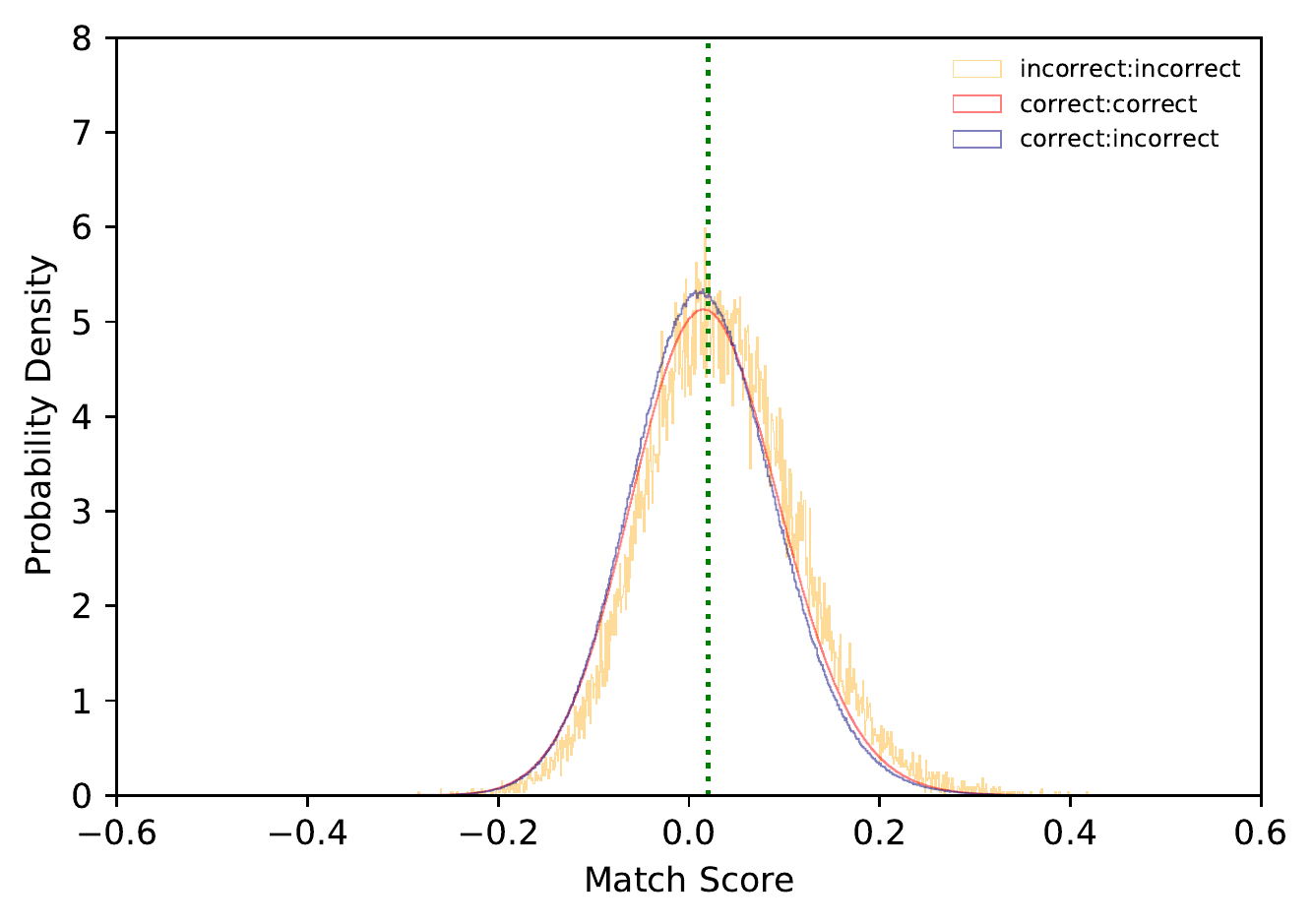}
      \end{subfigure}
      \begin{subfigure}[b]{0.239\linewidth}
        \centering
          \includegraphics[width=\linewidth]{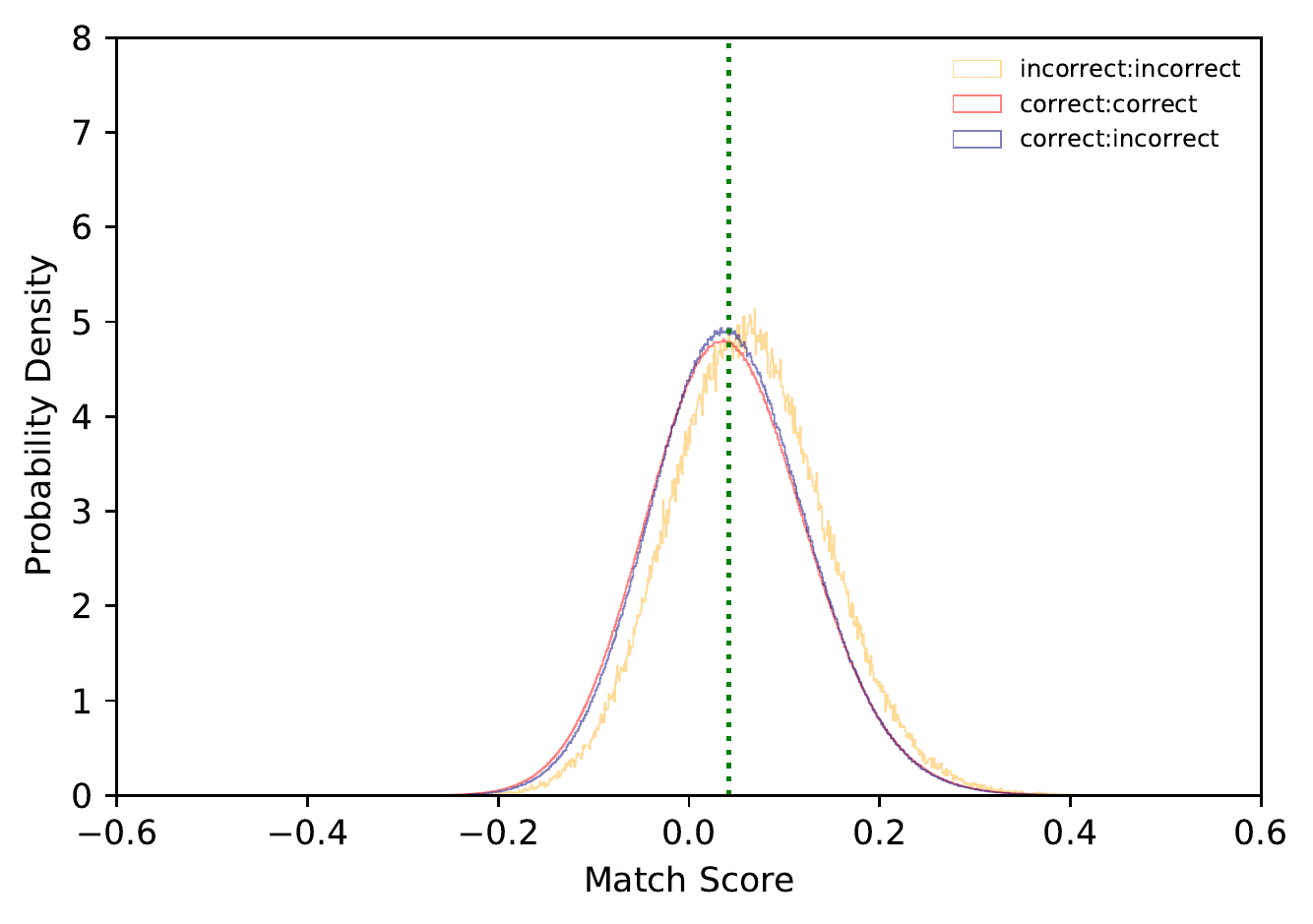}
      \end{subfigure}
      \centering
      \begin{subfigure}[b]{0.239\linewidth}
        \centering
          \includegraphics[width=\linewidth]{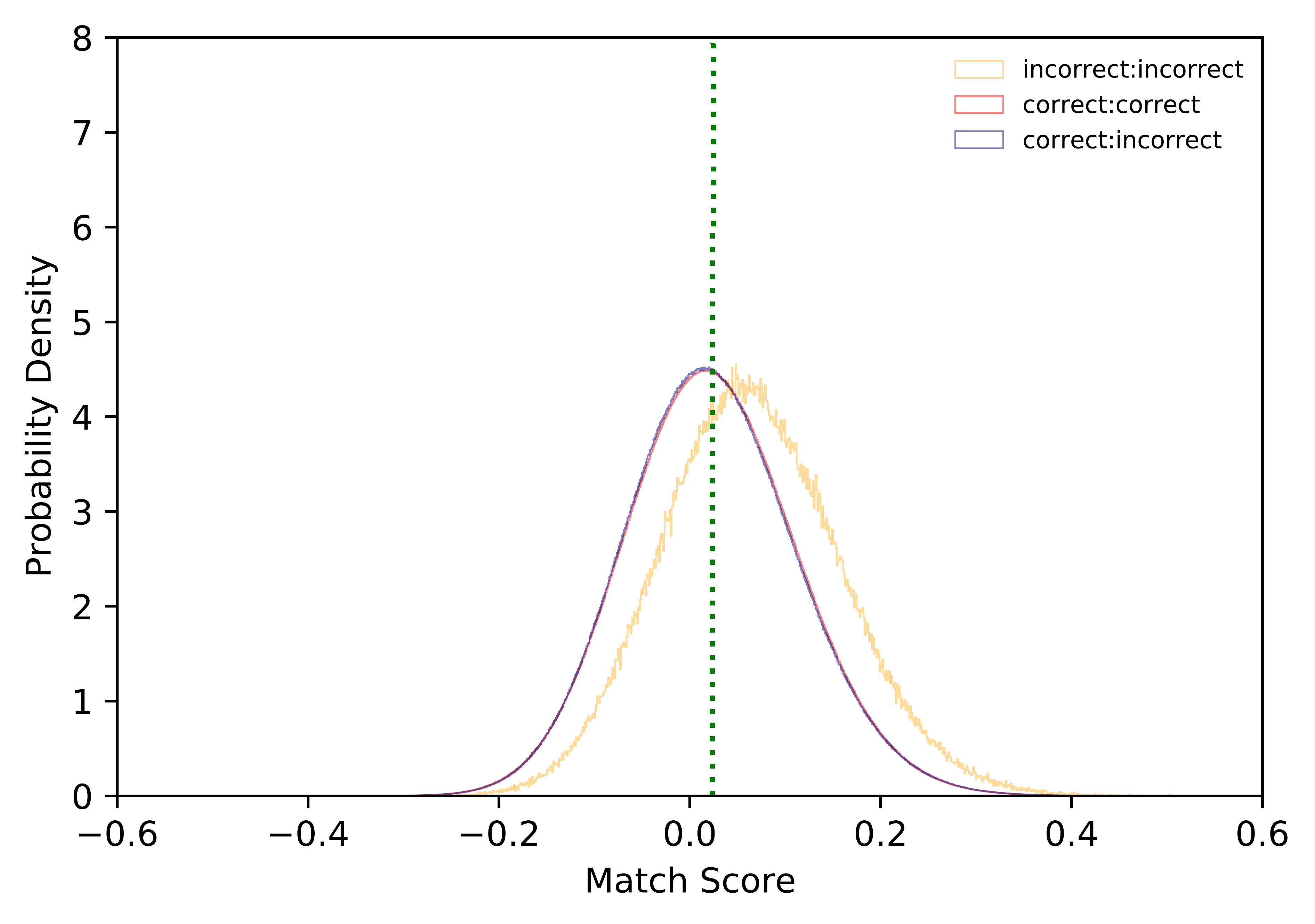}
      \end{subfigure}
      \begin{subfigure}[b]{0.239\linewidth}
        \centering
          \includegraphics[width=\linewidth]{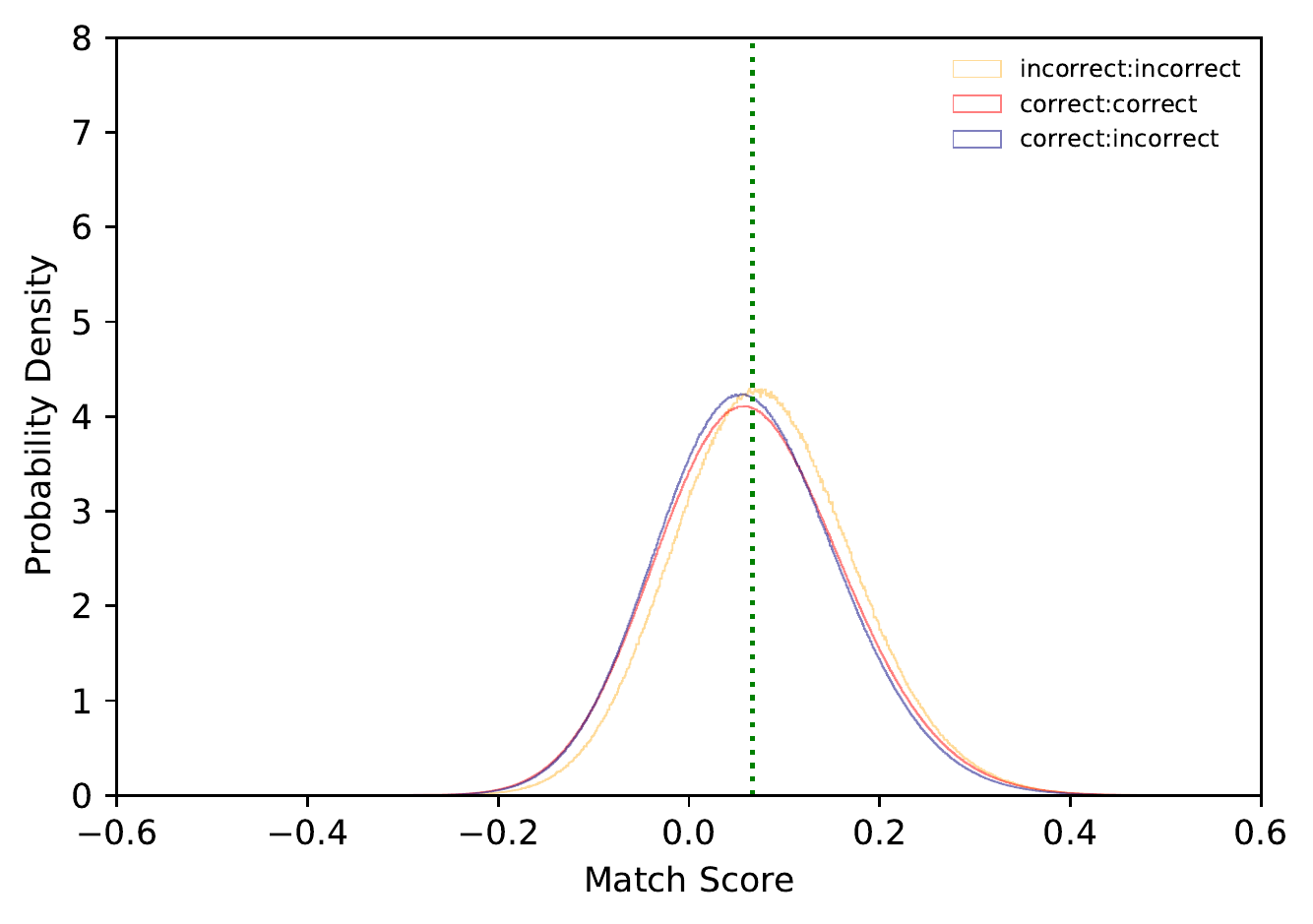}
      \end{subfigure}
      \caption{ArcFace and Open Source}
      \vspace{0.5em}
  \end{subfigure}
  
  \begin{subfigure}[b]{1\linewidth}
      \centering
      \begin{subfigure}[b]{0.239\linewidth}
        \centering
          \includegraphics[width=\linewidth]{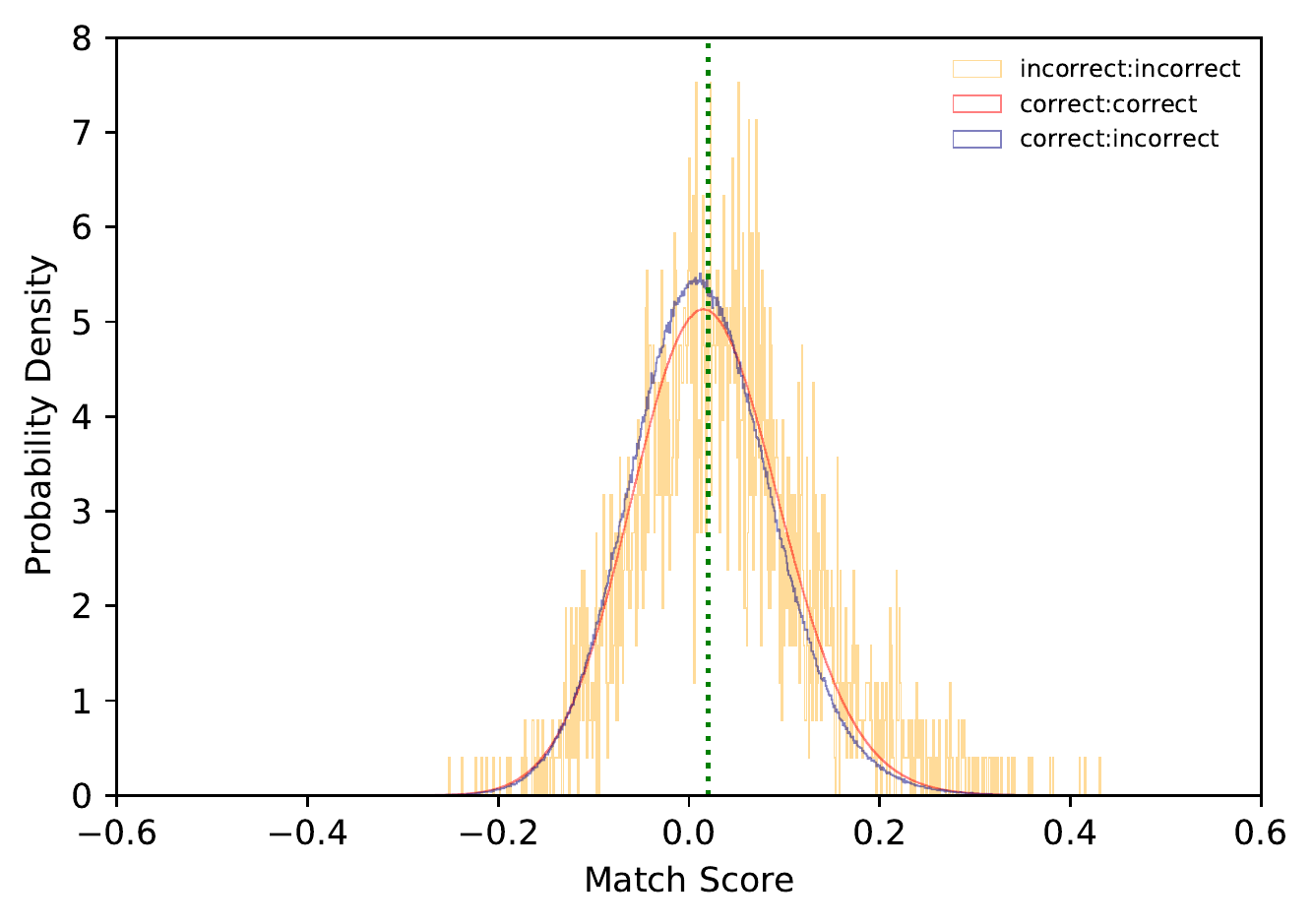}
      \end{subfigure}
      \begin{subfigure}[b]{0.239\linewidth}
        \centering
          \includegraphics[width=\linewidth]{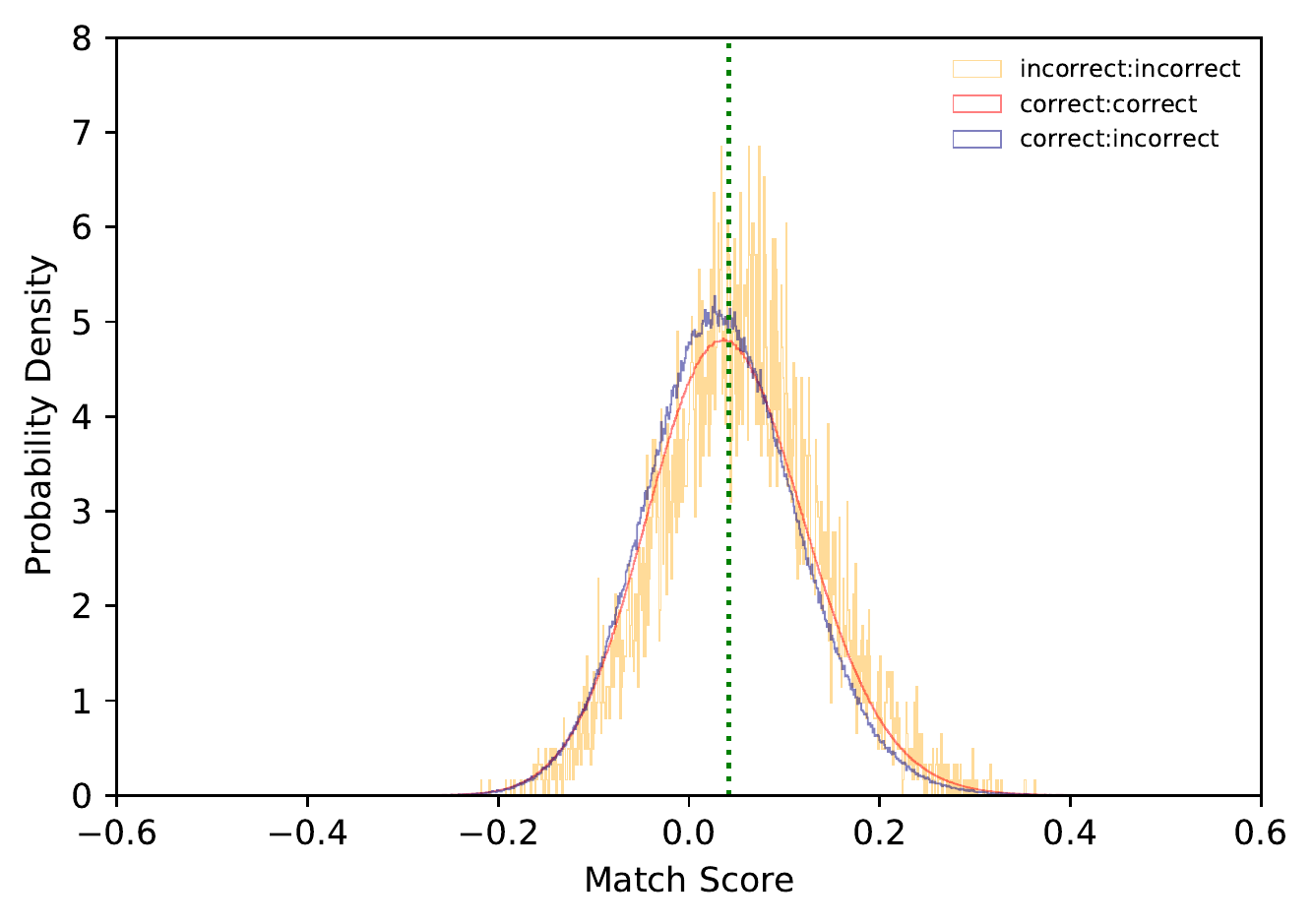}
      \end{subfigure}
      \centering
      \begin{subfigure}[b]{0.239\linewidth}
        \centering
          \includegraphics[width=\linewidth]{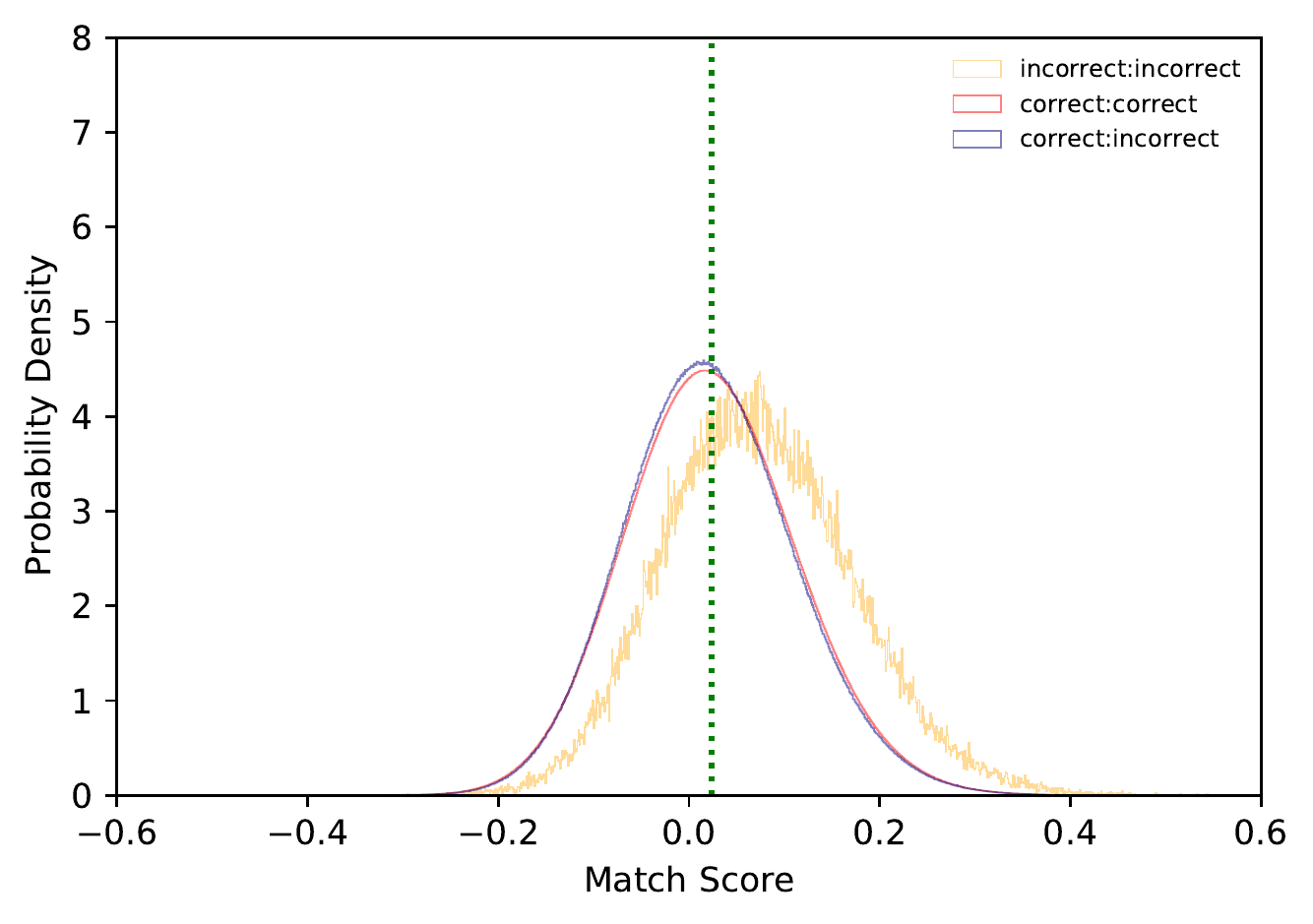}
      \end{subfigure}
      \begin{subfigure}[b]{0.239\linewidth}
        \centering
          \includegraphics[width=\linewidth]{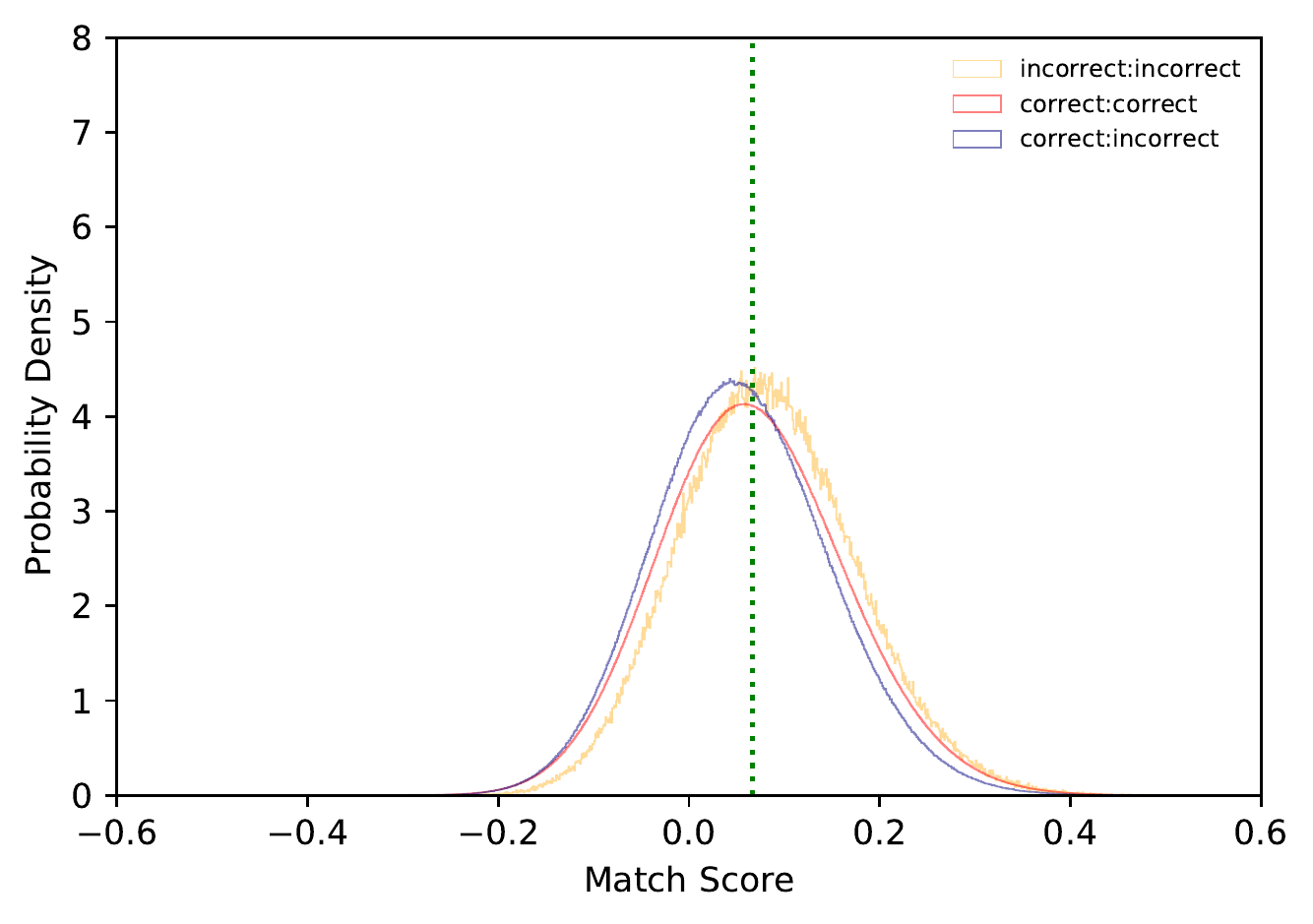}
      \end{subfigure}
      \caption{ArcFace and Microsoft Face API}
      \vspace{0.5em}
  \end{subfigure}
  
  \begin{subfigure}[b]{1\linewidth}
      \centering
      \begin{subfigure}[b]{0.239\linewidth}
        \centering
          \includegraphics[width=\linewidth]{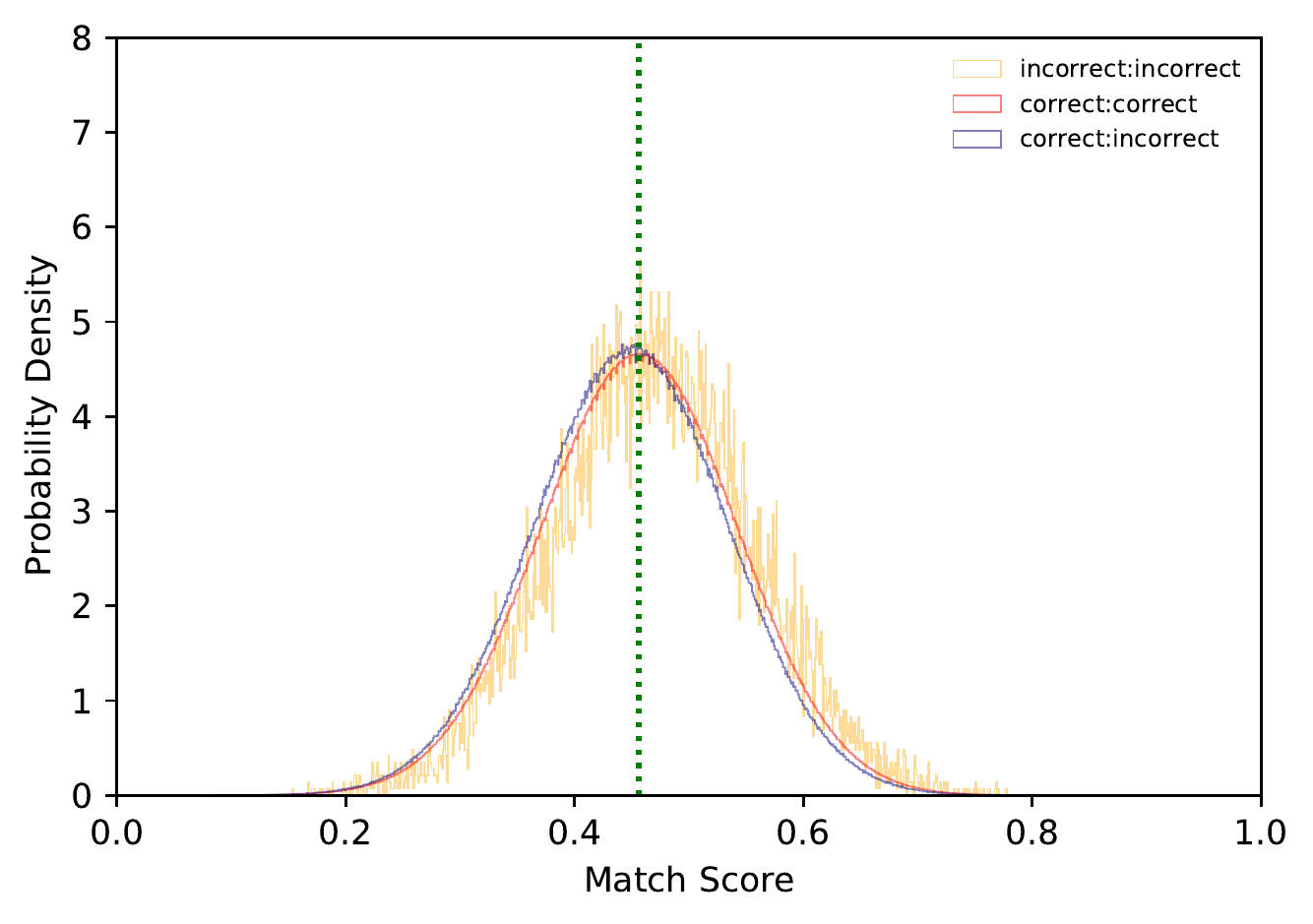}
      \end{subfigure}
      \begin{subfigure}[b]{0.239\linewidth}
        \centering
          \includegraphics[width=\linewidth]{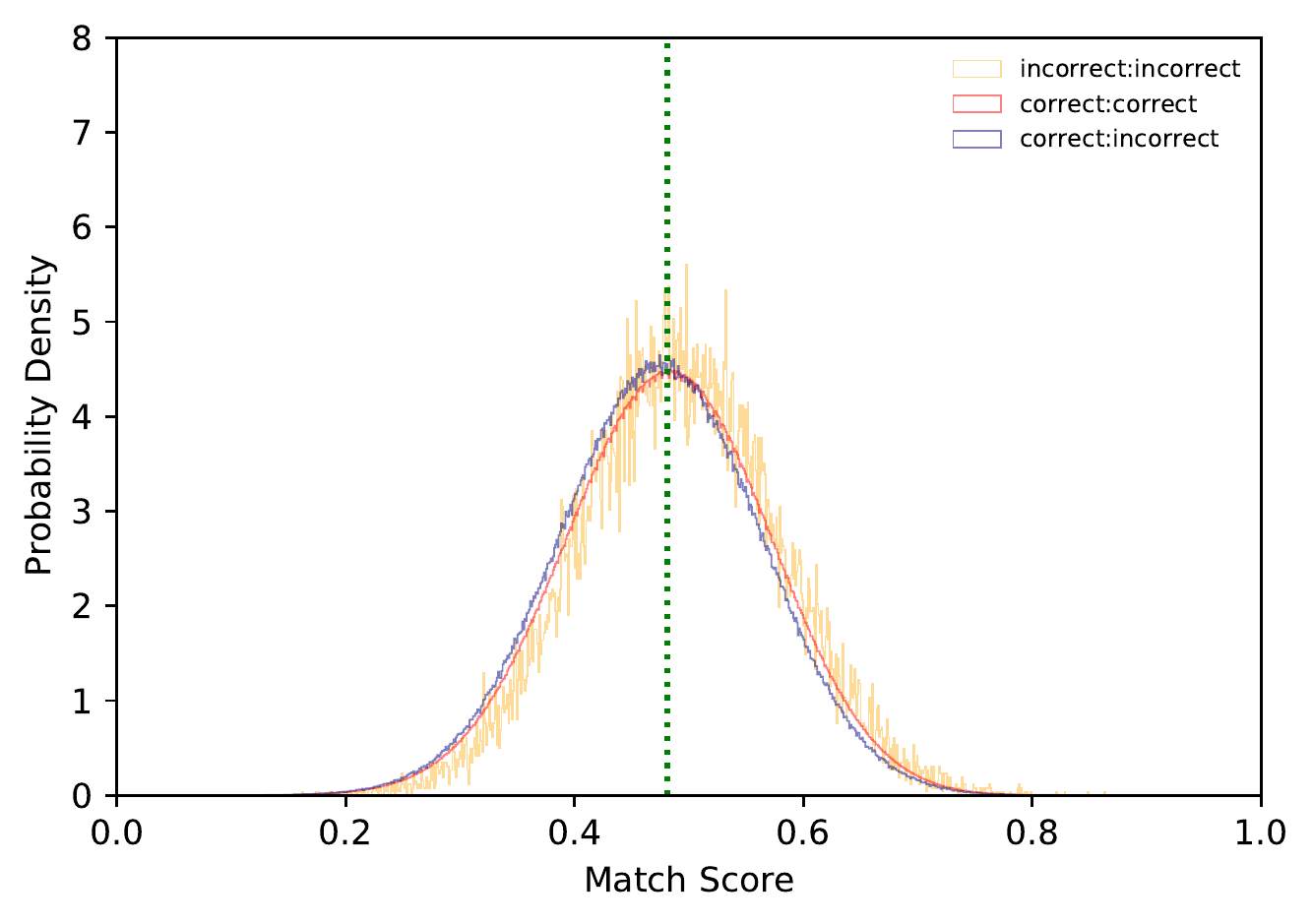}
      \end{subfigure}
      \centering
      \begin{subfigure}[b]{0.239\linewidth}
        \centering
          \includegraphics[width=\linewidth]{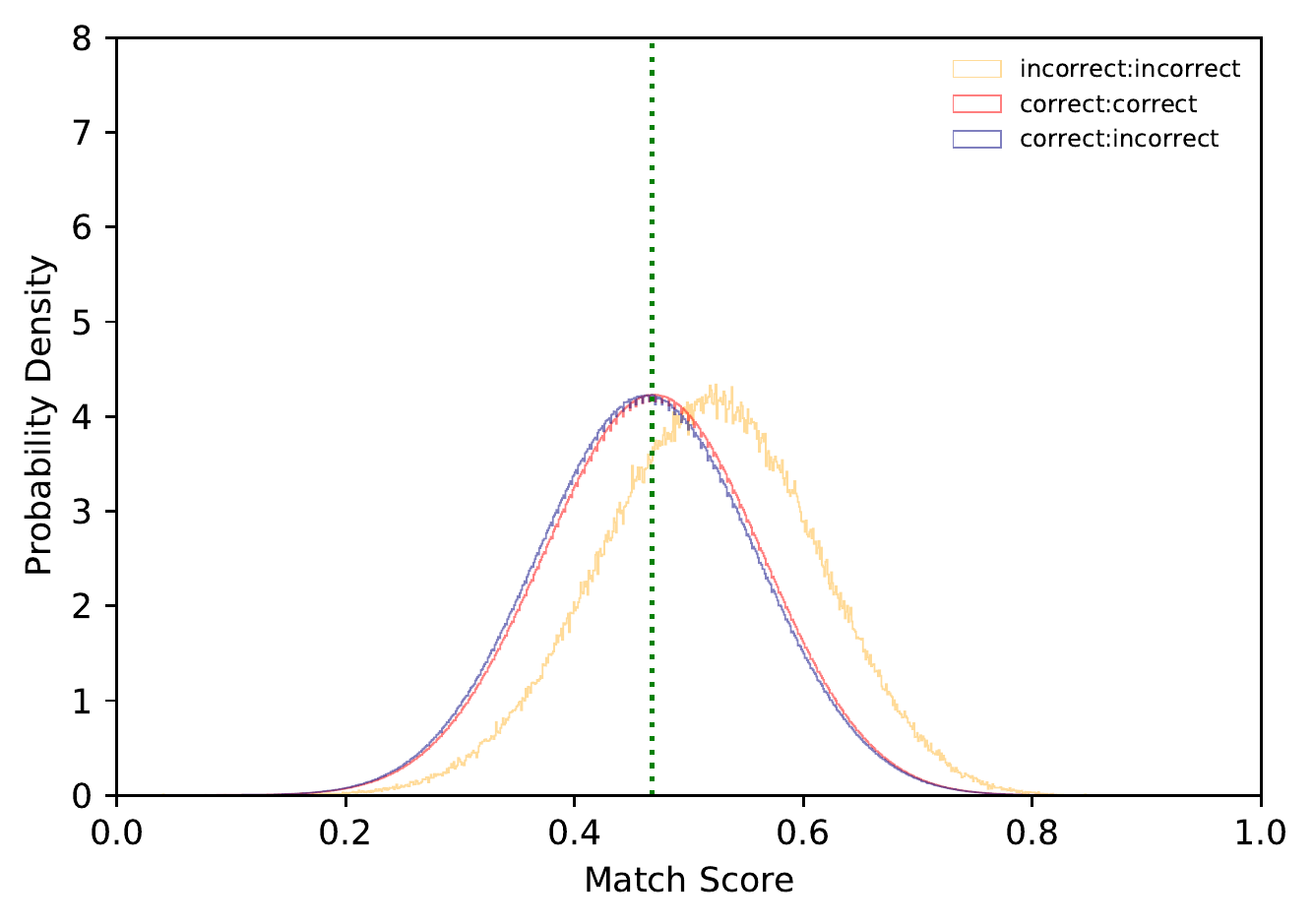}
      \end{subfigure}
      \begin{subfigure}[b]{0.239\linewidth}
        \centering
          \includegraphics[width=\linewidth]{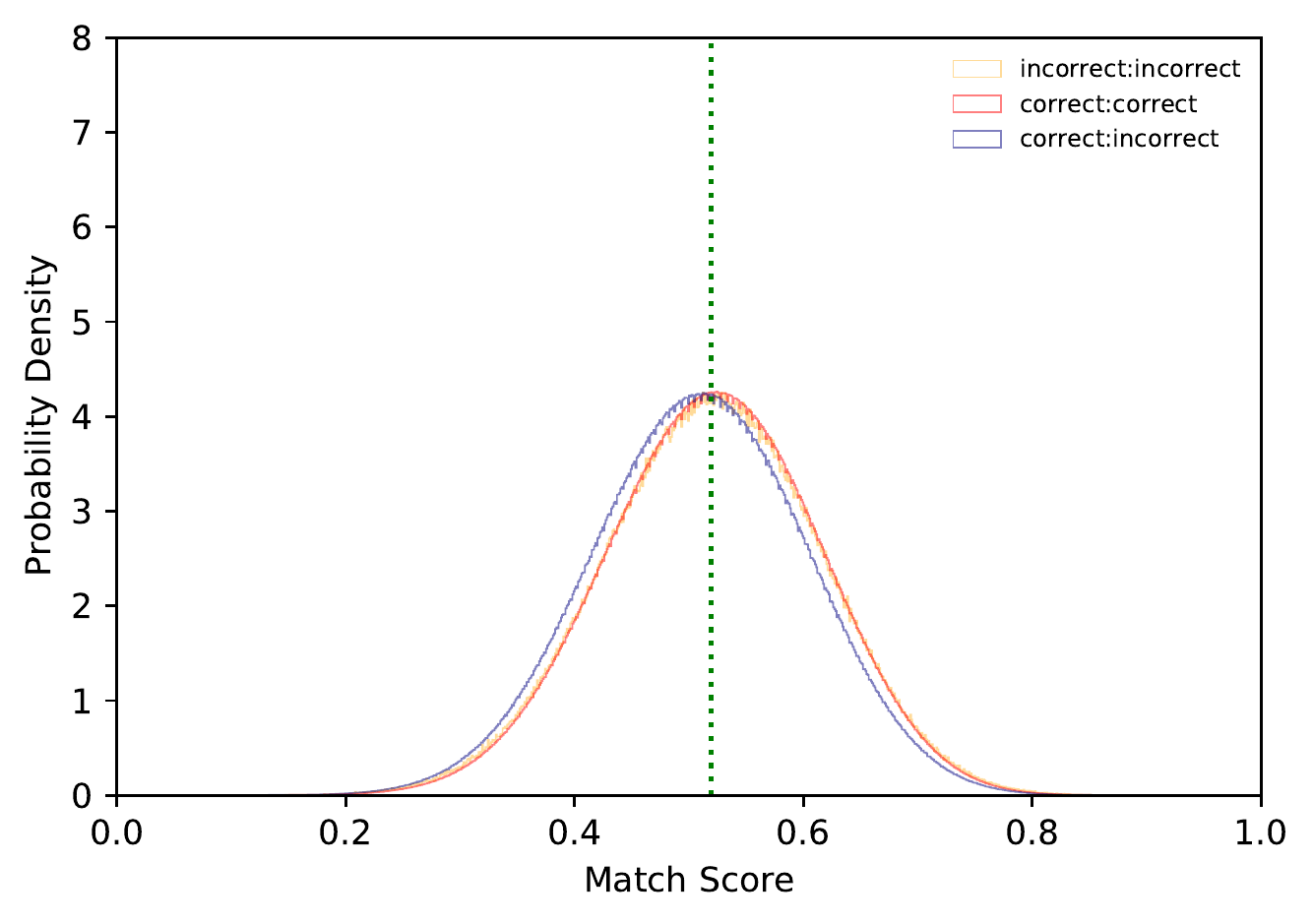}
      \end{subfigure}
      \caption{COTS and Amazon Face API}
      \vspace{0.5em}
  \end{subfigure}
  
  \begin{subfigure}[b]{1\linewidth}
      \centering
      \begin{subfigure}[b]{0.239\linewidth}
        \centering
          \includegraphics[width=\linewidth]{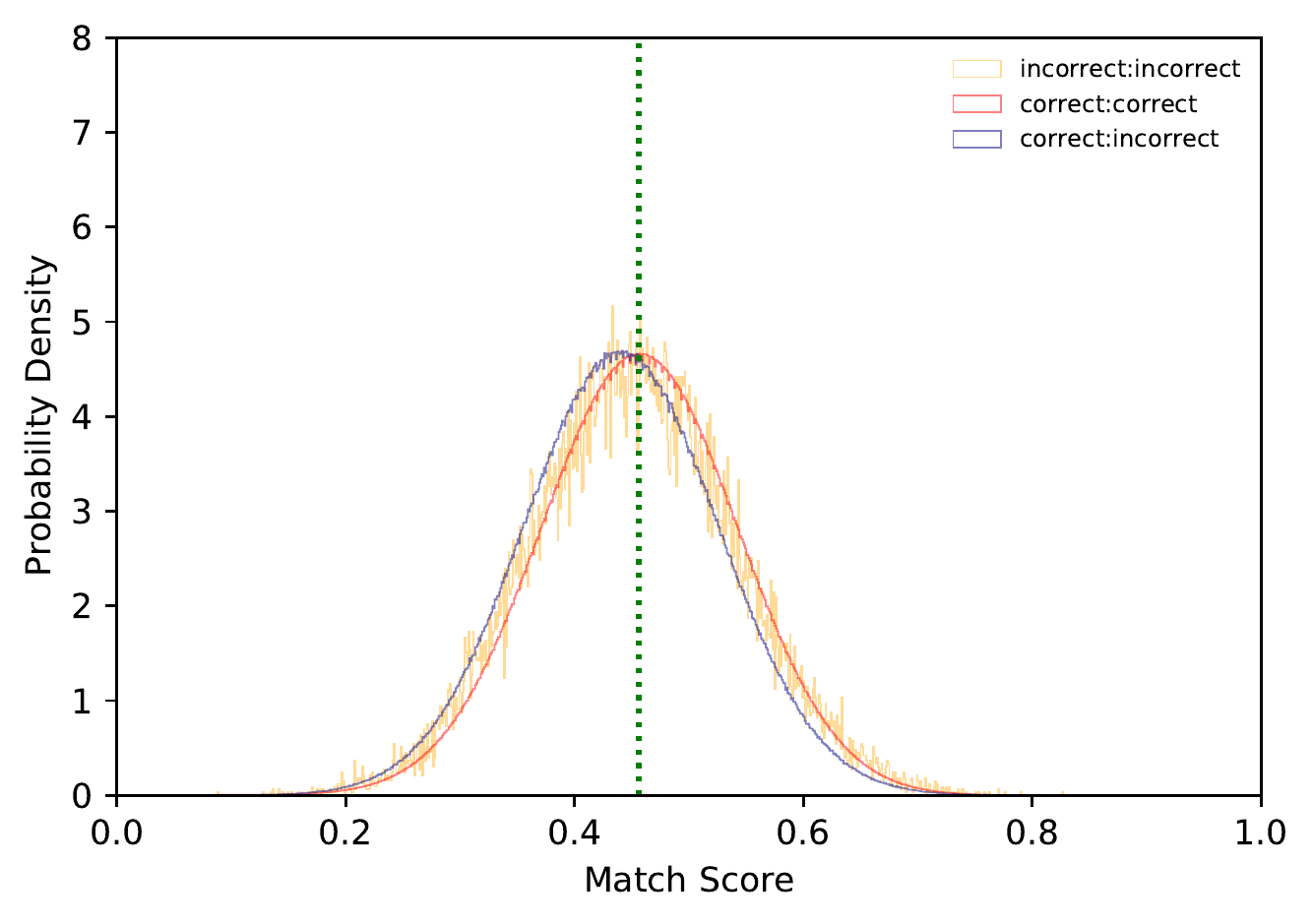}
      \end{subfigure}
      \begin{subfigure}[b]{0.239\linewidth}
        \centering
          \includegraphics[width=\linewidth]{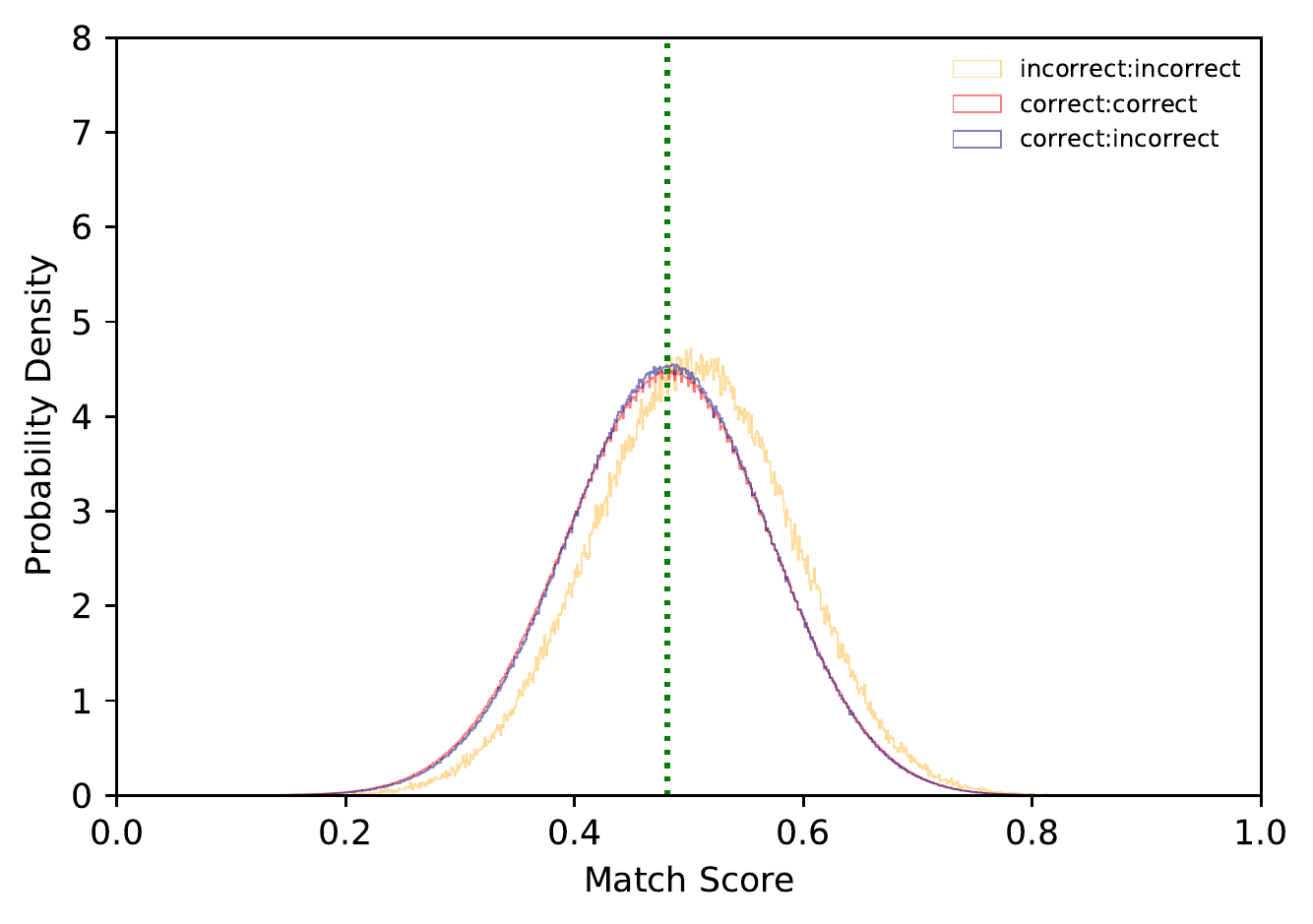}
      \end{subfigure}
      \centering
      \begin{subfigure}[b]{0.239\linewidth}
        \centering
          \includegraphics[width=\linewidth]{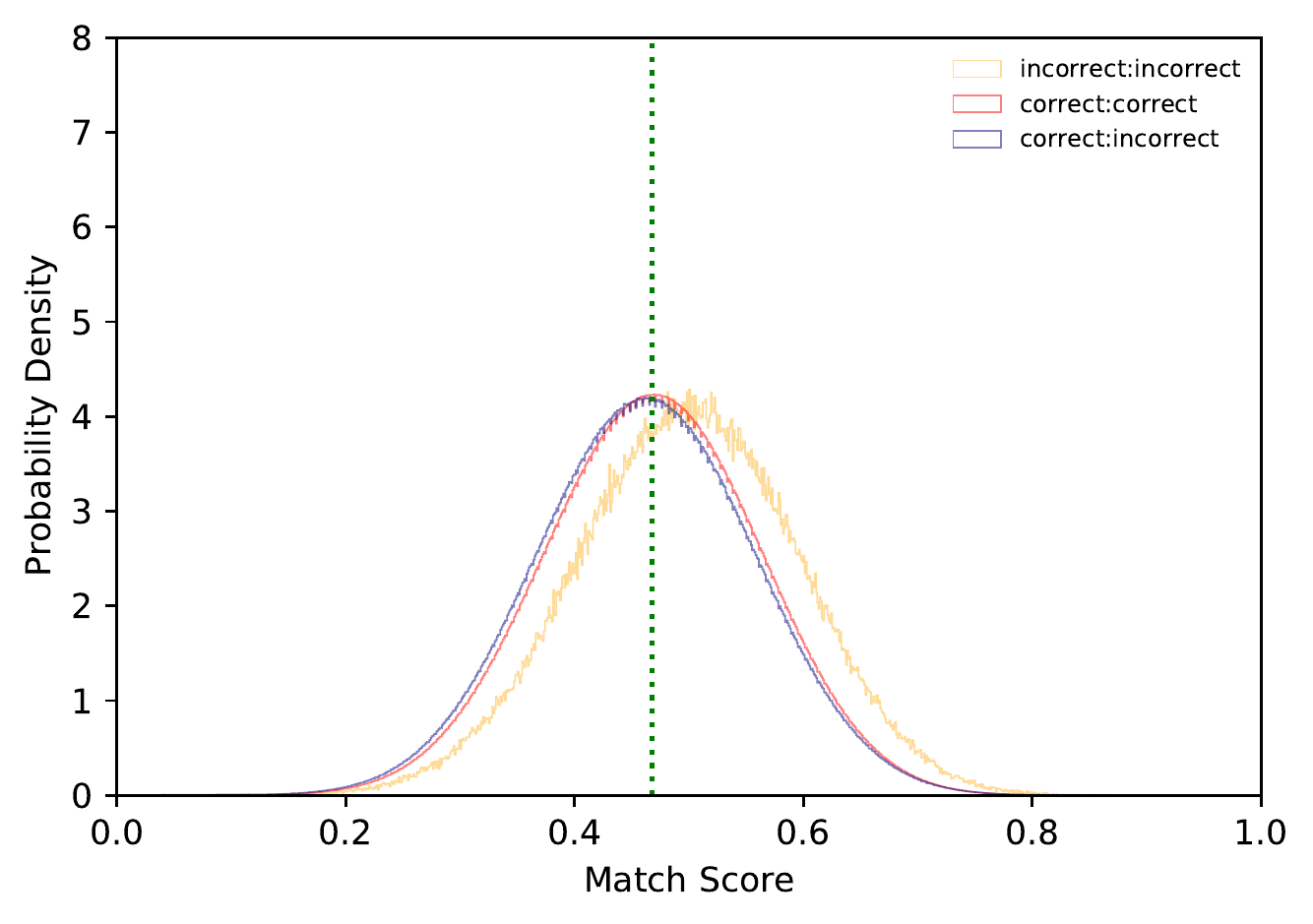}
      \end{subfigure}
      \begin{subfigure}[b]{0.239\linewidth}
        \centering
          \includegraphics[width=\linewidth]{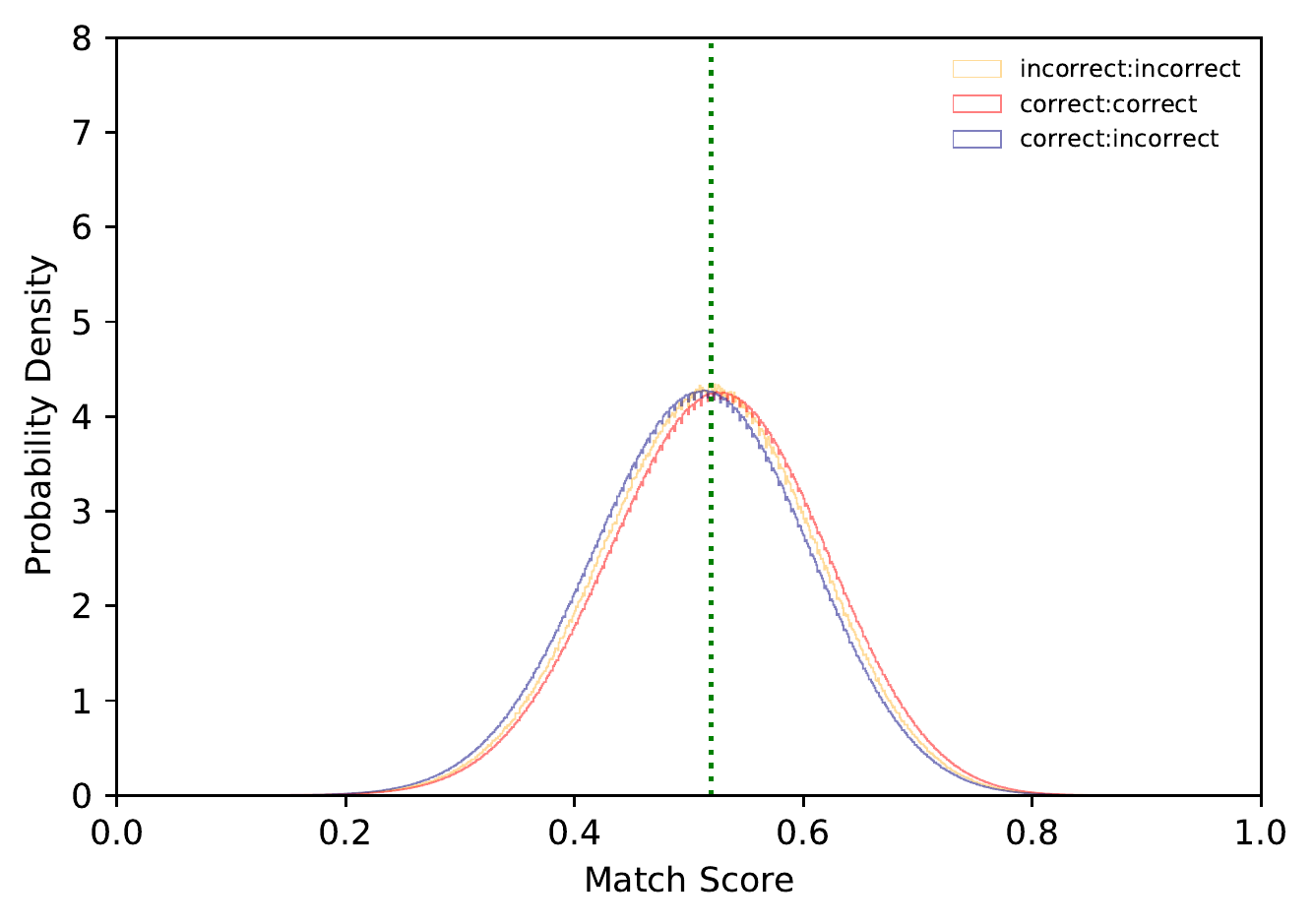}
      \end{subfigure}
      \caption{COTS and Open Source}
      \vspace{0.5em}
  \end{subfigure}
  
  \begin{subfigure}[b]{1\linewidth}
      \centering
      \begin{subfigure}[b]{0.239\linewidth}
        \centering
          \includegraphics[width=\linewidth]{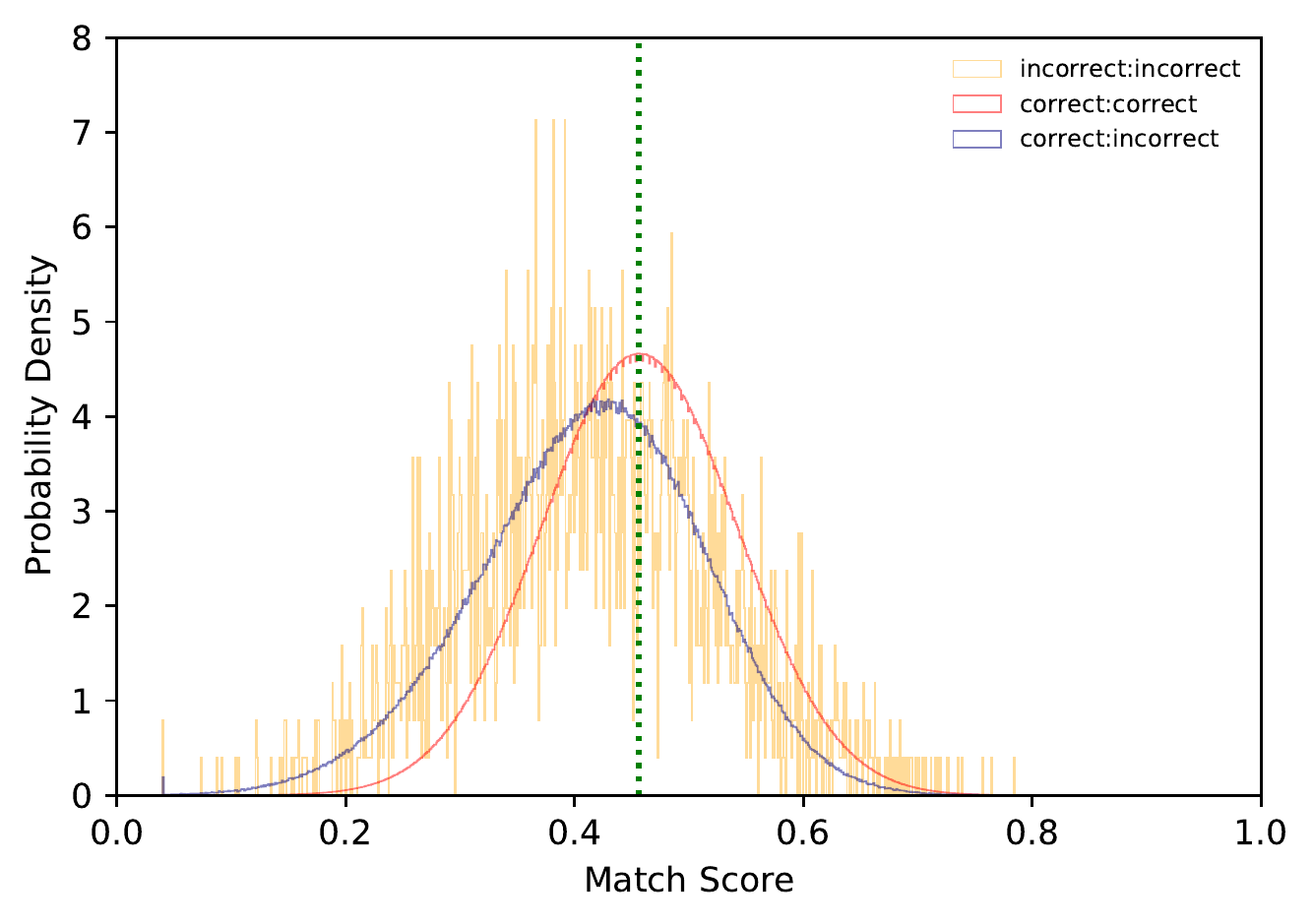}
      \end{subfigure}
      \begin{subfigure}[b]{0.239\linewidth}
        \centering
          \includegraphics[width=\linewidth]{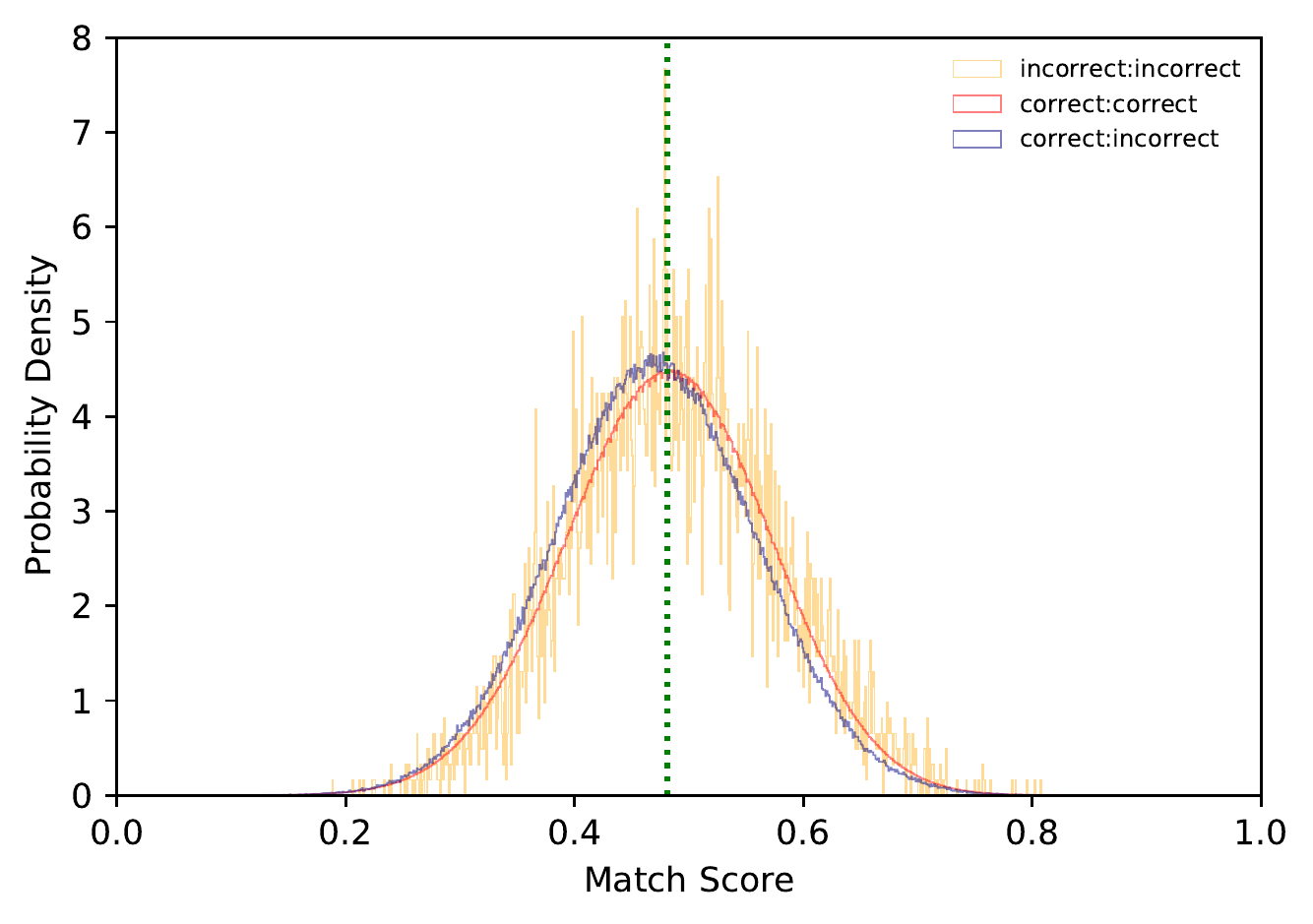}
      \end{subfigure}
      \centering
      \begin{subfigure}[b]{0.239\linewidth}
        \centering
          \includegraphics[width=\linewidth]{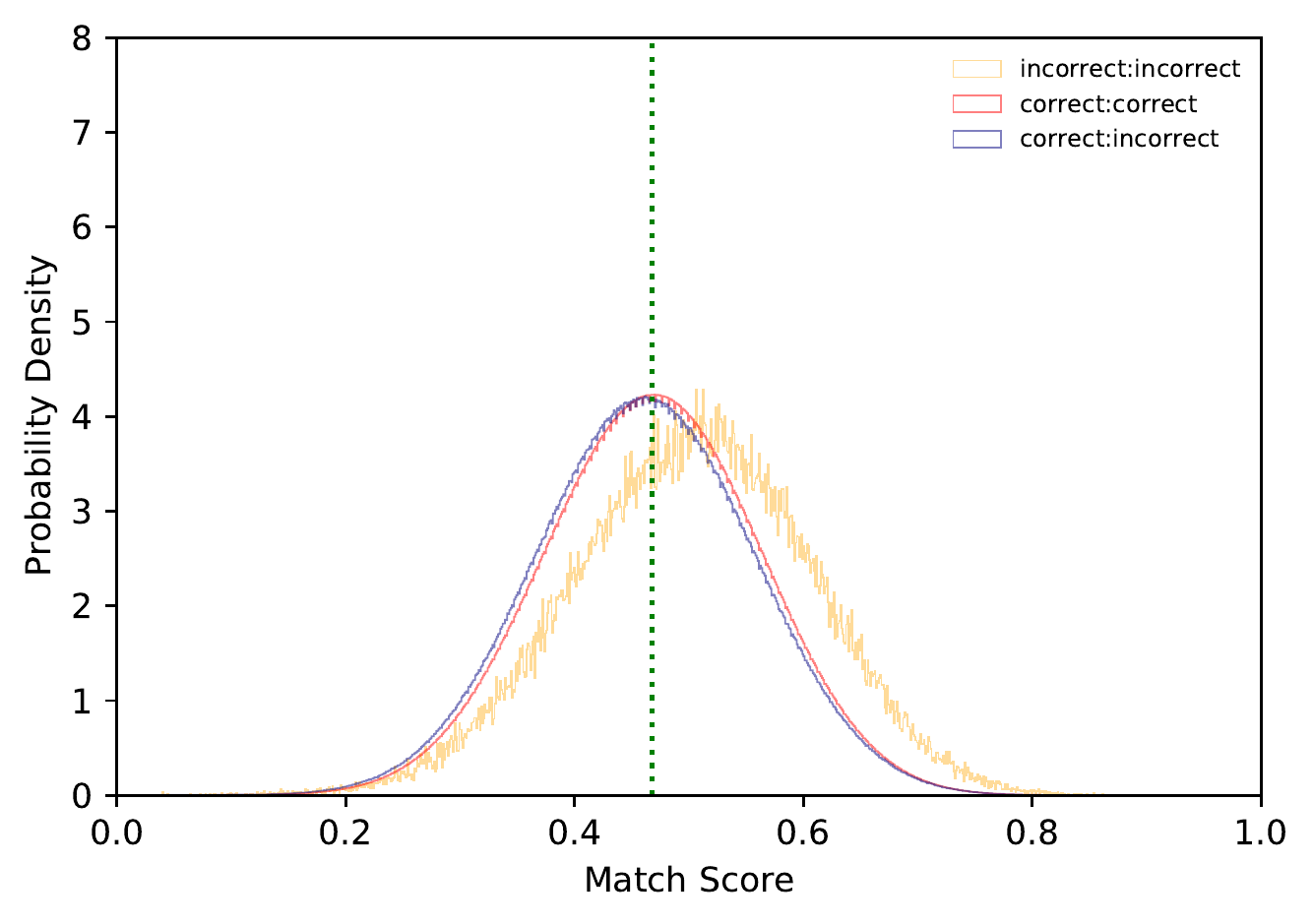}
      \end{subfigure}
      \begin{subfigure}[b]{0.239\linewidth}
        \centering
          \includegraphics[width=\linewidth]{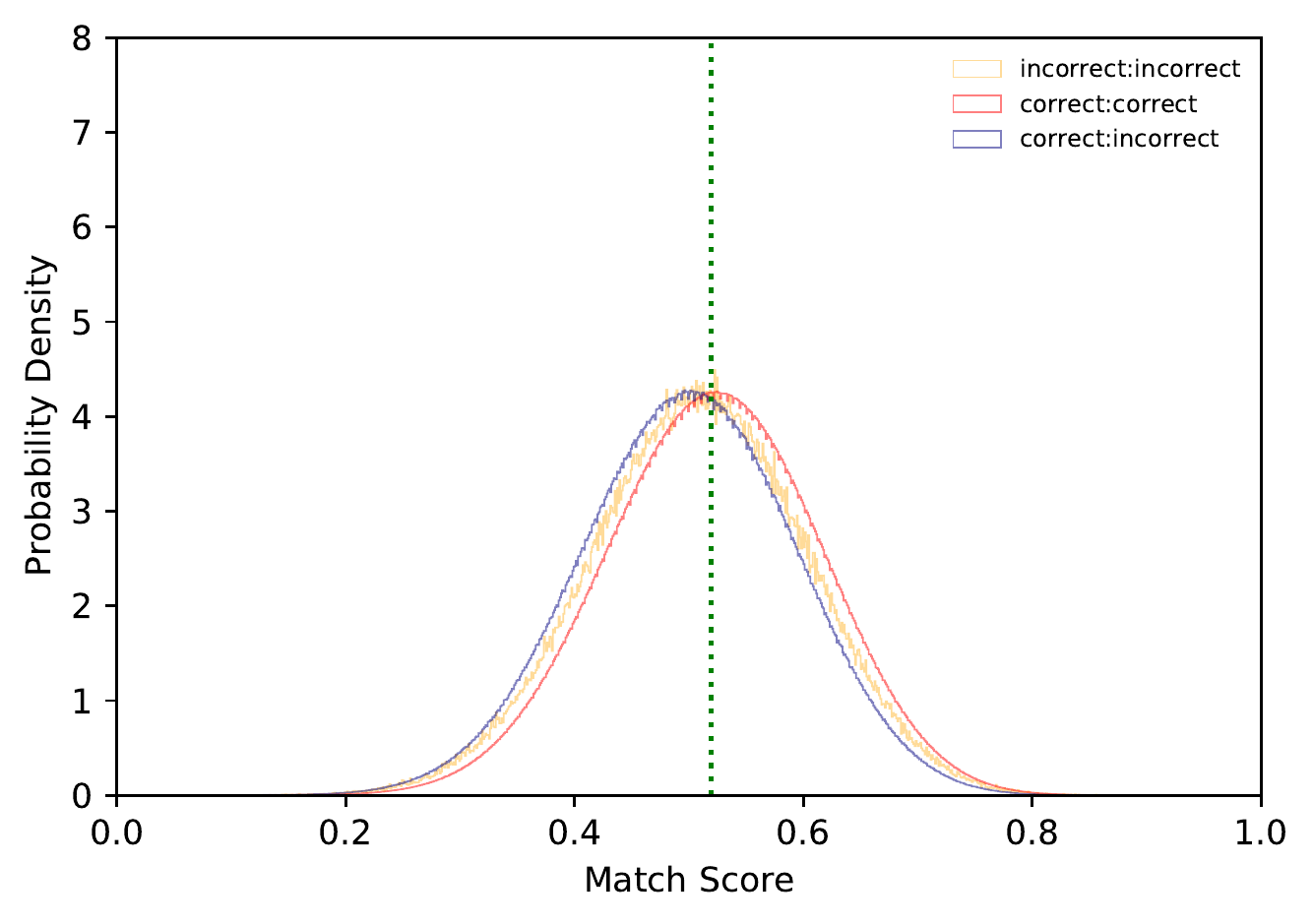}
      \end{subfigure}
      \caption{COTS and Microsoft Face API}
      \vspace{-0.5em}
  \end{subfigure}
  \caption{Impostor distribution split into images without gender error, with at least one with gender error, and with both with gender errors.}
  \label{fig:authentic}
\end{figure*}
\clearpage
\begin{table*}[t]
    \centering
        \begin{tabular}{|l|r|r|r|r|r|}
        \hline
        Gender Classifier & Cohort &  & Two Errors  & One Error & No Error \\
        \hline
\multirow{8}{*}{Microsoft}  & \multirow{2}{*}{A-A M}  & $\square$ &293 &96,381 &7,429,215    \\
                        \cline{3-6}
                        &  & $\blacksquare$ &0.05231 &-0.00268 &0.00003\\
                        \cline{2-6}

                        & \multirow{2}{*}{C M}  & $\square$ &15 &27,114 &7,498,760 \\
                        \cline{3-6}
                        &  & $\blacksquare$ &-0.02423 &-0.01790 &0.00006 \\
                       \cline{2-6}

                        & \multirow{2}{*}{A-A F }   & $\square$ &6,050 &418,360 &7,101,479 \\
                        \cline{3-6}
                        &   & $\blacksquare$ &0.01408 &-0.00971 &0.00056 \\ 
                        \cline{2-6}

                        & \multirow{2}{*}{C F}    & $\square$ &458 &119,308 &7,406,123 \\
                        \cline{3-6}
                        &   & $\blacksquare$ &0.01828 &-0.00614 &0.00010 \\
                        \hline
\multirow{8}{*}{Amazon}  & \multirow{2}{*}{A-A M}  & $\square$ &1,875 &236,715 &7,287,299    \\
                        \cline{3-6}
                        &  & $\blacksquare$  &0.05917 &-0.00178 &0.00004\\
                        \cline{2-6}

                        & \multirow{2}{*}{C M}  & $\square$ &248 &88,710 &7,436,931 \\
                        \cline{3-6}
                        &  & $\blacksquare$  &0.01605 &-0.00461 &0.00005 \\
                       \cline{2-6}

                        & \multirow{2}{*}{A-A F }   & $\square$ &26,492 &842,860 &6,656,537\\
                        \cline{3-6}
                        &   & $\blacksquare$  &0.01994 &-0.00255 &0.00024 \\ 
                        \cline{2-6}

                        & \multirow{2}{*}{C F}    & $\square$ &1,114 &183,919 &73,408,856 \\
                        \cline{3-6}
                        &   & $\blacksquare$ &0.03211 &-0.00365 &0.00009 \\
                        \hline

\multirow{8}{*}{Open Source}  & \multirow{2}{*}{A-A M}  & $\square$ &1,761 &229,185 &7,294,943    \\
                        \cline{3-6}
                        &  & $\blacksquare$ &0.04256 &-0.00196 &0.00005\\
                        \cline{2-6}

                        & \multirow{2}{*}{C M}  & $\square$ &589 &134,573 &7,390,727 \\
                        \cline{3-6}
                        &  & $\blacksquare$ &0.00713 &-0.00457 &0.00008 \\
                       \cline{2-6}

                        & \multirow{2}{*}{A-A F }   & $\square$ &171,172 &1,930,257 &5,424,460\\
                        \cline{3-6}
                        &   & $\blacksquare$ &0.01251 &-0.00315 &0.00072 \\
                        \cline{2-6}

                        & \multirow{2}{*}{C F}    & $\square$ &31,031 &907,484 &6,587,374 \\
                        \cline{3-6}
                        &   & $\blacksquare$ &0.02628 &0.00248 &-0.00047 \\
                        \hline

\end{tabular}
\vspace{-0.5em}
\caption{Number of image pairs ($\square$) and the relative average impostor match scores with respect to average match scores of the overall impostor distribution ($\blacksquare$) based on a MORPH subset that was balanced to have same number of subjects, with same amount of images, and same age distribution.
}
\label{table:impostor_BalancedDataset_ImagePairs_Mean}
\end{table*}
\begin{figure*}[ht]
  \centering
    \begin{subfigure}[b]{1\linewidth}
      \centering
      \captionsetup[subfigure]{labelformat=empty}
      \begin{subfigure}[b]{0.239\linewidth}
        \centering
          \caption{Caucasian Males}
          \vspace{-0.5em}
          \includegraphics[width=\linewidth]{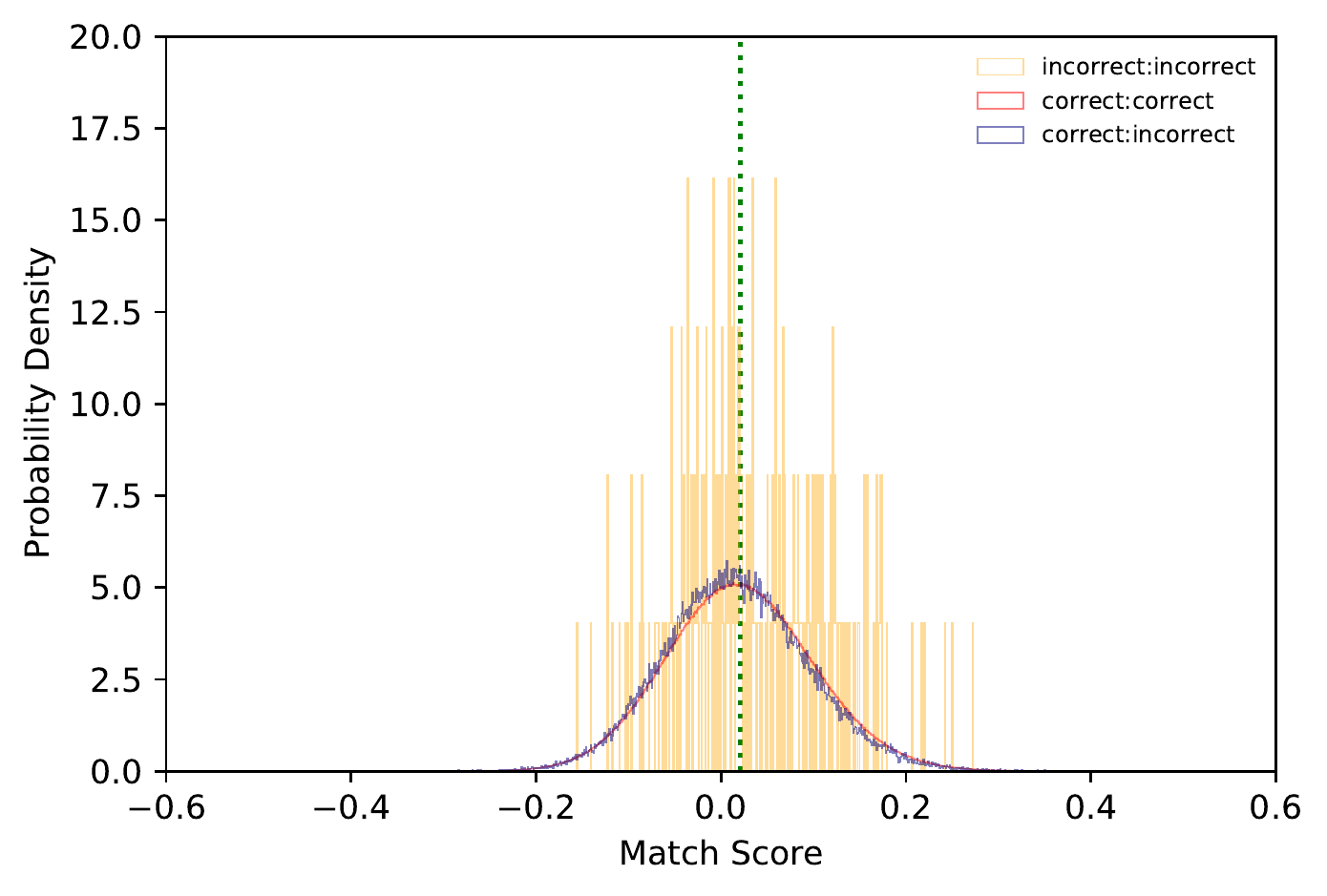}
      \end{subfigure}
      \begin{subfigure}[b]{0.239\linewidth}
        \centering
          \captionsetup[subfigure]{labelformat=empty}
          \caption{Caucasian Females}
          \vspace{-0.5em}
          \includegraphics[width=\linewidth]{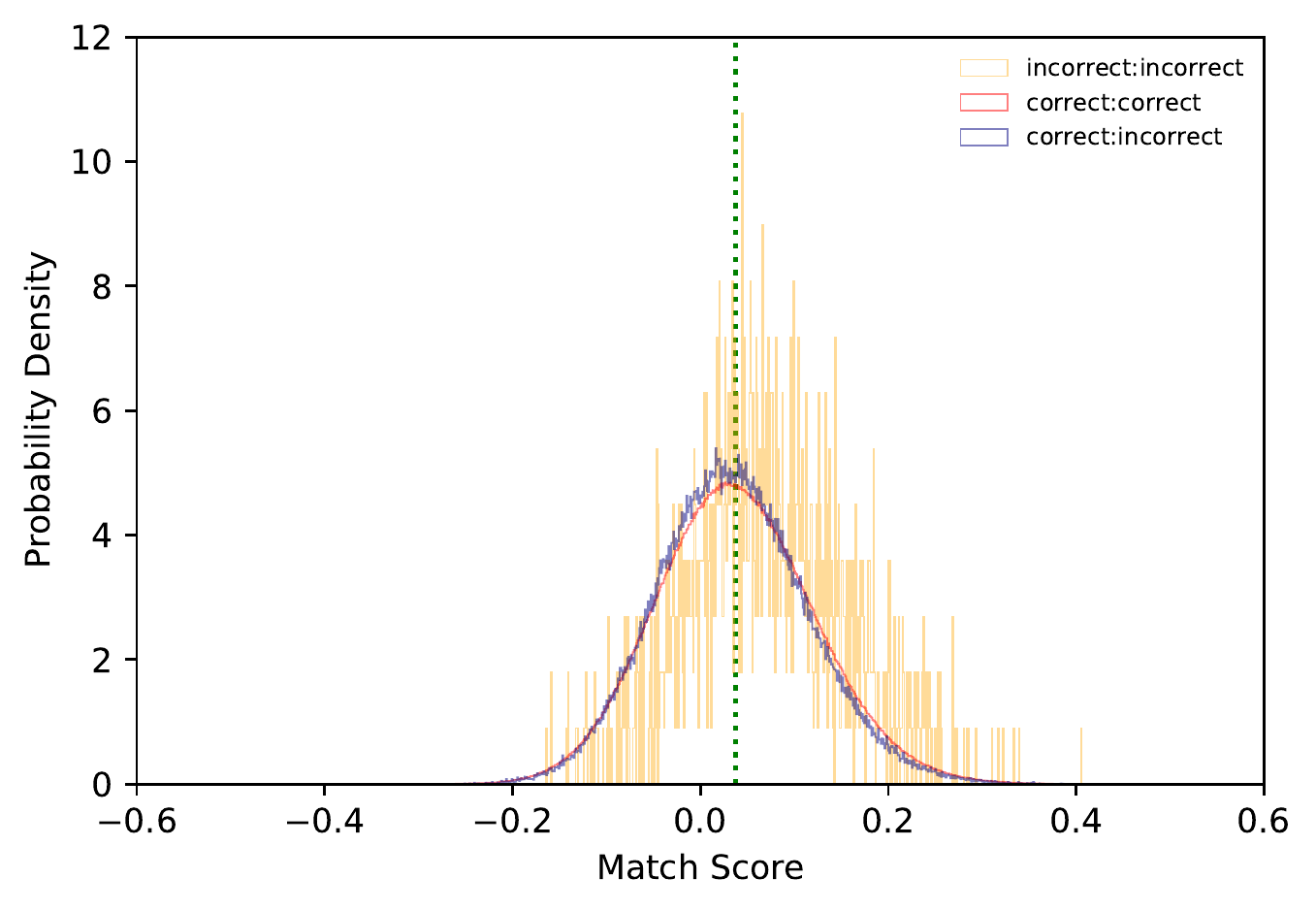}
      \end{subfigure}
      \centering
      \begin{subfigure}[b]{0.239\linewidth}
        \centering
          \captionsetup[subfigure]{labelformat=empty}
          \caption{African-American Males}
          \vspace{-0.5em}
          \includegraphics[width=\linewidth]{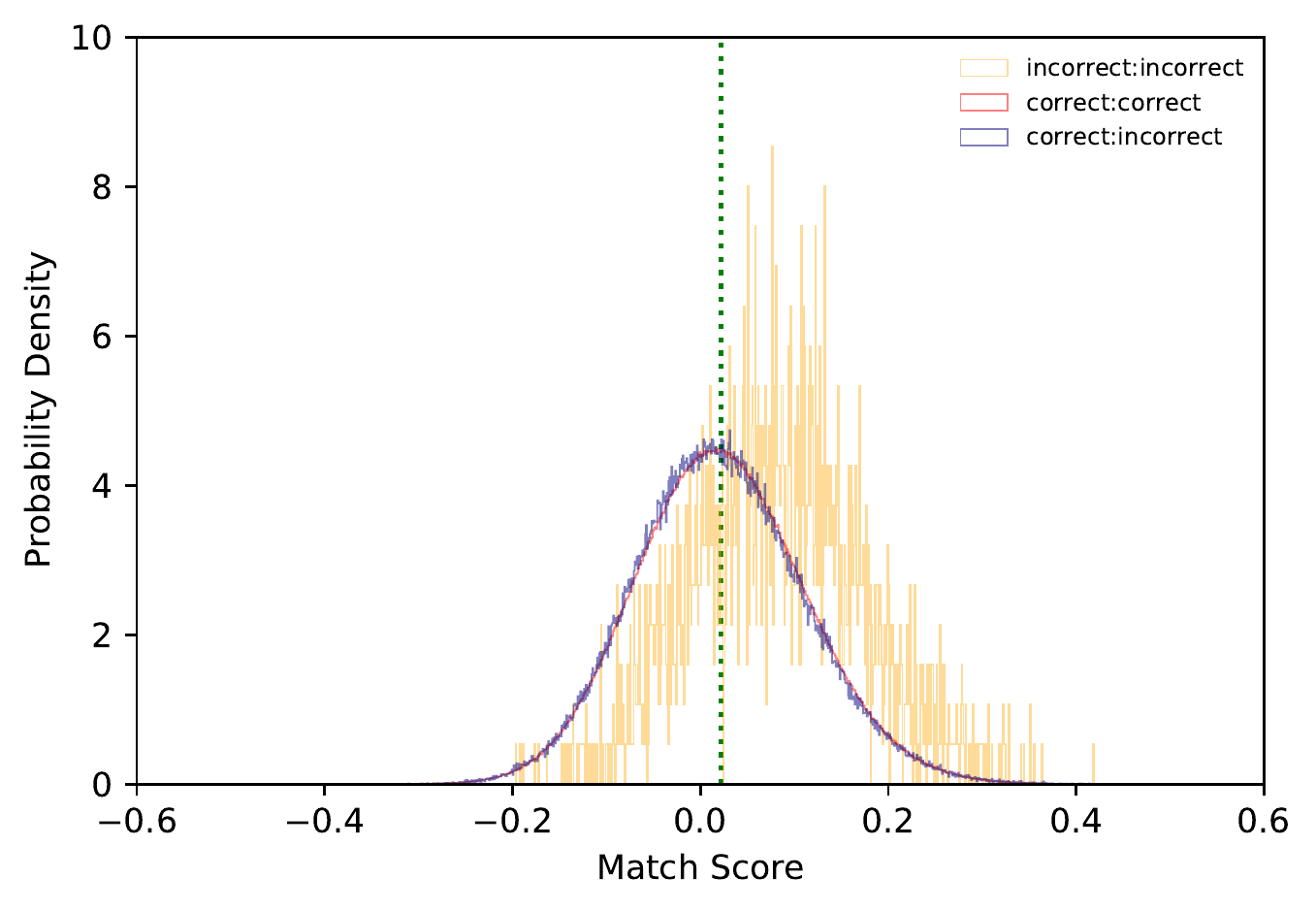}
      \end{subfigure}
      \begin{subfigure}[b]{0.239\linewidth}
        \centering
          \captionsetup[subfigure]{labelformat=empty}
          \caption{African-American Females}
          \vspace{-0.5em}
          \includegraphics[width=\linewidth]{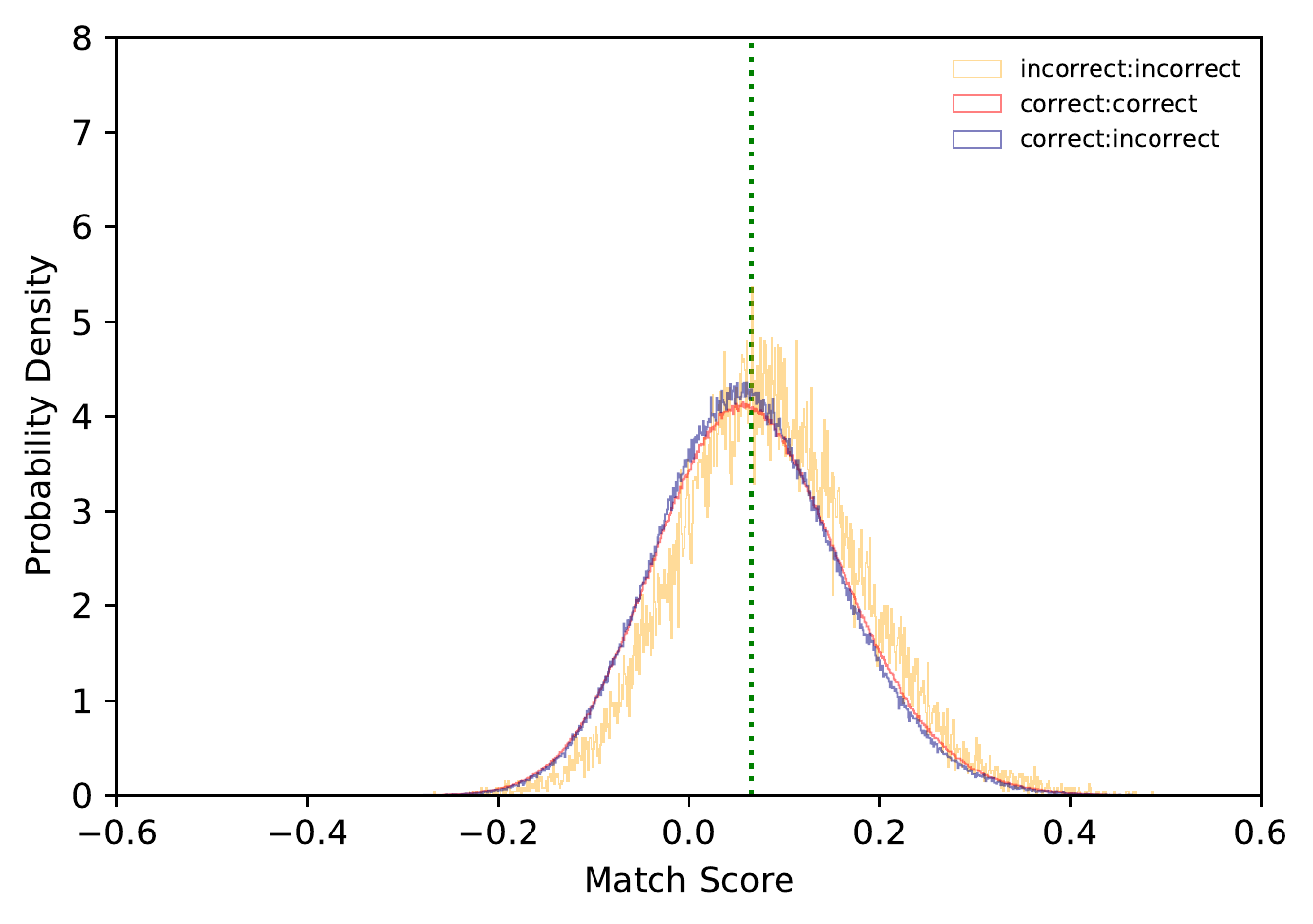}
      \end{subfigure}
      \addtocounter{subfigure}{-4}
      \vspace{-0.5em}
      \caption{ArcFace and Amazon Face API}
      \vspace{0.5em}
  \end{subfigure}
  \centering
  \begin{subfigure}[b]{1\linewidth}
      \centering
      \begin{subfigure}[b]{0.239\linewidth}
        \centering
          \includegraphics[width=\linewidth]{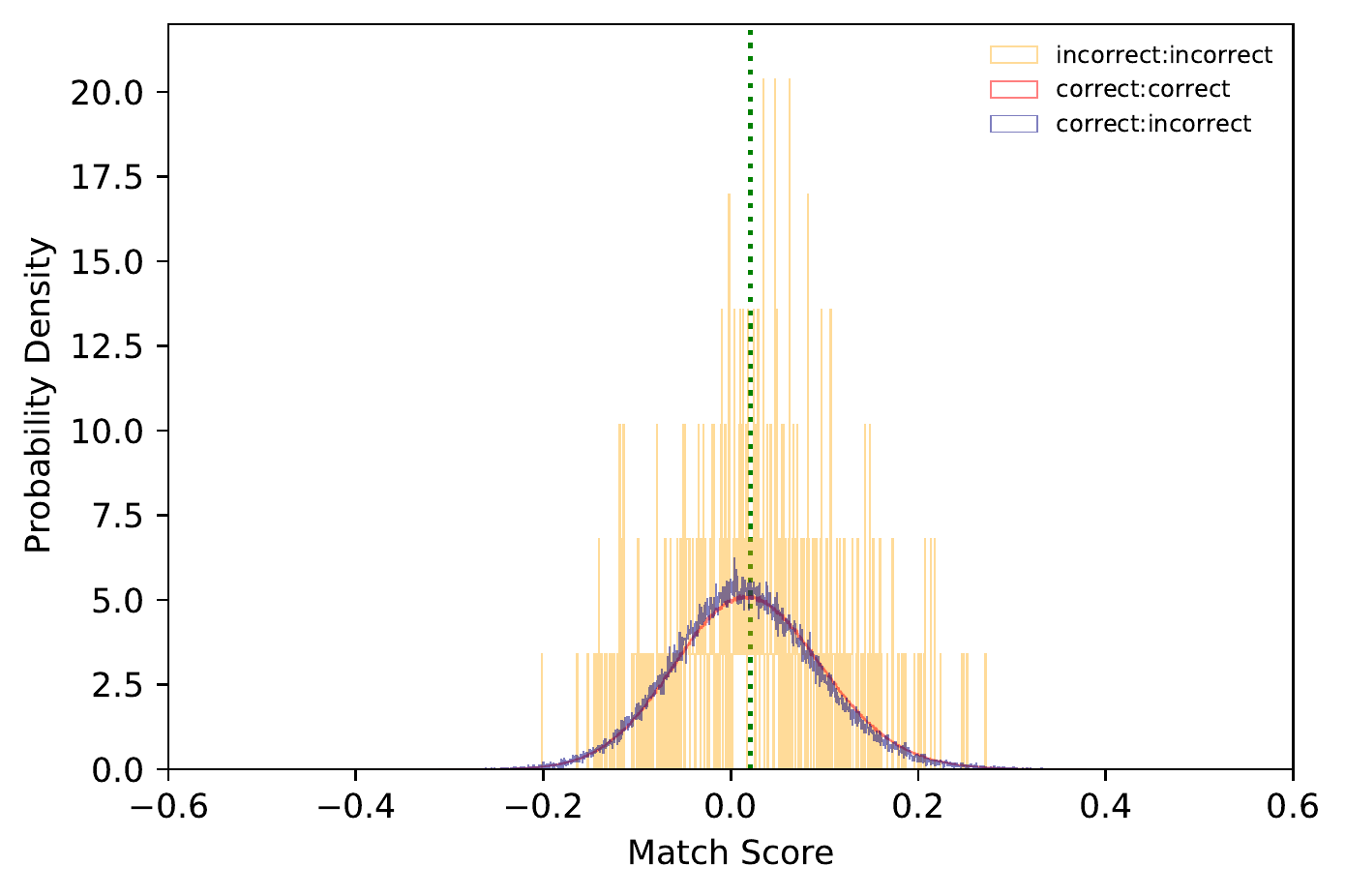}
      \end{subfigure}
      \begin{subfigure}[b]{0.239\linewidth}
        \centering
          \includegraphics[width=\linewidth]{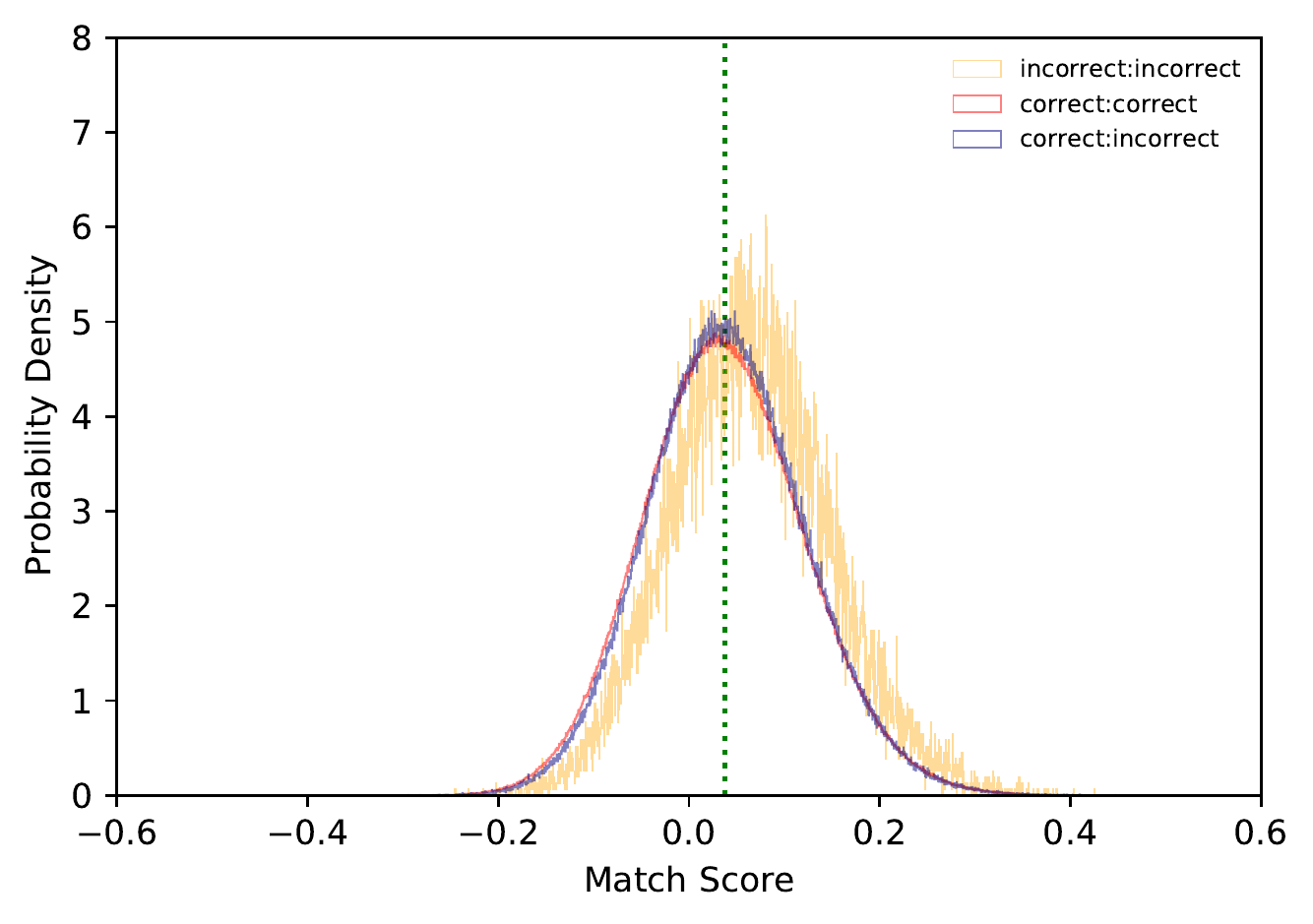}
      \end{subfigure}
      \centering
      \begin{subfigure}[b]{0.239\linewidth}
        \centering
          \includegraphics[width=\linewidth]{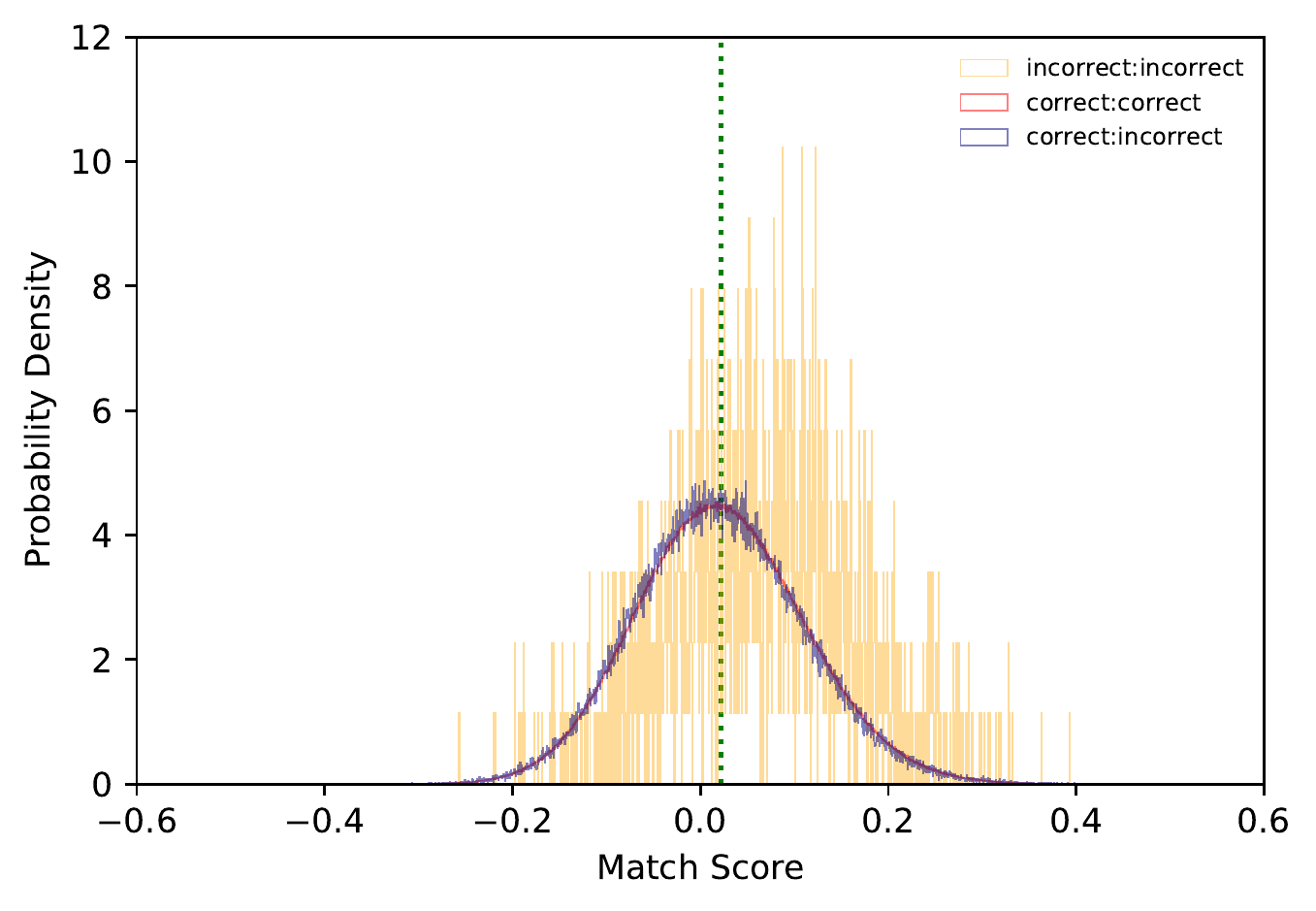}
      \end{subfigure}
      \begin{subfigure}[b]{0.239\linewidth}
        \centering
          \includegraphics[width=\linewidth]{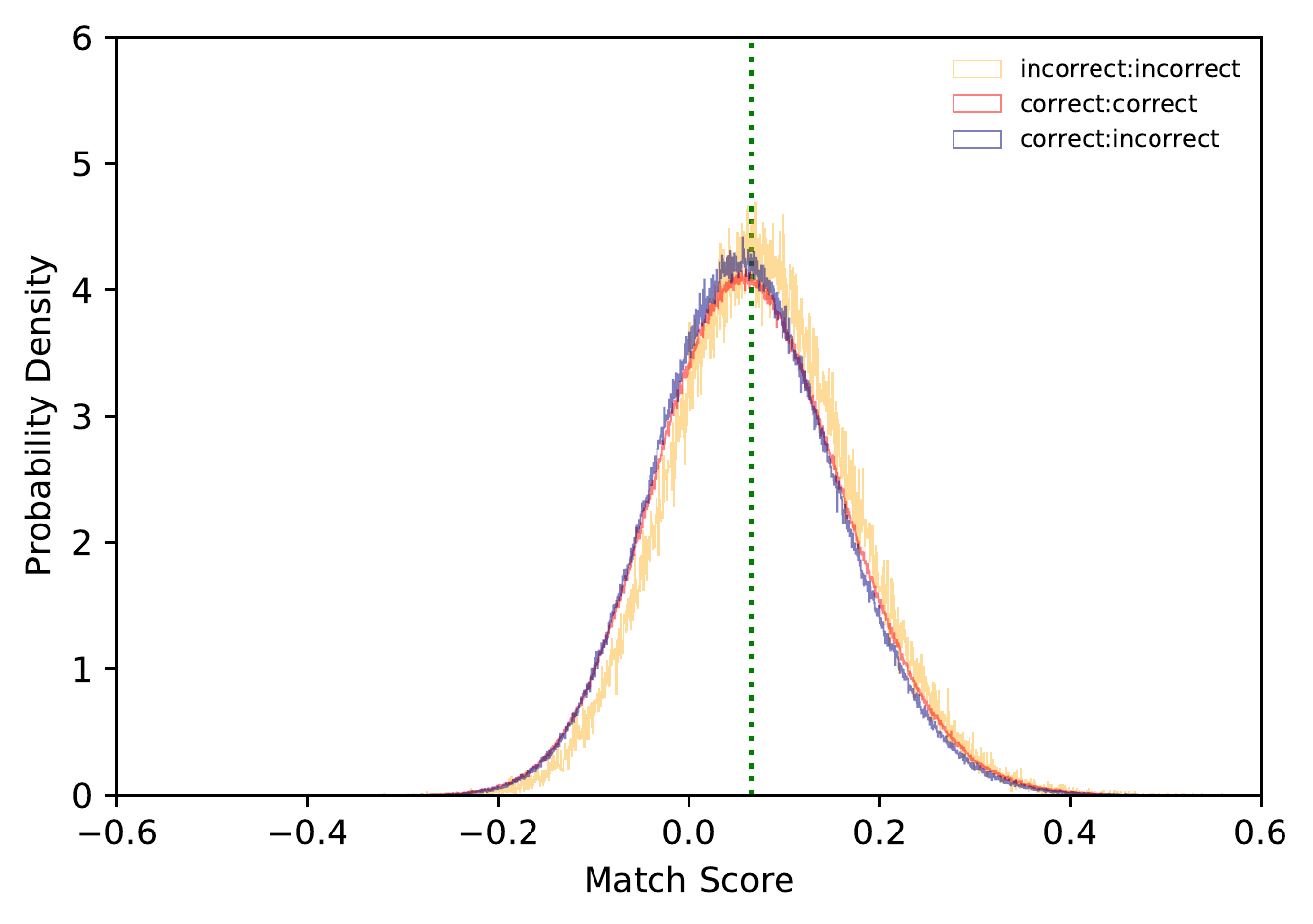}
      \end{subfigure}
      \vspace{-0.5em}
      \caption{ArcFace and Open Source}
      \vspace{0.5em}
  \end{subfigure}
  \begin{subfigure}[b]{1\linewidth}
      \centering
      \begin{subfigure}[b]{0.239\linewidth}
        \centering
          \includegraphics[width=\linewidth]{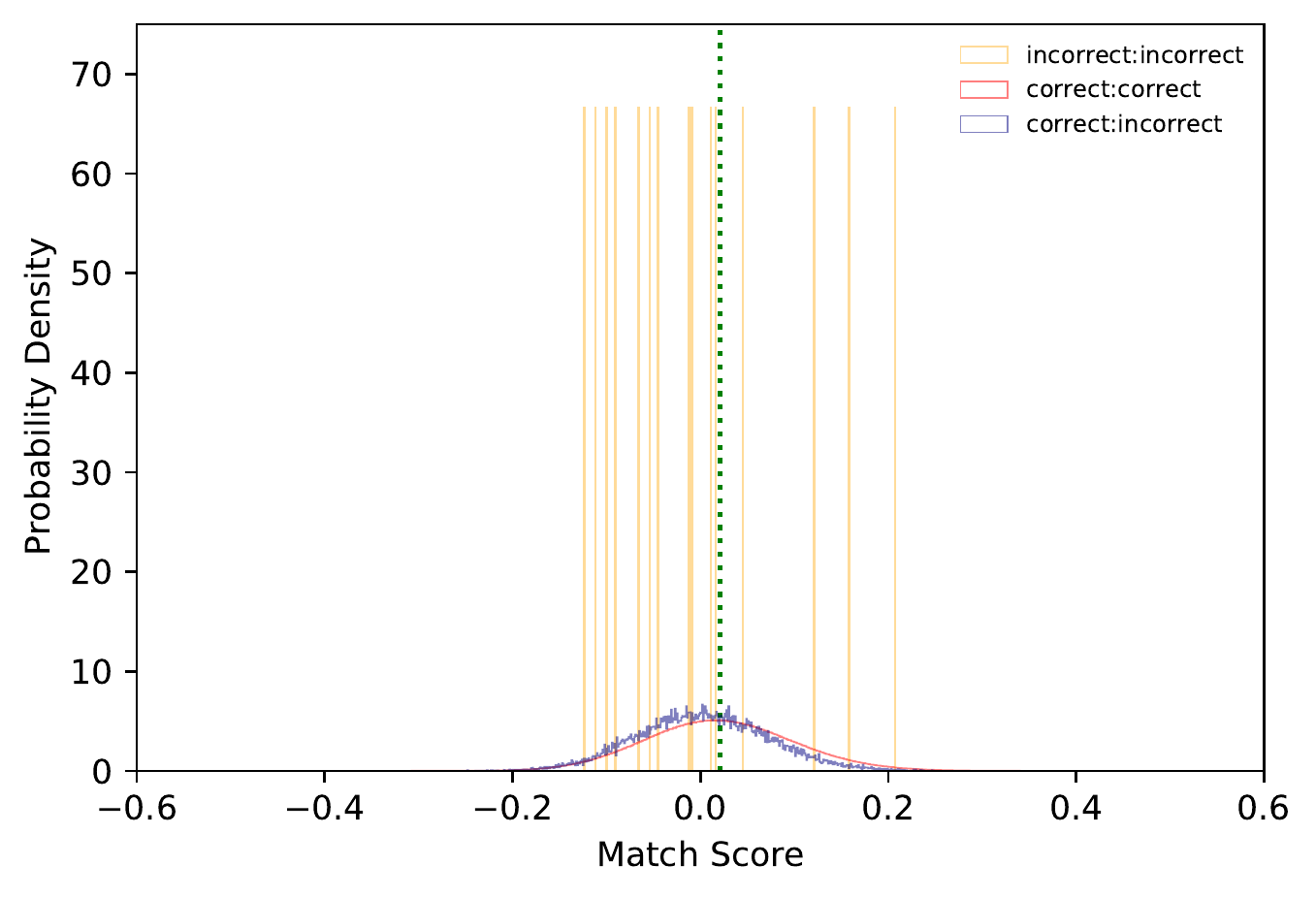}
      \end{subfigure}
      \begin{subfigure}[b]{0.239\linewidth}
        \centering
          \includegraphics[width=\linewidth]{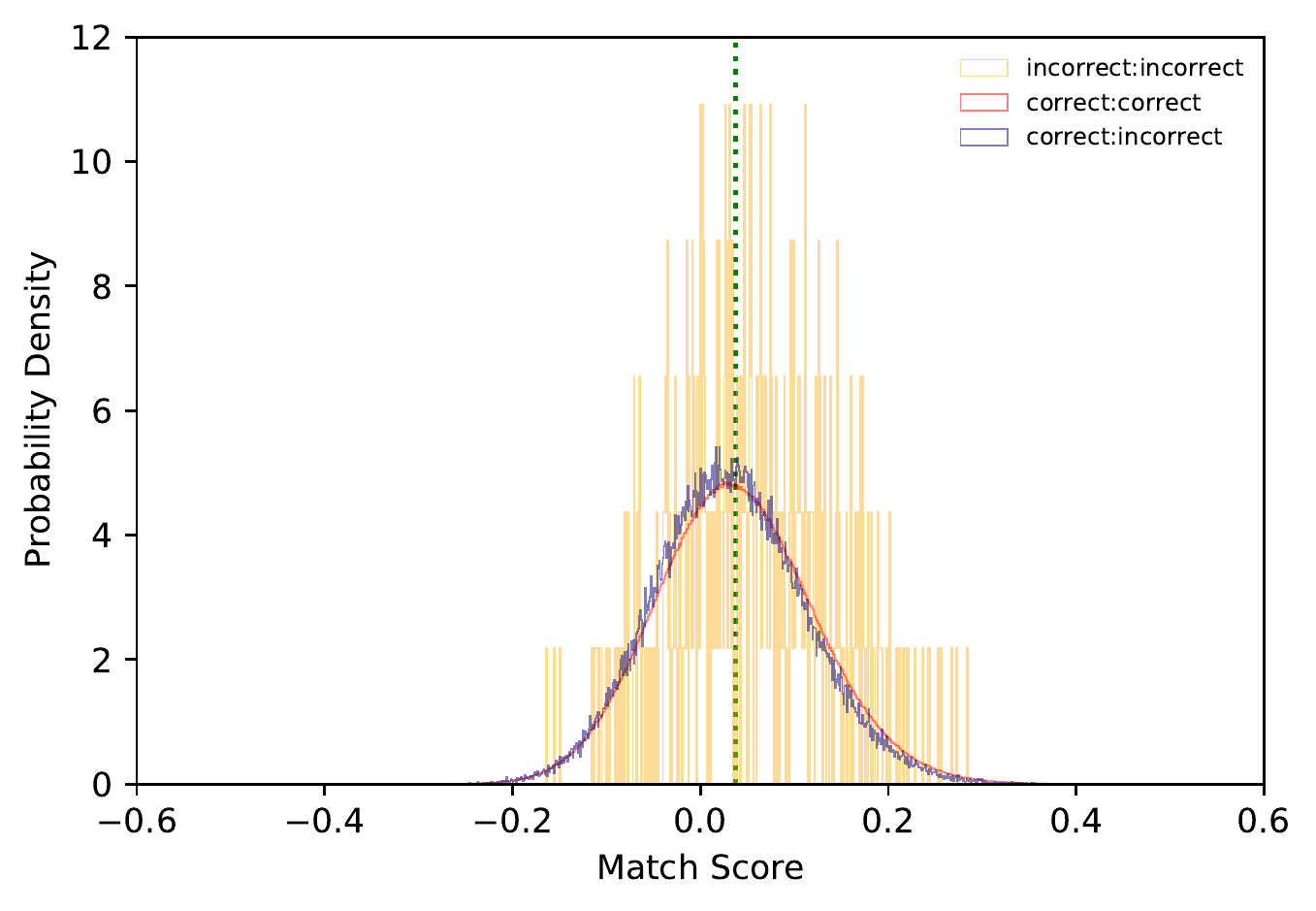}
      \end{subfigure}
      \centering
      \begin{subfigure}[b]{0.239\linewidth}
        \centering
          \includegraphics[width=\linewidth]{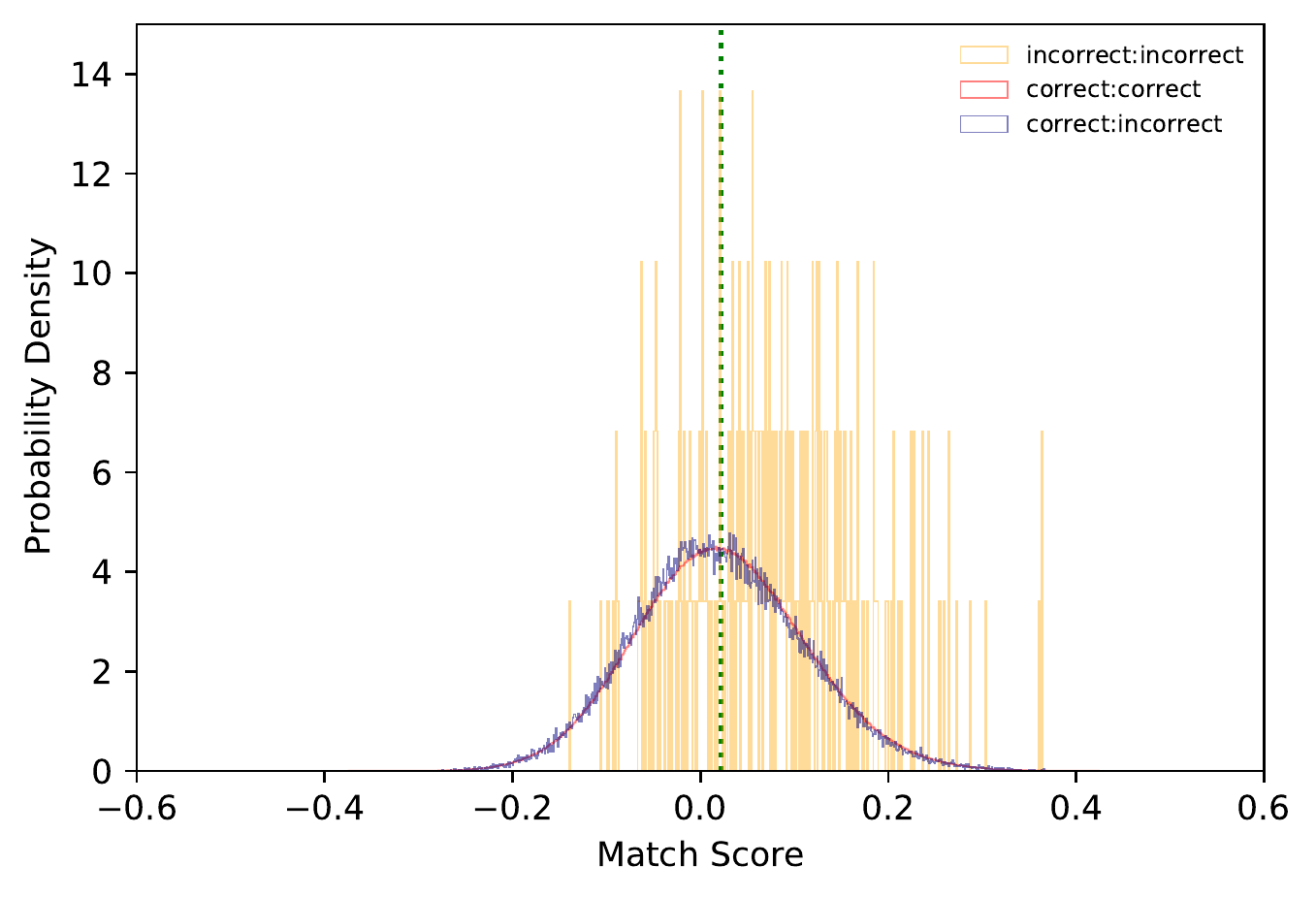}
      \end{subfigure}
      \begin{subfigure}[b]{0.239\linewidth}
        \centering
          \includegraphics[width=\linewidth]{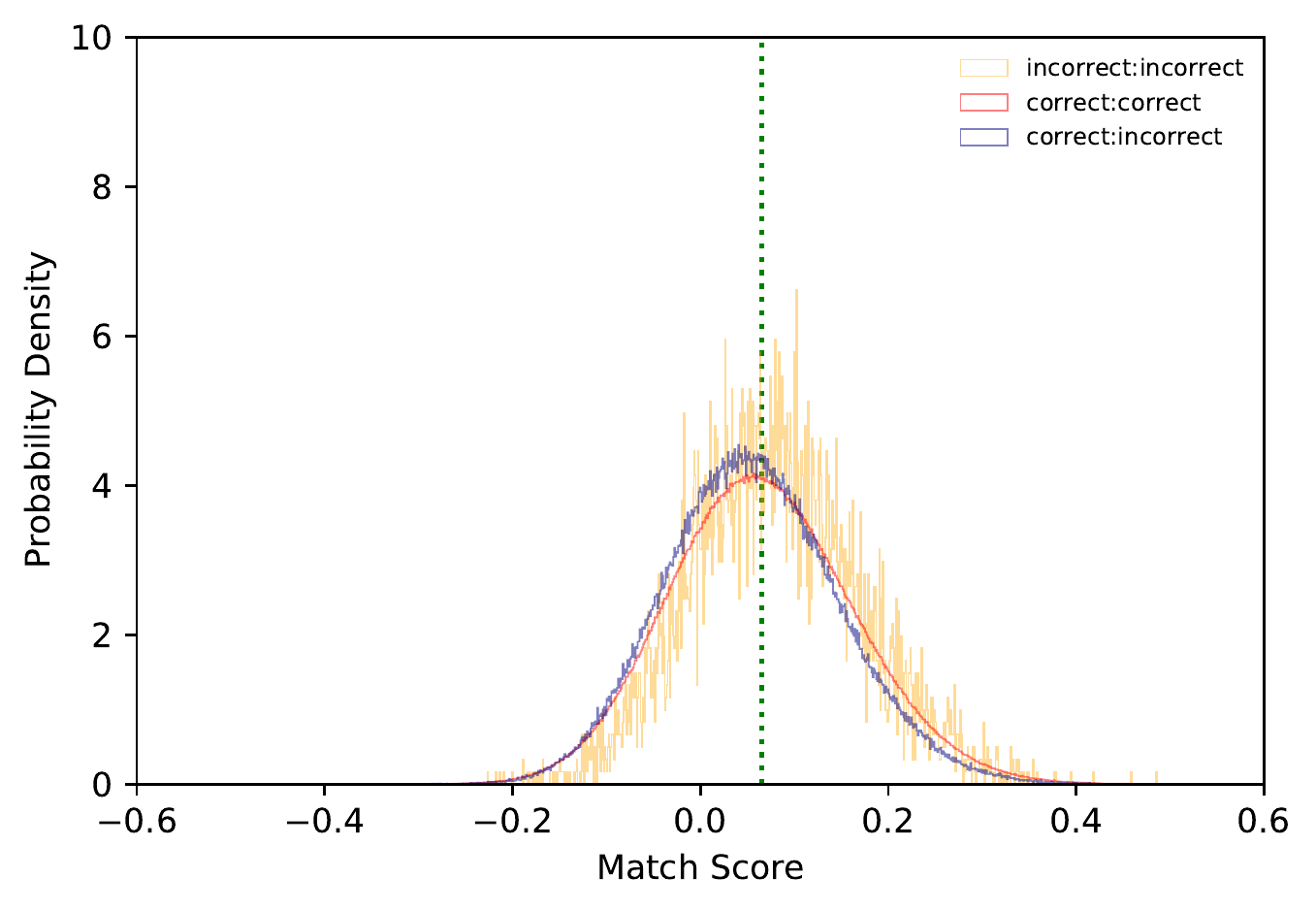}
      \end{subfigure}
      \vspace{-0.5em}
      \caption{ArcFace and Microsoft Face API}
      \vspace{-0.5em}
  \end{subfigure}
  \caption{Impostor distribution split into images without gender error, with at least one with gender error, and with both with gender error based on a MORPH subset that was balanced to have same number of subjects, with same amount of images, and same age distribution.}
  \label{fig:authentic}
\end{figure*}
\begin{figure*}[t]
  \centering
    \begin{subfigure}[b]{1\linewidth}
      \centering
      \captionsetup[subfigure]{labelformat=empty}
      \begin{subfigure}[b]{0.239\linewidth}
        \centering
          \vspace{-0.5em}
          \caption{Caucasian Males}
          \includegraphics[width=\linewidth]{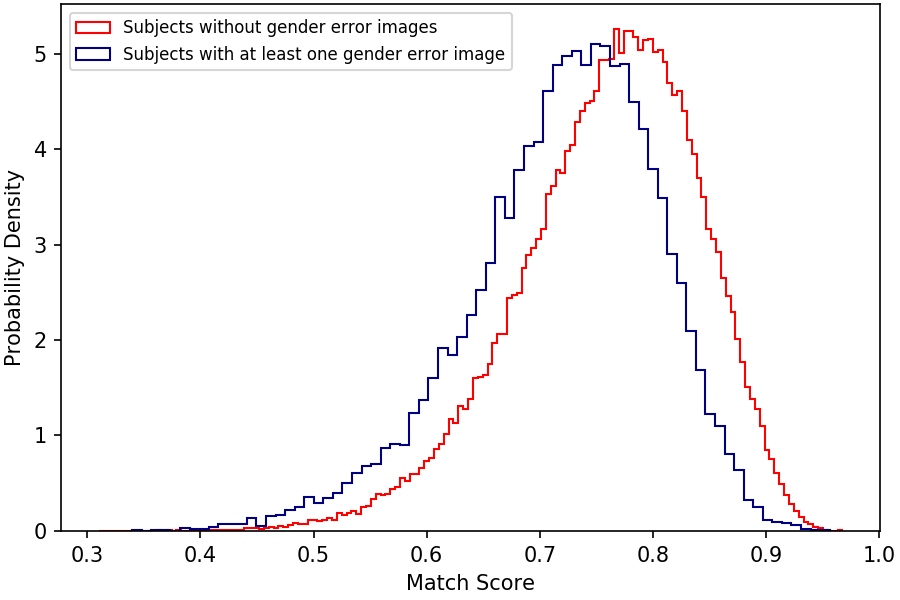}
      \end{subfigure}
      \begin{subfigure}[b]{0.239\linewidth}
        \centering
          \captionsetup[subfigure]{labelformat=empty}
          \vspace{-0.5em}
          \caption{Caucasian Females}
          \includegraphics[width=\linewidth]{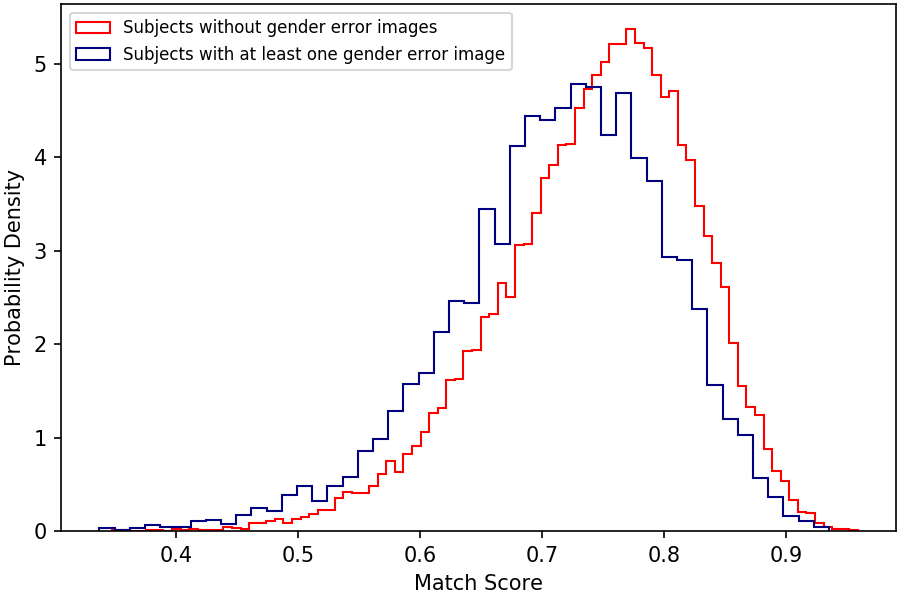}
      \end{subfigure}
      \centering
      \begin{subfigure}[b]{0.239\linewidth}
        \centering
          \captionsetup[subfigure]{labelformat=empty}
          \vspace{-0.5em}
          \caption{African-American Males}
          \includegraphics[width=\linewidth]{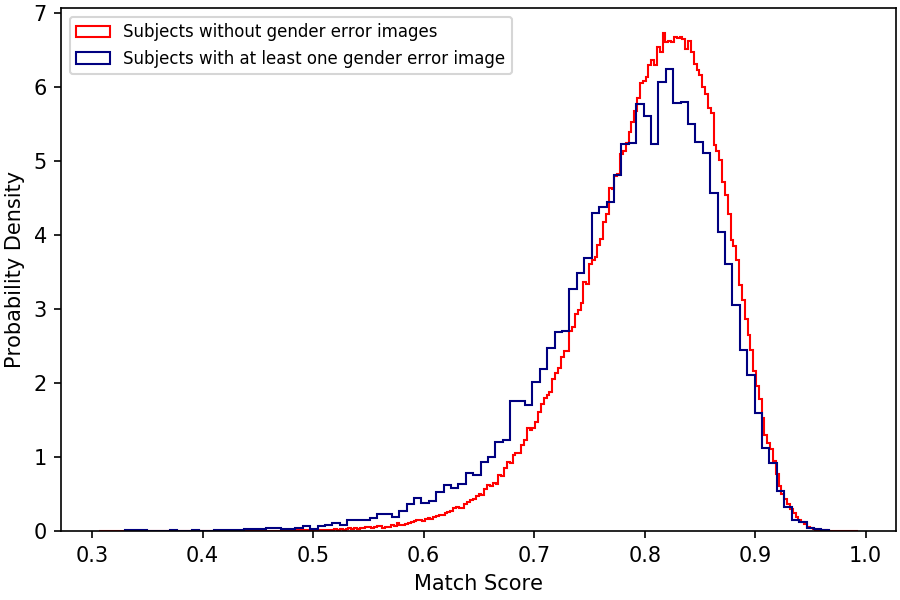}
      \end{subfigure}
      \begin{subfigure}[b]{0.239\linewidth}
        \centering
          \captionsetup[subfigure]{labelformat=empty}
          \vspace{-0.5em}
          \caption{African-American Females}
          \includegraphics[width=\linewidth]{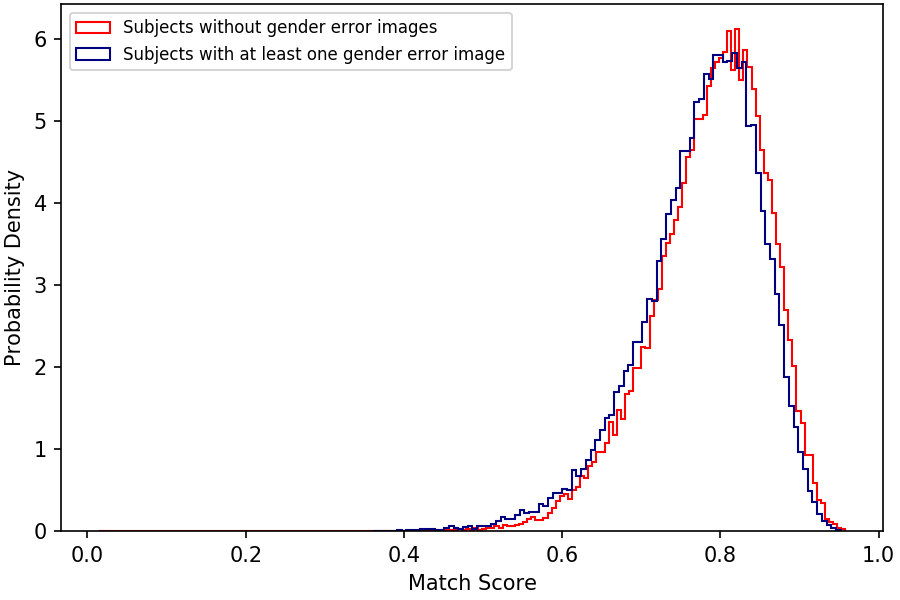}
      \end{subfigure}
      \addtocounter{subfigure}{-4}
      \caption{ArcFace and Amazon Face API}
      \vspace{0.5em}
  \end{subfigure}
  \centering
  \begin{subfigure}[b]{1\linewidth}
      \centering
      \begin{subfigure}[b]{0.239\linewidth}
        \centering
          \includegraphics[width=\linewidth]{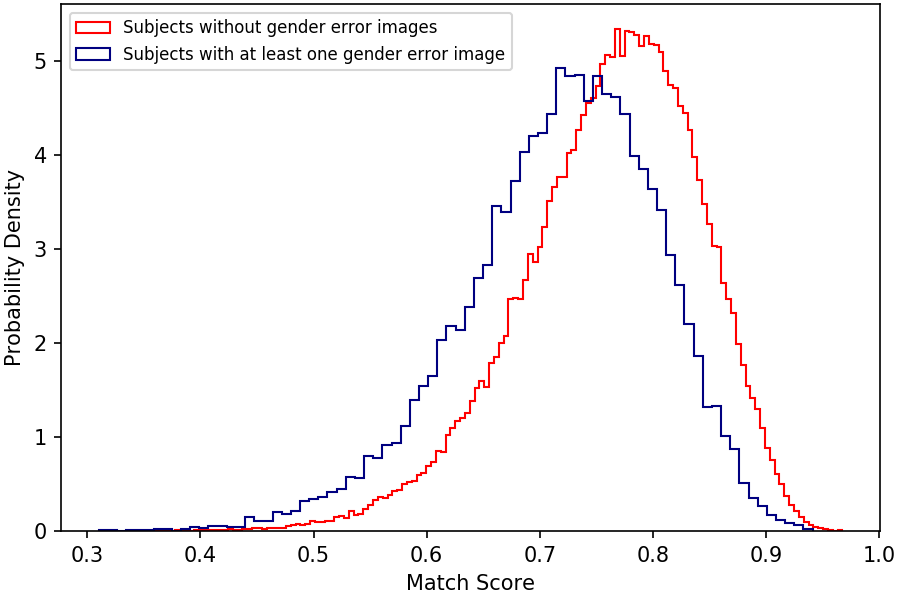}
      \end{subfigure}
      \begin{subfigure}[b]{0.239\linewidth}
        \centering
          \includegraphics[width=\linewidth]{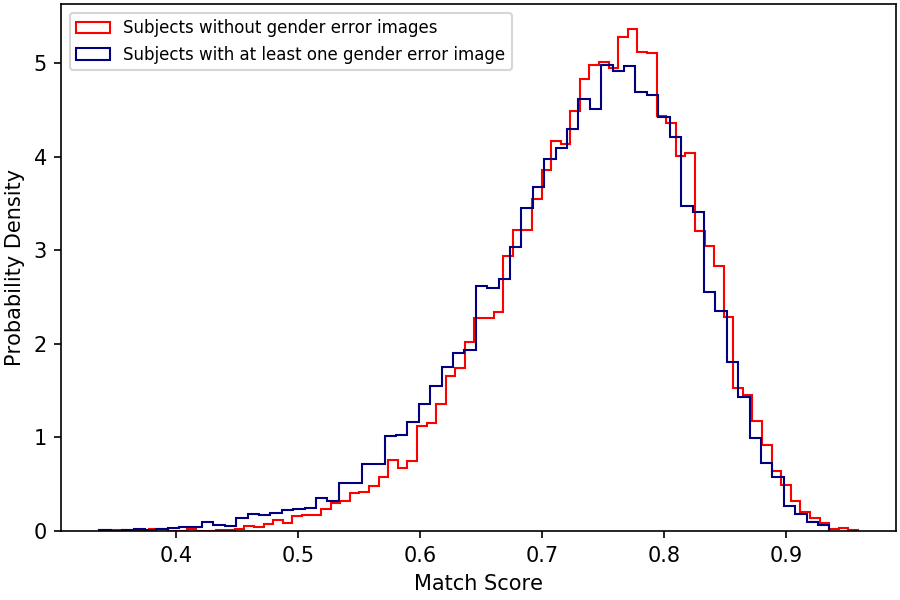}
      \end{subfigure}
      \centering
      \begin{subfigure}[b]{0.239\linewidth}
        \centering
          \includegraphics[width=\linewidth]{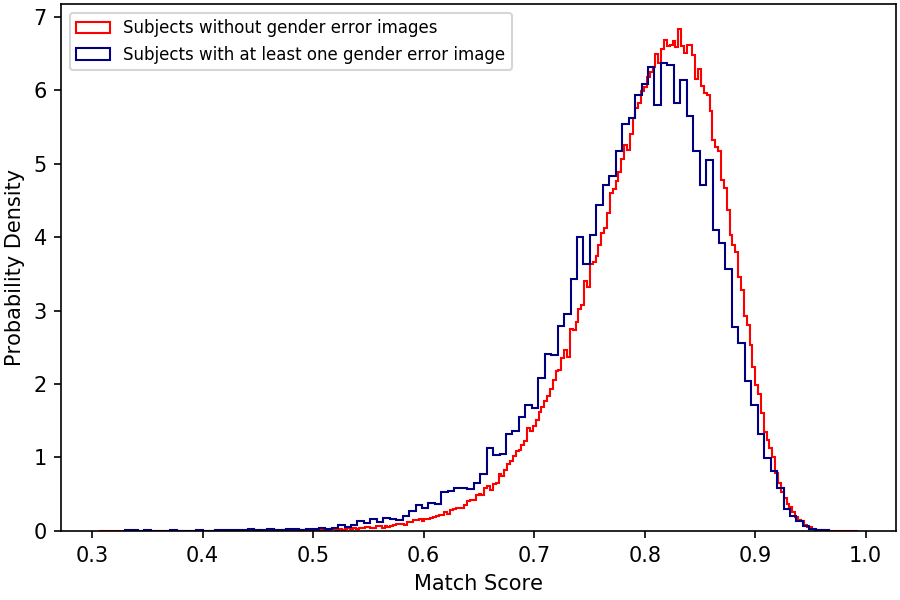}
      \end{subfigure}
      \begin{subfigure}[b]{0.239\linewidth}
        \centering
          \includegraphics[width=\linewidth]{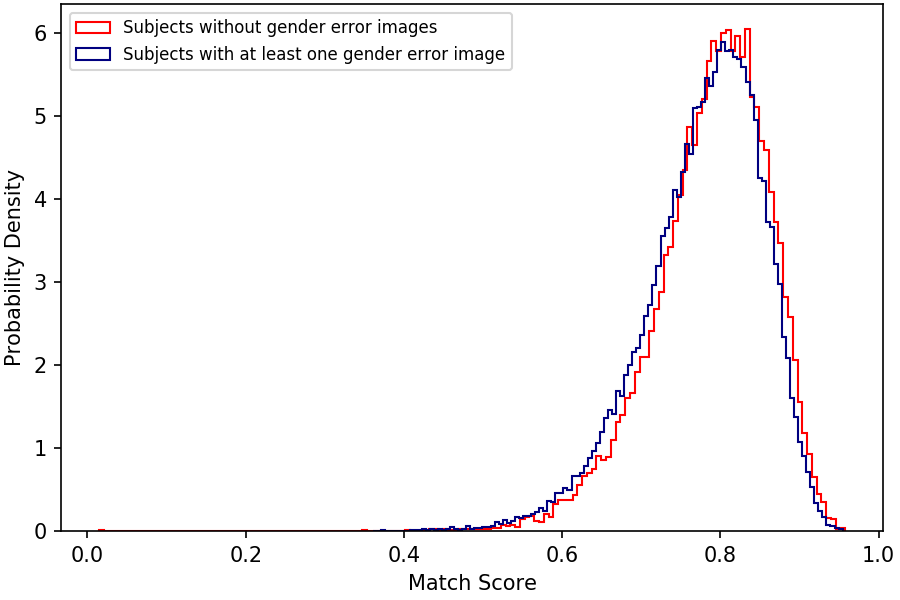}
      \end{subfigure}
      \caption{ArcFace and Open Source}
      \vspace{0.5em}
  \end{subfigure}
  \begin{subfigure}[b]{1\linewidth}
      \centering
      \begin{subfigure}[b]{0.239\linewidth}
        \centering
          \includegraphics[width=\linewidth]{figures/authentic/subject_based/arcface/hist_C_M_Microsoft_arcface.png}
      \end{subfigure}
      \begin{subfigure}[b]{0.239\linewidth}
        \centering
          \includegraphics[width=\linewidth]{figures/authentic/subject_based/arcface/hist_C_F_Microsoft_arcface.png}
      \end{subfigure}
      \centering
      \begin{subfigure}[b]{0.239\linewidth}
        \centering
          \includegraphics[width=\linewidth]{figures/authentic/subject_based/arcface/hist_AA_M_Microsoft_arcface.png}
      \end{subfigure}
      \begin{subfigure}[b]{0.239\linewidth}
        \centering
          \includegraphics[width=\linewidth]{figures/authentic/subject_based/arcface/hist_AA_F_Microsoft_arcface.png}
      \end{subfigure}
      \caption{ArcFace and Microsoft Face API}
      \vspace{0.5em}
  \end{subfigure}
  \begin{subfigure}[b]{1\linewidth}
      \centering
      \begin{subfigure}[b]{0.239\linewidth}
        \centering
          \includegraphics[width=\linewidth]{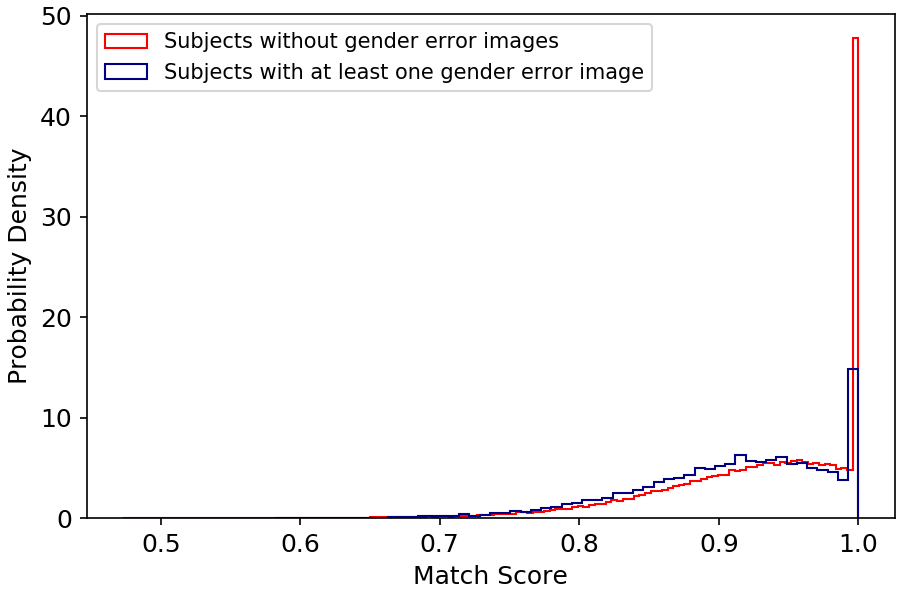}
      \end{subfigure}
      \begin{subfigure}[b]{0.239\linewidth}
        \centering
          \includegraphics[width=\linewidth]{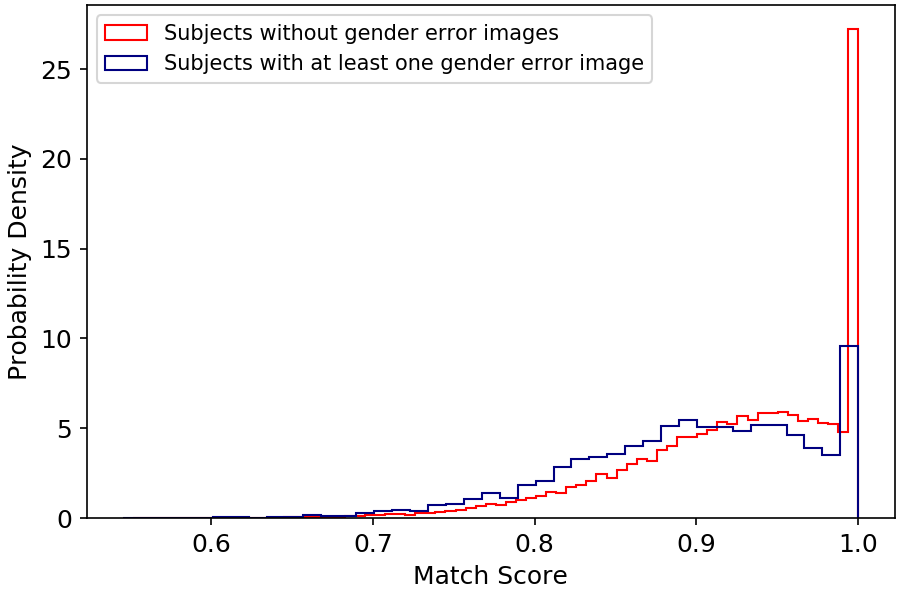}
      \end{subfigure}
      \centering
      \begin{subfigure}[b]{0.239\linewidth}
        \centering
          \includegraphics[width=\linewidth]{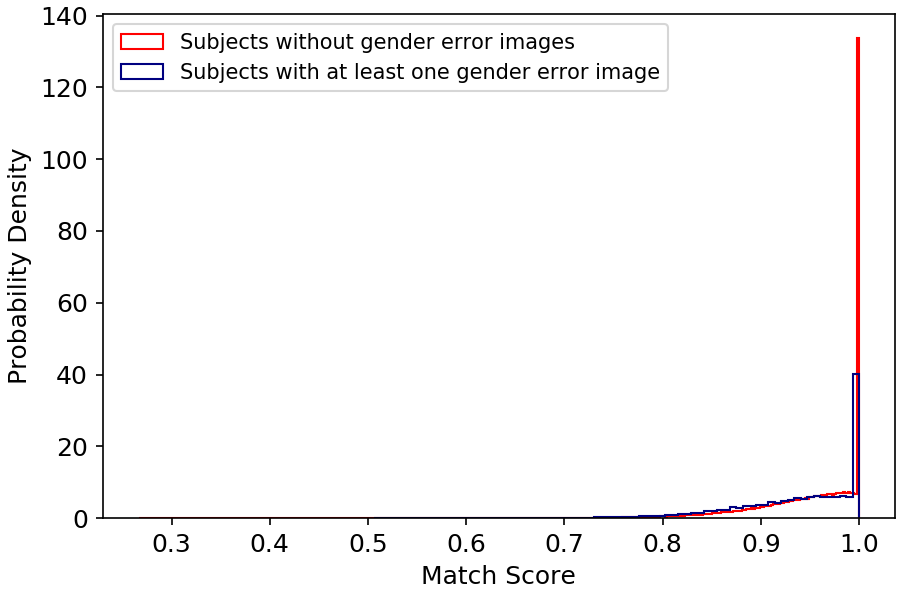}
      \end{subfigure}
      \begin{subfigure}[b]{0.239\linewidth}
        \centering
          \includegraphics[width=\linewidth]{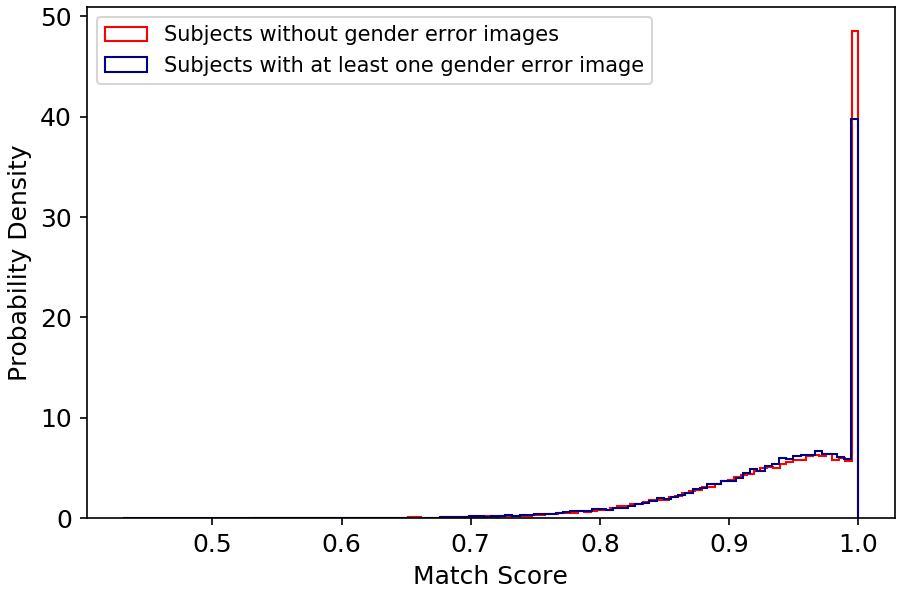}
      \end{subfigure}
      \caption{COTS and Amazon Face API}
      \vspace{0.5em}
  \end{subfigure}
  \begin{subfigure}[b]{1\linewidth}
      \centering
      \begin{subfigure}[b]{0.239\linewidth}
        \centering
          \includegraphics[width=\linewidth]{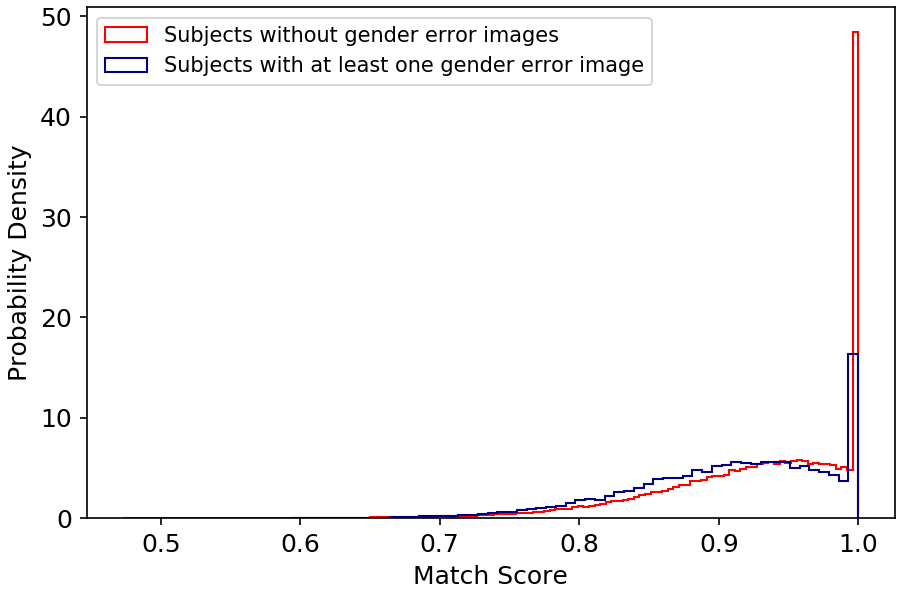}
      \end{subfigure}
      \begin{subfigure}[b]{0.239\linewidth}
        \centering
          \includegraphics[width=\linewidth]{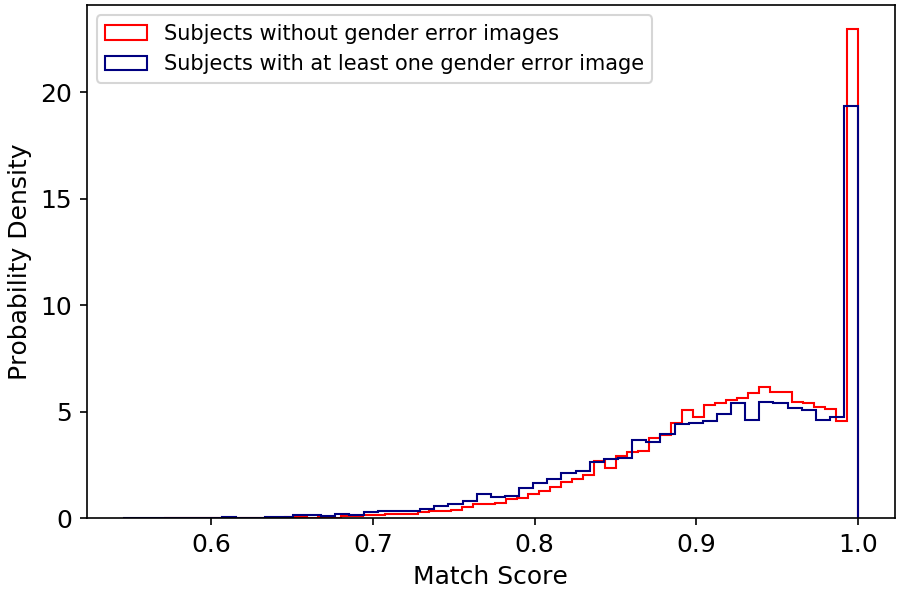}
      \end{subfigure}
      \centering
      \begin{subfigure}[b]{0.239\linewidth}
        \centering
          \includegraphics[width=\linewidth]{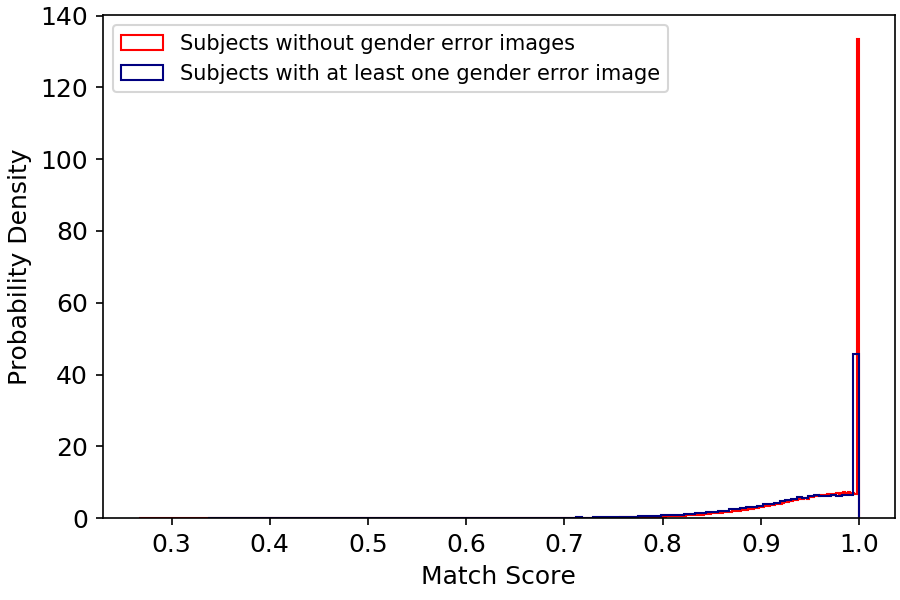}
      \end{subfigure}
      \begin{subfigure}[b]{0.239\linewidth}
        \centering
          \includegraphics[width=\linewidth]{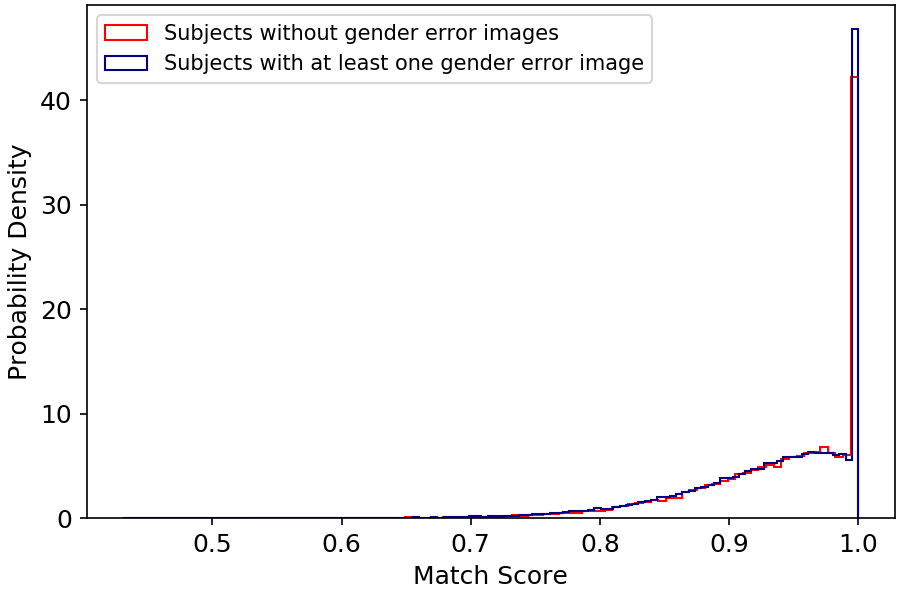}
      \end{subfigure}
      \caption{COTS and Open Source}
      \vspace{0.5em}
  \end{subfigure}
  \begin{subfigure}[b]{1\linewidth}
      \centering
      \begin{subfigure}[b]{0.239\linewidth}
        \centering
          \includegraphics[width=\linewidth]{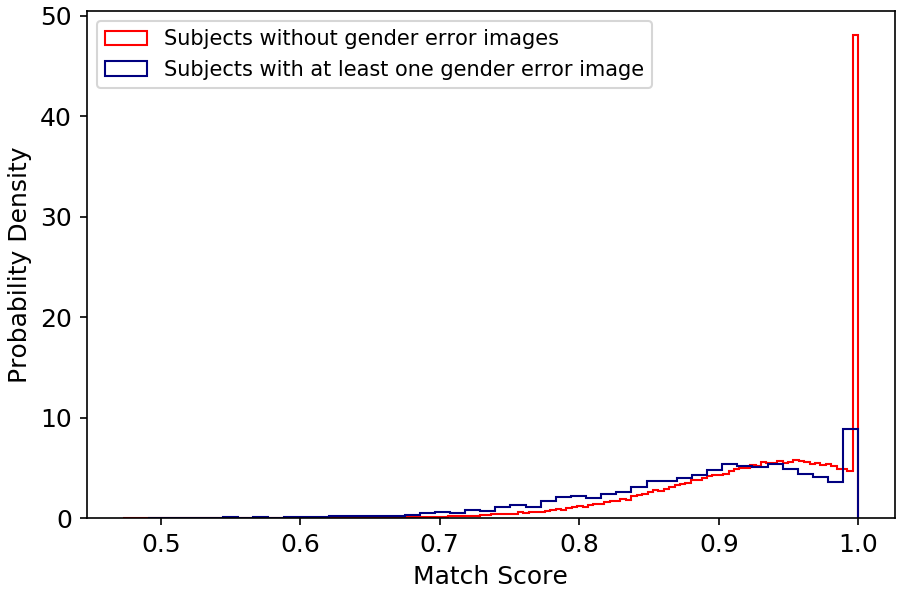}
      \end{subfigure}
      \begin{subfigure}[b]{0.239\linewidth}
        \centering
          \includegraphics[width=\linewidth]{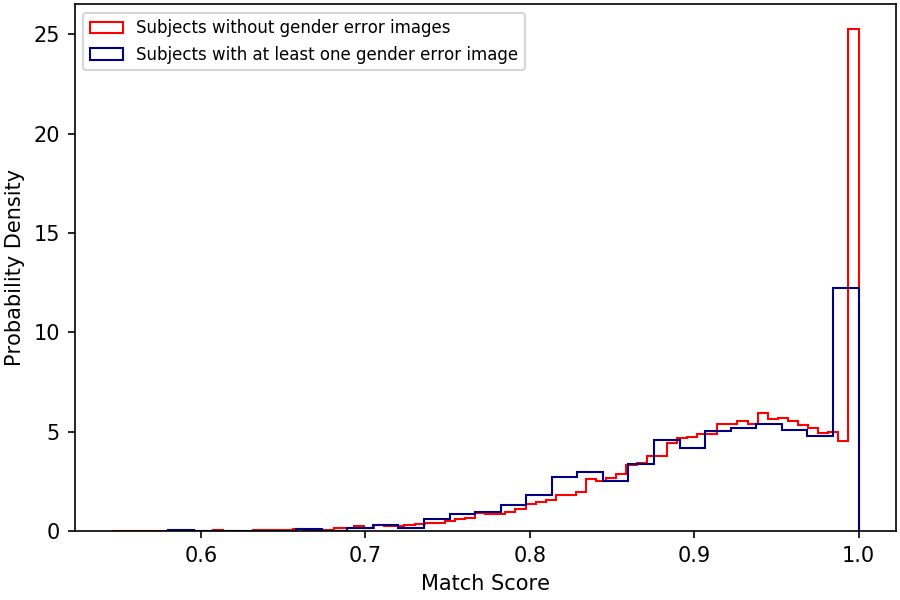}
      \end{subfigure}
      \centering
      \begin{subfigure}[b]{0.239\linewidth}
        \centering
          \includegraphics[width=\linewidth]{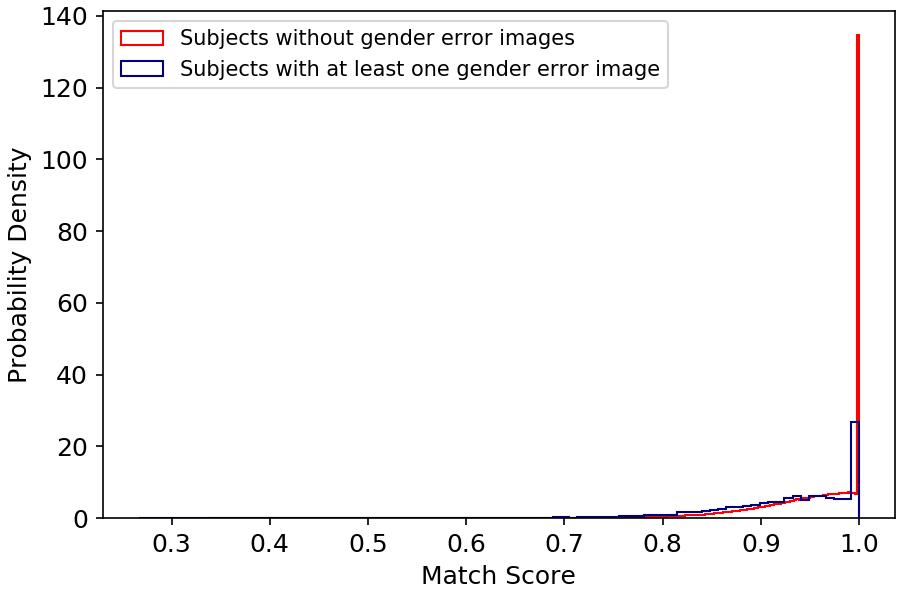}
      \end{subfigure}
      \begin{subfigure}[b]{0.239\linewidth}
        \centering
          \includegraphics[width=\linewidth]{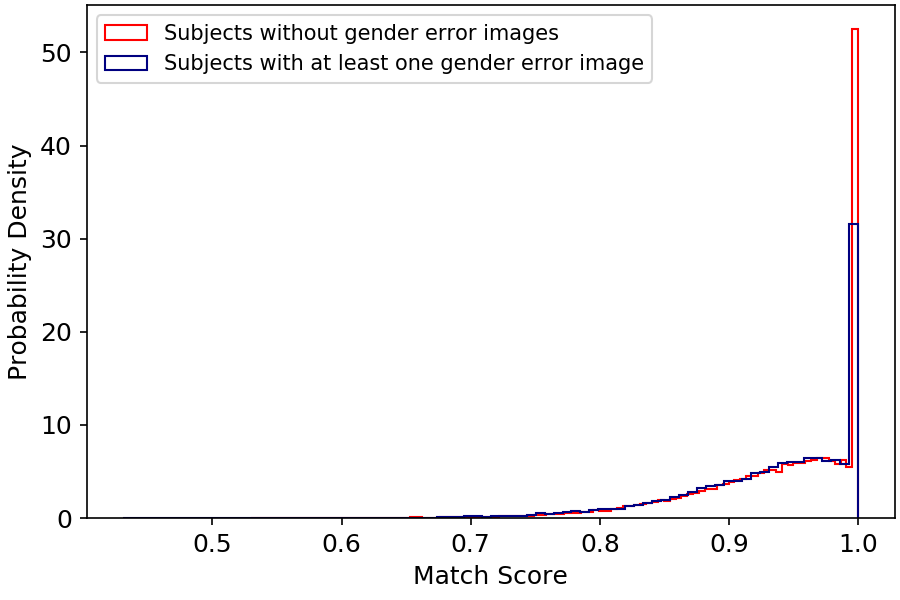}
      \end{subfigure}
      \caption{COTS and Microsoft Face API}
      \vspace{0.5em}
  \end{subfigure}
 
  \caption{Authentic distribution split into images without gender error, with at least one with gender error, and with both with gender error.}
  \label{fig:authentic}
\end{figure*}

\end{document}